\documentclass[10pt,twocolumn,letterpaper]{article}

\usepackage[pagenumbers]{cvpr} % To force page numbers, e.g. for an arXiv version

\usepackage{soul}
\usepackage{array}
\definecolor{cvprblue}{rgb}{0.21,0.49,0.74}
\usepackage[pagebackref,breaklinks,colorlinks,allcolors=cvprblue]{hyperref}
\usepackage{graphicx}
\usepackage{amsmath}
\usepackage{amssymb}
\usepackage{comment}
\usepackage{booktabs,multirow}
\usepackage{bm}
\usepackage{relsize}
\def\paperID{*****} % *** Enter the Paper ID here
\def\confName{CVPR}
\def\confYear{2026}

\title{A Difference-in-Difference Approach to Detecting AI-Generated Images}

\author{
Xinyi Qi\thanks{Xinyi Qi and Kai Ye contributed equally to this work and are listed in alphabetical order.}\\
Tsinghua University\\
{\tt\small qixy25@mails.tsinghua.edu.cn}
\and
Kai Ye\footnotemark[1]\\
The London School of Economics and Political Science\\
{\tt\small k.ye1@lse.ac.uk}
\and
Chengchun Shi\\
The London School of Economics and Political Science\\
{\tt\small c.shi7@lse.ac.uk}
\and
Ying Yang\\
Tsinghua University\\
{\tt\small yangying@tsinghua.edu.cn}
\and
Hongyi Zhou\thanks{Corresponding authors: Hongyi Zhou and Jin Zhu are listed in alphabetical order. Emails: zhou-hy21@mails.tsinghua.edu.cn, j.zhu.7@bham.ac.uk}\\
Tsinghua University\\
{\tt\small zhou-hy21@mails.tsinghua.edu.cn}
\and
Jin Zhu\footnotemark[2]\\
University of Birmingham\\
{\tt\small j.zhu.7@bham.ac.uk}
}

\begin{document}
\maketitle
\begin{abstract}
   Diffusion models are able to produce AI-generated images that are almost indistinguishable from real ones. This raises concerns about their potential misuse and poses substantial challenges for detecting them. Many existing detectors rely on reconstruction error — the difference between the input image and its reconstructed version — as the basis for distinguishing real from fake images. However, these detectors become less effective as modern AI-generated images become increasingly similar to real ones. To address this challenge, we propose a novel difference-in-difference method. Instead of directly using the reconstruction error (a first-order difference), we compute the difference in reconstruction error -- a second-order difference -- for variance reduction and improving detection accuracy. Extensive experiments demonstrate that our method achieves strong generalization performance, enabling reliable detection of AI-generated images in the era of generative AI.
   %ion approaches rely on geometric assumptions about the discrepancy between real and synthetic image spaces. However, this assumption become less effective as modern generators achieve higher performance and the gap between the two spaces narrows. To address this limitation, we propose a novel Difference-in-Difference (DID) detection framework that capture the difference in variance of reconstruction residuals between real and fake images. Our method measures reconstruction stability across repeated reconstructions and introduces a Dual-Classifier framework that jointly exploits geometric and variance-based features for complementary robustness. Extensive experiments across multiple diffusion backbones and post-processing conditions demonstrate that the proposed approach achieves strong generalization and consistent performance, providing a principled solution for detecting AI-generated images in the era of advanced diffusion models.
\end{abstract}    
\section{Introduction}\label{sec:intro}
The rapid evolution of large-scale generative models has led to groundbreaking progress in image generation in recent years \cite{arkhipkin2023kandinsky,betker2023improving,baldridge2024imagen,sheynin2024emu,bandyopadhyay2025sd3}
 . Diffusion models, in particular, have become the foundation of modern image generation. %including Generative Adversarial Networks (GANs), Variational Autoencoders (VAEs) and Denoising Diffusion Models (DMs) . 
%Machine-generated images have achieved unprecedented levels of realism, diversity, and controllability.  
However, the proliferation of AI-generated images has raised substantial concerns about authenticity and public trust, creating an urgent need for effective detectors that can distinguish AI-generated images from real ones.
\begin{figure}[t]
  \centering
   \includegraphics[width=1\linewidth]{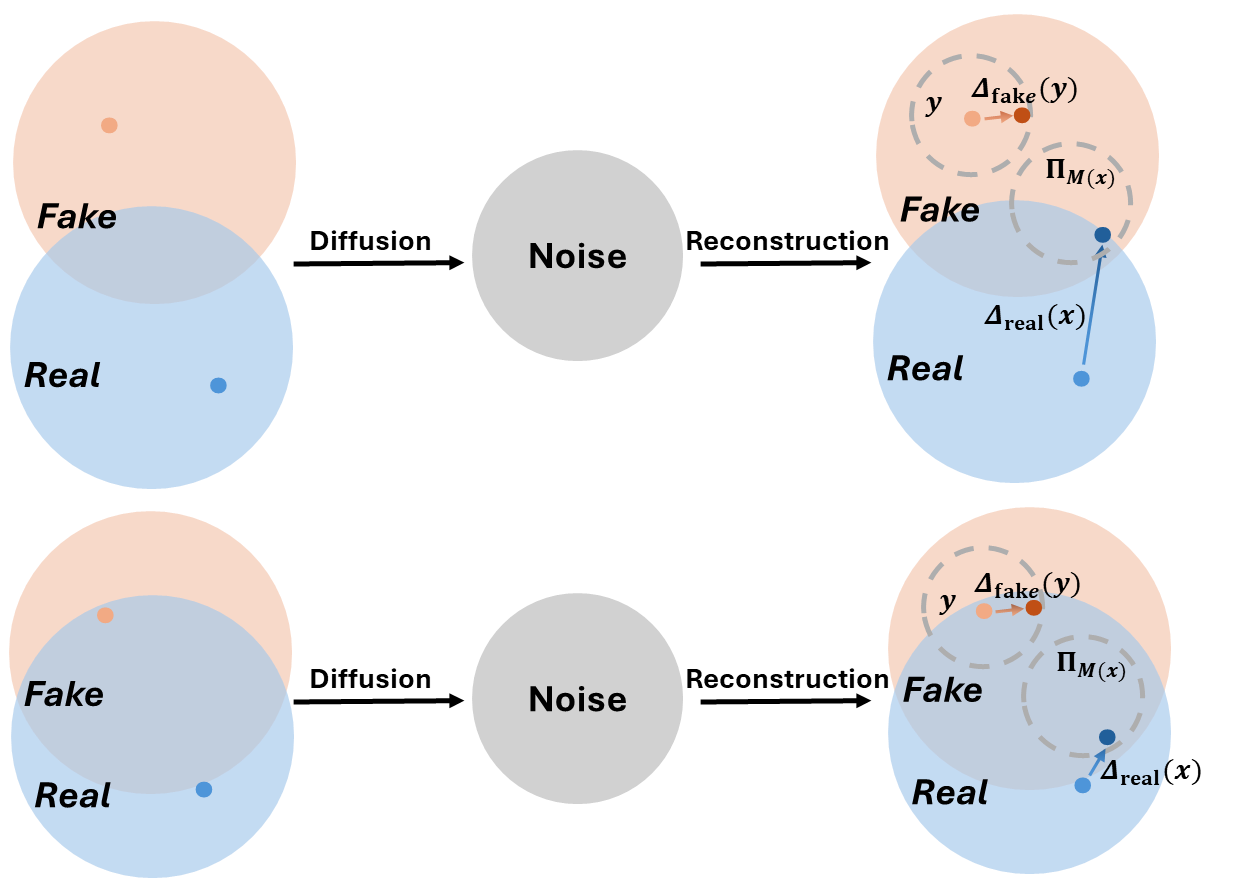}
   \caption{\textbf{Left:} The real image $x$ and its distribution on the manifold $\mathcal{M}$ are shown in blue, whereas the synthetic image $y$ and its distribution are shown in orange.  
\textbf{Right:} Reconstructions of the real and synthetic images. $\Delta_{\textrm{fake}}$ and $\Delta_{\textrm{real}}$ denote the reconstruction errors of synthetic and real images, respectively, and $\Pi_{\mathcal{M}}$ represents the projection of the real image onto $\mathcal{M}$.  
\textbf{Top:} The generator is weak, resulting in a large discrepancy between the real and synthetic image distributions. Consequently, the reconstruction error of the synthetic image, $\Delta_{\textrm{fake}}(y)$, is much larger than that of the real image, $\Delta_{\textrm{real}}(x)$, making detection easier.  
\textbf{Bottom:} The generator is strong, so the real and synthetic image distributions are close. Here, $\Delta_{\textrm{fake}}(y)$ is similar to $\Delta_{\textrm{real}}(x)$, making AI-generated images difficult to detect.}\label{fig:illustration}
\end{figure}
There is a growing literature on detecting AI-generated images; see Section \ref{sec:relatedworks} for a detailed review. Early detection algorithms were tailored for specific models and exploited model-based artifacts — such as inconsistencies in the frequency domain, irregularities in color filter responses, or statistical traces introduced by up-sampling layers \cite{frank2020leveraging,zhang2019detecting,corvi2023intriguing}. Although these methods achieved strong performance in detecting images from specific generators, they lacked robustness across different models. %and generated content (e.g., portraits, medical images, satellite data). 

More recently, reconstruction-based algorithms have emerged as both conceptually intuitive and effective practically. As suggested by their name, these methods (a) reconstruct the image to be detected using a generative model (e.g., a diffusion model), (b) compute the reconstruction error as the difference between the original and reconstructed images, and (c) classify the image as real or synthetic based on that error. Their core idea is straightforward: given a particular generative model for reconstruction, synthetic images that lie on or near the model manifold should be reconstructed with smaller residual errors than real images, which reside outside that manifold (see the upper panel of Figure~\ref{fig:illustration} for an illustration). This line of work, represented by DIRE~\cite{Wang2023DIRE} and its variants LaRE$^{2}$ and FIRE \cite{Luo2024LaRE, chu2025fire}, has generalized well across different generative models, while maintaining a transparent interpretation. %. By computing the reconstruction discrepancy $|x-x'|$ or its latent-space analog, these methods achieved empirical success cross-architecture performance while maintaining a transparent interpretation.

%Despite extensive research in the literature, reliably detecting AI-generated images remains difficult. Existing 

% As generative models advance rapidly, and the range of generated content (e.g., portraits, medical images, satellite data) becomes increasingly diverse, we require detectors that are both architecture-agnostic and data-agnostic, capable of generalizing well across models and data distributions.

%Among the growing number of detection paradigms, 
\begin{figure}
  \centering
  \includegraphics[width=1\linewidth]{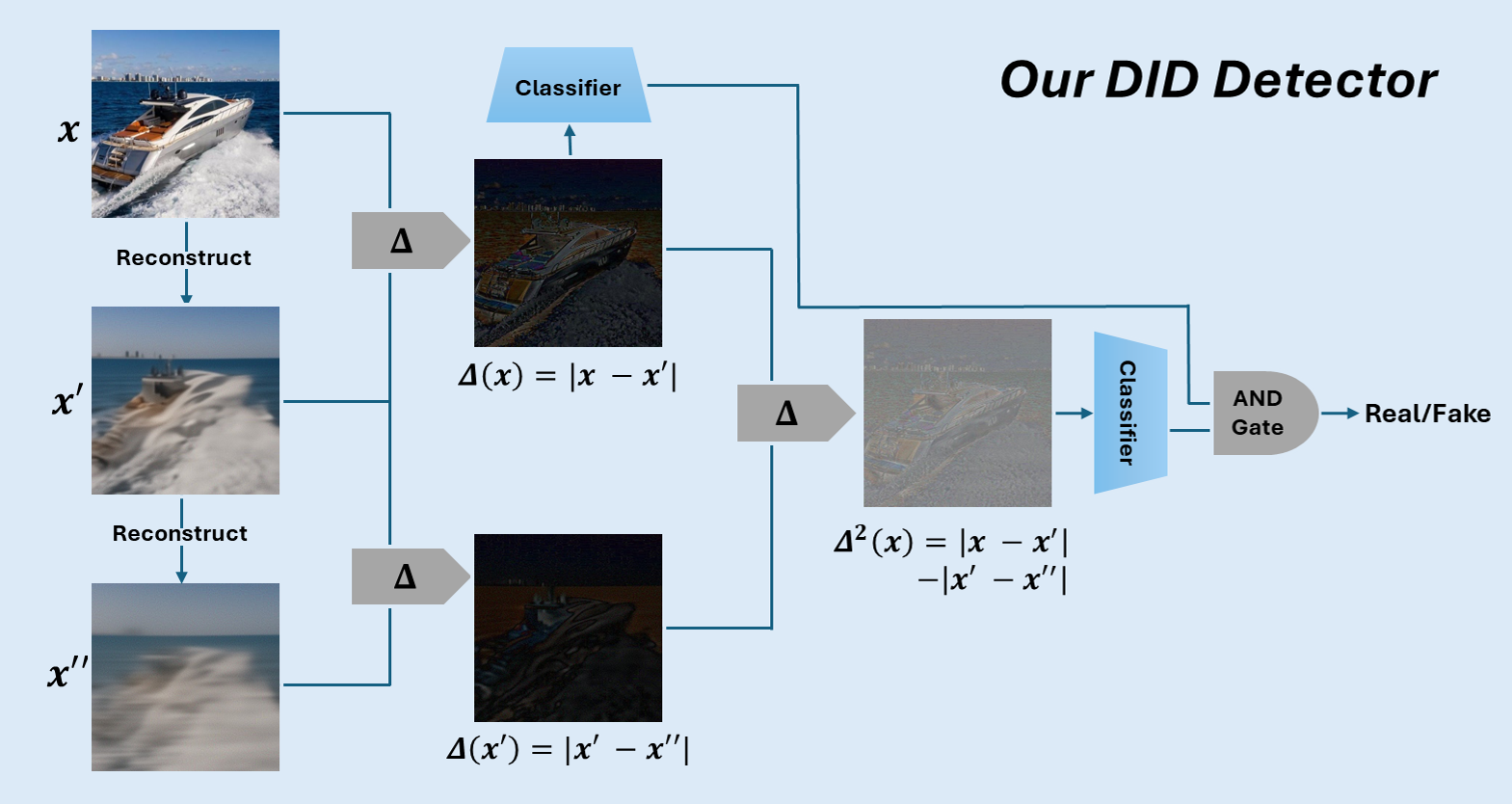}
  \caption{Visualizations of our DID detector. $\Delta$ denotes the differencing operator that computes the pixel-wise difference between two images.}
  \label{fig:workflow}
\end{figure}
However, the effectiveness of existing reconstruction-based algorithms relies on a sufficiently large gap between real and AI-generated images to be detectable. As generative models advance rapidly, this assumption becomes increasingly invalid. Meanwhile, images found on the internet are often post-processed, through resizing, compression, or partial AI editing (where only specific regions are synthesized), which further confounds detection signals. Recent studies have empirically observed that as the quality of the image generator improves or under adversarial scenarios (e.g., with partial AI edits), reconstruction-based algorithms struggle to detect AI-generated content effectively \cite{chu2025fire, Wang2025End4}. See also, the bottom panel of Figure~\ref{fig:illustration} for an illustration. Therefore, as high-fidelity diffusion models become prevalent, it is necessary to move beyond existing reconstruction-based methods and develop more effective algorithms capable of detecting subtler differences between real and AI-generated content.

This paper proposes a new algorithm, termed \ul{\textit{difference-in-differences}} (DID), for detecting AI-generated content. Instead of relying solely on the reconstruction error -- which can be viewed as a \ul{\textit{first-order}} difference between the input image and its reconstruction -- we compute a \ul{\textit{second-order}} difference by taking the difference between the reconstruction error of the input image and the reconstruction error of its reconstructed version. Our analytical analysis in Section \ref{sec:method} shows that this second-order differencing can effectively remove fluctuations introduced during synthetic image generation, thus amplifying the detection signal particularly when it is weak. 
We next employ both first-order and second-order differences as the basis for detection (see Figure~\ref{fig:workflow} for an illustration of our proposal). 
Empirically, our extensive numerical experiments demonstrate that DID consistently outperforms existing detectors across diverse combinations of image datasets and generative models, achieving improvements around 20\% - 30\% over the strongest baseline. %(Sections \ref{sec:compare} and \ref{sec:sensitivity}). Moreover, our ablation study (Section \ref{sec:ablation}) 
Moreover, our findings shed light on the effectiveness of DID: in simpler settings where AI-generated images are sufficiently different from real ones, first-order differences alone are adequate for detection; however, in more complex experimental settings, employing the second-order difference substantially improves the detection accuracy. These observations are consistent with our analytical insights. By leveraging both orders of differences, our method generalizes robustly across a wide range of detection tasks.

\section{Related Works}\label{sec:relatedworks}
This section is organized as follows. We begin by reviewing various diffusion models used for image generation. We then review existing algorithms for AI-image detection. Finally, we introduce the difference-in-difference (DID) method from econometrics, which is conceptually related to our proposal but fundamentally different in terms of its problem setting and application.  
\subsection{Diffusion Models}
Diffusion models have become a cornerstone of modern image generation and multimodal large language models  \citep[MLLMs,][]{huang2023language, yin2024survey}, powering a wide range of tasks such as text-to-image, text-to-3D generation, and text-to-video generation~\cite{nichol2021glide, crowson2022vqgan, gu2022vector, ramesh2022hierarchical, saharia2022photorealistic, podell2023sdxl, ruiz2023dreambooth, zhang2024motiondiffuse, yang2025mmada, lin2023magic3d,openai_sora_world_simulators_2024,ho2022imagen}. They constitute an important family of probabilistic generative models that progressively perturb data with noise and learn a reverse process to generate synthetic samples~\cite{yang2023diffusion}. Broadly speaking, these models can be categorized into three main types: denoising diffusion probabilistic models~\cite{sohl2015deep,ho2020denoising}, score-based generative models~\cite{song2019generative, song2020improved}, and stochastic differential equation (SDE)-based models ~\cite{song2020score, song2021maximum}. Building on the three types of models described above, each can be further decomposed into two subtypes: unconditional and conditional diffusion models. For unconditional diffusion models, prior research has focused on three directions: (i) accelerating the sampling process~\citep{song2019generative, watson2021learning, dockhorn2022genie, karras2022elucidating, zhang2022fast, zhang2022gddim}, (ii) tightening the variational lower bound on the data log-likelihood~\citep{song2020denoising, kingma2021variational, nichol2021improved, bao2022analytic, lu2022maximum}, and (iii) extending the model to accommodate new data modalities~\citep{niu2020permutation, vahdat2021score, Bortoli2022RiemannianSG, huang2022riemannian, jo2022score}. By contrast, research on conditional diffusion models has focused on guiding the generation process to produce content that aligns with users'  desired specifications. For instance, ADM uses an auxiliary classifier to improve the quality of generated samples~\cite{dhariwal2021diffusion}.  ImageReward~\cite{xu2023imagereward} and Diffusion-DPO~\cite{wallace2024diffusion} leverage reinforcement learning from human feedback to align diffusion models with human preferences. Scalable diffusion models~\cite{peebles2023scalable} replace standard layer normalization in transformers with adaptive layer normalization to more effectively encode conditioning variables. DiffuSeq~\cite{gong2022diffuseq}, unCLIP~\cite{ramesh2022hierarchical}, ConPreDif~\cite{yang2023improving}, and RPG~\cite{yang2024mastering} focus on guiding diffusion models with textual inputs.

\subsection{AI-Generated Image Detection}\label{sec:imagedetectreview}
With the rapid advancement of AI image generators, there is an urgent need to detect AI-generated images; see  \cite{lin2024detecting} for a recent review of detection algorithms. 

Early works leverage physical~\cite{mccloskey2018detecting,mccloskey2019detecting, farid2022perspective,borji2023qualitative} and physiological features~\cite{ciftci2020fakecatcher} in the images, such as inconsistencies in color, perspective, shadows, reflections or human anatomy for detection. These methods rely on the assumption that AI-generated images often fail to adhere to real-world physical and physiological constraints. Later approaches employ spatial-domain features, such as the image texture to build classifiers \citep{lorenz2023detecting,nguyen2023unmasking,zhong2023patchcraft}.  %~\cite{zhong2023patchcraft}, gradients~\cite{nguyen2023unmasking}, and local intrinsic dimensionality~\cite{lorenz2023detecting} as features. 
However, such classifiers often fail to capture frequency-domain artifacts~\cite{wolter2022wavelet}. To overcome this limitation, several methods have been proposed that combine spatial- and frequency-domain features for detection \cite{wolter2022wavelet,bammey2023synthbuster,xi2023ai,lanzino2024faster}. Other approaches leverage alternative features, such as the negative log-likelihood of the input image \cite{cozzolino2024zero}, or directly fine-tune an MLLM \cite{lin2025seeing} for detection. 

Different from the aforementioned approaches, our proposal is most closely related to reconstruction-based detection algorithms. As discussed in the introduction, a representative example is DIRE~\cite{Wang2023DIRE}, which employs a diffusion model with a forward diffusion process and a reverse denoising process to obtain the reconstruction, computes the difference from the original image as the reconstruction error, and trains the classifier using this reconstruction error. SeDID~\cite{ma2023exposing} extends this approach by considering not only the final reconstruction error between the reconstructed and input image, but also the intermediate errors accumulated along the forward diffusion and reverse denoising processes. Other variants, such as DRCT~\cite{chen2024drct}, modify the objective function by employing a contrastive loss to train the classifier based on the reconstruction error. One potential limitation of DIRE is its computational time required to obtain the reconstruction, as the forward diffusion and reverse denoising processes involve multiple steps and can be time-consuming. To address this, LaRE$^2$~\cite{Luo2024LaRE} projects the image into a latent space and applies the forward and reverse procedures only once, computing the reconstruction error in the latent space to speed up the computation. AEROBLADE~\cite{ricker2024aeroblade} adopts a training-free strategy: instead of training a classifier with the reconstruction error, it relies on simple statistical measures derived from the reconstruction error for detection. FakeInversion~\cite{cazenavette2024fakeinversion} inverts an open-source stable diffusion process model to compute the reconstruction error. 
Finally, FIRE additionally leverages frequency-domain information to compute reconstruction errors, which enhances the detector’s robustness in adversarial settings and across different generative models~\cite{chu2025fire}.

%These kinds of detectors increasingly leverage diffusion models’ internal mechanics via reconstruction signals. 
%DIRE~\cite{Wang2023DIRE} measures DDIM~\cite{song2020denoising} inversion error to separate real from generated images, and 
%SeDID~\cite{ma2023exposing} extends this idea with multi-step error accumulation along the denoising trajectory. 
%To improve robustness and data efficiency, DRCT~\cite{chen2024drct} introduces a reconstruction-contrastive objective to improve cross-set performance, and 
%LaRE$^2$~\cite{Luo2024LaRE} uses single-step latent reconstruction error with an error-guided refinement module to speed up the feature extraction. FakeInversion~\cite{cazenavette2024fakeinversion} uses inversion features from a pre-trained Stable Diffusion to detect images from unseen generators using text-conditioned inversion. %Moreover, FIRE~\cite{chu2025fire} further guides reconstruction by frequency cues, yielding spectrum-aware errors that are more robust to diverse perturbations and unseen diffusion models.

\subsection{Difference-in-Differences}\label{sec:didreview}
Our proposal is conceptually similar to the difference-in-difference approach for causal inference, which estimates the treatment effect of a particular policy. DID originated in social science research on quasi-experimental designs \cite{hyman1982quasi,campbell2015experimental} and was later systematized and popularized in the econometrics literature for analyzing panel data with multiple subjects observed over time \cite{ashenfelter1984using, card1993minimum, blundell2009alternative}. 

In the typical DID setting, some subjects are continuously exposed to a control policy, while others switch from the control to a new policy at a certain time point. For the latter group, one can compute a first-order difference in their average outcomes before and after the policy change. However, this difference captures not only the effect of the policy but also any time-related effects. To remove the confounding effect of time, a corresponding first-order difference is computed for the control group, and a second-order difference between the two first-order differences is then taken to isolate the treatment effect. For more details, refer to \cite{angrist2009mostly,callaway2021difference,goodman2021difference,sun2021estimating}. 

Despite sharing the same name and similar conceptual ideas, our proposal is designed for a fundamentally different application of AI-generated image detection, as opposed to causal inference.

%Difference-in-Differences (DID) originates in the broader social-science and statistics literature on quasi-experimental designs \cite{campbell2015experimental, hyman1982quasi}. It formalizes “before-and-after with a control group” comparisons by using untreated units to proxy the counterfactual for treated units. Econometrics later systematized and popularized DID for panel and repeated cross-section data \cite{ashenfelter1984using, card1993minimum, blundell2009alternative}. The central idea is to compare the pre-post change in outcomes for the treated group with the contemporaneous change for the control group, thereby isolating the policy’s causal effect and mitigating confounding from time-varying shocks and selection bias \cite{angrist2009mostly}. In practice, standard implementations employ two-way fixed effects (unit and time) and event-study designs to interrogate the parallel-trends assumption and to trace dynamic treatment effects over time\cite{callaway2021difference,goodman2021difference,sun2021estimating}.
\section{Methodology}\label{sec:method}

This section first provides background on diffusion models (Section \ref{sec:back}), then discusses the intuition and limitations of existing reconstruction-based methods (Section \ref{sec:intui}), which motivates our use of DID (Section \ref{sec:DID}). Finally, we detail our proposal (Section \ref{sec:ourproposal}).

\subsection{Background on Diffusion Models}\label{sec:back}
Image generation via diffusion models typically consists of two processes: a \textit{forward diffusion} process that gradually corrupts data with Gaussian noise, and a \textit{reverse denoising} process that reconstructs the underlying image signal from noise. Specifically, in the forward process, an image $x_0$ is progressively transformed into a sequence of noisy images $\{x_t\}_{t=1}^{T}$ according to
\begin{equation*}
x_t = \sqrt{\alpha_t}x_{t-1} + \sqrt{1-\bar{\alpha}_t}\,\epsilon_t,
\end{equation*}
for $t = 1, \dots, T$, where $x_t$ denotes the noisy image at timestep $t$, and $\bar{\alpha}_t = \prod_{i=1}^{t}\alpha_i$ for a sequence of pre-defined weights $\{\alpha_t\}_t$ that controls the signal-to-noise ratio and $\epsilon_t \sim \mathcal{N}(0, I)$ is the Gaussian noise.  

In contrast, during the reverse process, the model learns to iteratively denoise $x_t$ back to a clean sample using a parameterized neural network $\epsilon_\theta(x_t, t)$ that takes $x_t$ and $t$ as input and aims to predict the added noise $\epsilon_t$. This process can be mathematically represented by
\begin{equation*}
x_{t-1} = \frac{1}{\sqrt{\alpha_t}}\left(x_t - \frac{1-\alpha_t}{\sqrt{1-\bar{\alpha}_t}}\,\epsilon_\theta(x_t, t)\right)
+ \sqrt{1-\alpha_{t-1}}\,\epsilon_t,
\end{equation*}
for $t = T, \dots, 1$, where $\epsilon_\theta(x_t, t)$ is trained to minimize the prediction error between the true and the predicted noise.

\subsection{Intuition and Limitations in Reconstruction-Based Detection}\label{sec:intui}

Let $\mathcal{X} \subseteq \mathbb{R}^{3\times256\times256}$ denote the space of all natural images, and $\mathcal{M}$ represent the manifold of images generated by a given pre-trained diffusion model. 
In general, $\mathcal{M}$ does not coincide with $\mathcal{X}$, as the model’s learned distribution cannot perfectly align with the true image distribution. However, as generative models become increasingly expressive, $\mathcal{M}$ tends to approximate $\mathcal{X}$ more closely, forming a high-fidelity submanifold near the real image space.

The reconstruction can thus be interpreted as a projection of an input image $x$ onto the generative manifold $\mathcal{M}$, followed by a stochastic perturbation\footnote{Similar interpretations have been proposed in the context of LLM-generated text detection \citep{zhou2026learn}.}:
\begin{equation}\label{eq:reconstruction-operator}
    \mathcal{R}(x) = \Pi_{\mathcal{M}}(x) + \delta(x),
\end{equation}
where $\mathcal{R}(\bullet)$ denotes the reconstruction operator, $\Pi_{\mathcal{M}}(x)$ denotes the projection of $x$ onto $\mathcal{M}$, and $\delta(x)$ is a perturbation term capturing the randomness introduced during the reconstruction.  

Consequently, for a synthetic image $x \in \mathcal{M}$, its projection onto $\mathcal{M}$ equals $x$ itself, resulting in a reconstruction error of
\begin{equation}\label{eqn:fake}
    \Delta_{\text{fake}}(x) = |\delta(x)|,
\end{equation}
where the absolute value is taken elementwise. In contrast, for a real image $x \in \mathcal{X}$, its reconstruction error is given by
\begin{equation}\label{eqn:real}
    \Delta_{\text{real}}(x) = |x - \Pi_{\mathcal{M}}(x) - \delta(x)|.
\end{equation}

When the generative model is weak, the discrepancy between $\mathcal{M}$ and $\mathcal{X}$ is large, so $|x - \Pi_{\mathcal{M}}(x)|$ dominates the reconstruction error. As a result, the real image achieves a much larger reconstruction error than the synthetic image (see the upper panel of Figure \ref{fig:illustration} for an illustration), and we classify an image as synthetic if its reconstruction error is small. This explains the rationale behind existing reconstruction-based methods. When the generative model is strong, however, the real and synthetic image distributions are closely aligned, so the signal $|x - \Pi_{\mathcal{M}}(x)|$ becomes small. Consequently, the reconstruction errors of both real and synthetic images are dominated by the perturbation and become similar (see the bottom panel of Figure \ref{fig:illustration}), and existing reconstruction-based methods struggle to distinguish real from synthetic images.

%However, as capacity of advanced models increases, $\mathcal{M}$ becomes a dense subspace of $\mathcal{X}$, and thus $|x - \Pi_{\mathcal{M}}(x)|$ becomes uniformly small for almost all $x\in\mathcal{X}$. In this case, the reconstruction errors $\Delta_{\text{real}}(x)$ and $\Delta_{\text{fake}}(x)$ are both primarily dominated by the stochastic term $\delta_x$, whose magnitude and distribution may idiosyncratically vary across samples and thereby overshadowing the geometric distinction between $\mathcal{M}$ and $\mathcal{X}$. This fundamentally limits the discriminative ability of traditional reconstruction-error-based detectors.

%Nevertheless, although the geometric discrepancy may vanish, the statistical behavior of the perturbation term $\delta_x$ may still differ between real and synthetic images after removing sample-specific effects. This observation naturally motivates the exploration of more subtle, higher-order statistical indicators — beyond first-order reconstruction errors — to capture residual structural discrepancies that remain informative when mean errors converge.

%Motivated by this, 
In summary, existing reconstruction-based methods are limited in settings where synthetic images closely resemble real ones, due to the strong capacity of modern generative models. This indicates that relying solely on first-order differences is insufficient in such challenging scenarios, which motivates our difference-in-difference approach that seeks higher-order differences to remove perturbation errors while amplifying the weak signal. We will demonstrate how DID addresses this challenge in the next section.

%we introduce a \emph{Difference-In-Difference (DID) detection framework} that characterizes the statistical consistency of reconstruction variance as an additional discriminative indicator. The central observation is that the distributional variability of reconstruction errors differs between real and synthetic images. Specifically, when the model reconstructs its own outputs, the reconstruction variance tends to be small, whereas real images, which do not strictly follow the model manifold, exhibit higher instability across repeated reconstruction procedures. 

\subsection{Difference-In-Differences}\label{sec:DID}
%Motivated by this, we introduce a \emph{Difference-In-Difference (DID)} detection framework that characterizes the statistical consistency of reconstruction variance as an additional discriminative indicator. The key insight is that the distributional variability of reconstruction errors differs between real and synthetic images. When the model reconstructs its own outputs, the reconstruction process is stable with low variance, whereas real images often exhibit larger fluctuations across repeated reconstructions.
Unlike existing methods which conduct reconstruction only once, DID performs two consecutive reconstructions using the same pre-trained diffusion model:
\begin{equation*}
    x'=\mathcal{R}(x), \quad x''=\mathcal{R}(x'),
\end{equation*}
where $\mathcal{R}(\bullet)$ is the reconstruction operator defined in~\eqref{eq:reconstruction-operator}, $x'$ denotes the reconstruction of the input image $x$ and $x''$ denotes the reconstruction of the reconstructed image. 

Taking the difference between $x'$ and $x$ yields the first-order reconstruction error, $\Delta(x) = |x - \mathcal{R}(x)|$, which forms the basis of existing reconstruction-based detectors for classification. However, as discussed in the previous section, relying solely on the first-order reconstruction error is limited when the signal $|x - \Pi_{\mathcal{M}}(x)|$ for a real image is small, causing the reconstruction error to be dominated by the perturbation $\delta(x)$. To address this limitation, DID computes the second-order difference between the two reconstruction errors:
\begin{equation}
    \Delta^{2}(x)=|x-x'|-|x'-x''|.
\end{equation}
To illustrate the intuition behind DID, note that, similar to \eqref{eqn:fake} and \eqref{eqn:real}, one can derive that for a synthetic image $x$, the second-order reconstruction error is given by
\begin{equation}\label{eqn:secondorderfake}
    \Delta_{\text{fake}}^{2}(x) = |\delta(x)| - |\delta(x')|,
\end{equation}
whereas for a real image, it is
\begin{equation}\label{eqn:secondorderreal}
    \Delta_{\text{real}}^{2}(x) = |x - \Pi_{\mathcal{M}}(x) - \delta(x)| - |\delta(x')|.
\end{equation}

When the signal is weak, $x$ is close to $x'$. In this case, if the perturbation error $\delta$ is highly correlated across space, then $\delta(x) \approx \delta(x')$. As a result, the two perturbations cancel out, and \eqref{eqn:secondorderfake} and \eqref{eqn:secondorderreal} simplify to
\begin{equation}\label{eqn:secondorderfake2}
    \Delta_{\text{fake}}^{2}(x) \approx 0,
\end{equation}
and
\begin{equation}\label{eqn:secondorderreal2}
    |\Delta_{\text{real}}^{2}(x)| = |x - \Pi_{\mathcal{M}}(x) - \delta(x) + \delta(x')| \approx |x - \Pi_{\mathcal{M}}(x)|,
\end{equation}
where the first equality holds because the weak signal makes the magnitude of $|x - \Pi_{\mathcal{M}}(x)|$ much smaller than that of $|\delta(x)|$, so that the sign of $x - \Pi_{\mathcal{M}}(x) - \delta(x)$ is dominated by that of $-\delta(x)$.

Comparing \eqref{eqn:secondorderfake2} and \eqref{eqn:secondorderreal2} with \eqref{eqn:fake} and \eqref{eqn:real}, it becomes clear that, unlike the first-order difference, the second-order difference is no longer confounded by the perturbation error and depends only on the signal, even when it is weak. This highlights the advantage of DID over existing reconstruction-based algorithms: by taking a second-order difference, it effectively removes the perturbation noise and uncovers the subtle signal.

\subsection{Our Proposal}\label{sec:ourproposal}
%Although the variance-based detector captures more subtle statistical discrepancies, repeated reconstruction introduces the risk of error accumulation, which may reduce its reliability when the first-order reconstruction signal remains discriminative. To integrate both types of information, we propose a \textbf{Dual-Classifier Framework} that jointly utilizes mean- and variance-based representations.
We present the details of our proposed method in this section. Based on the aforementioned analysis, the first-order reconstruction error $\Delta(x) = |x - x'|$ is effective for distinguishing real from synthetic images when their discrepancy is large, whereas the second-order reconstruction error $\Delta^2(x) = |x - x'| - |x' - x''|$ remains reliable when real and synthetic images are highly similar. To handle both cases simultaneously, our detector integrates both first- and second-order reconstruction errors for robust detection
(Figure \ref{fig:diff-maps}); more visualizations are reported in the Appendix Section \ref{sup:visualization}.

Specifically, we train two classifiers independently based on $\Delta(x)$ and $\Delta^{2}(x)$, respectively, using the standard cross-entropy loss:
\begin{equation}
\mathcal{L} = - \frac{1}{N} \sum_{i=1}^{N} \big[y_i \log(y'_i) + (1 - y_i) \log(1 - y'_i)\big],
\end{equation}
where $y_i$ denotes the ground-truth label indicating whether the image is real or synthetic, $y'_i$ is the predicted probability produced by the classifier, and $N$ is the number of training samples.

Each classifier outputs a numerical probability score and predicts an image as real if the score is smaller than a predefined threshold $c$. Finally, an image is classified as real only when both classifiers identify it as real.
\section{Experiments}

\begin{table*}[t]
\caption{Accuracy (\%) of different detectors when trained on \textbf{ImageNet \& ADM}. The best results are highlighted in \textbf{bold}, while the second best results are \underline{underlined}. All averages are computed from the raw values and then rounded to two decimals.} %The last column reports the improvement of DID over the second-best baseline.}
\label{tab:imagenet_adm_full}
\centering
{\setlength{\tabcolsep}{15pt} % ← 这里只对这一张表生效 
% \resizebox{0.7\linewidth}{!}{%
\begin{tabular}{ll| c c c c c}
\toprule
\multicolumn{2}{c|}{\textbf{Train set \& Gen. model}} &
\multicolumn{5}{c}{\textbf{ImageNet \& ADM}} \\
\cmidrule(lr){1-2}\cmidrule(lr){3-7}
\textbf{Eval set} &
\textbf{Gen. model} &
\textbf{DID} &
\textbf{DIRE} &
\textbf{LaRE$^2$} &
\textbf{AERO.} &
\textbf{UFD} \\
\midrule
% ===== LSUN-B =====
\multirow{6}{*}{LSUN-B}
 & ADM            & \textbf{99.5} & \textbf{99.5} & 50.0 & 52.3 & \underline{63.9} \\
 & PNDM           & \textbf{97.6} & \textbf{97.6} & 55.4 & 49.5 & \underline{91.8} \\
 & DDPM           & \underline{99.5} & \textbf{99.8} & 56.6 & 56.7 & 76.5 \\
 & iDDPM          & \underline{99.4} & \textbf{99.6} & 50.0 & 50.3 & 85.1 \\
 & SDv1           & 99.7 & \underline{99.9} & \textbf{100} & 50.6 & 94.5 \\
 \cmidrule(lr){2-7}
 & Avg.           & \underline{99.12} & \textbf{99.25} & 62.39 & 51.87 & 82.36 \\
\midrule
% ===== ImageNet =====
\multirow{3}{*}{ImageNet}
 & ADM            & \textbf{100} & \textbf{100} & \textbf{100} & 51.0 & \underline{85.8} \\
 & SDv1           & \textbf{99.3} & 98.9 & \underline{99.2} & 37.0 & 79.9 \\
 \cmidrule(lr){2-7}
 & Avg.           & \textbf{99.64} & 99.44 & \underline{99.60} & 43.99 & 82.80 \\
\midrule
% ===== LAION =====
\multirow{6}{*}{LAION}
 & Kan.3          & \textbf{99.5} & \underline{99.3} & 64.5 & 66.7 & 72.6 \\
 & SDXL           & \textbf{99.6} & \underline{99.4} & 66.4 & 55.5 & 69.5 \\
 & Vega           & \textbf{99.7} & \underline{99.5} & 66.4 & 52.2 & 68.6 \\
 & Playground v2.5& \textbf{99.8} & \textbf{99.8} & 66.0 & 53.1 & \underline{82.8} \\
 & Stable Cascade & \textbf{99.6} & \underline{99.4} & 66.3 & 73.4 & 81.4 \\
 \cmidrule(lr){2-7}
 & Avg.           & \textbf{99.60} & \underline{99.45} & 65.89 & 60.14 & 75.95 \\
\midrule
\multicolumn{2}{l|}{Total average} &
\textbf{99.41} & \underline{99.37} & 70.05 & 54.00 & 79.35 \\
\bottomrule
\end{tabular}
}
% }%
\end{table*}

\begin{figure}[t]
    \centering
    \setlength{\tabcolsep}{2pt}
    \renewcommand{\arraystretch}{0}

\begin{tabular}{c@{\hspace{2pt}}cccc}

        % ===== row 1: real =====
        \raisebox{0.04\textwidth}{\rotatebox[origin=c]{90}{\footnotesize Real}} &
        \includegraphics[width=0.1\textwidth]{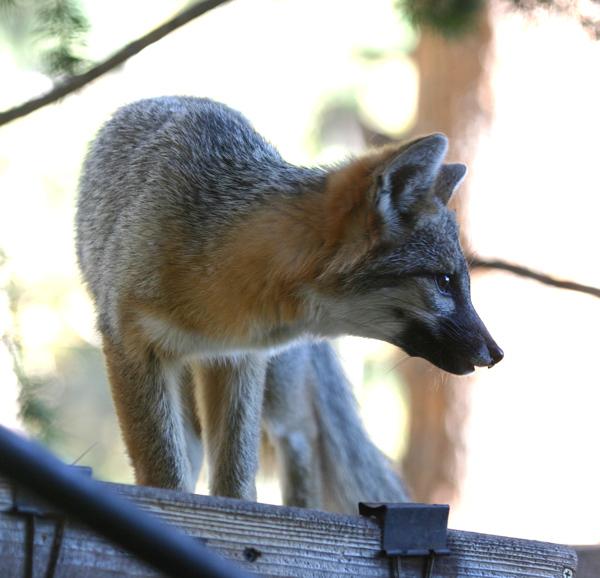} &
        \includegraphics[width=0.1\textwidth]{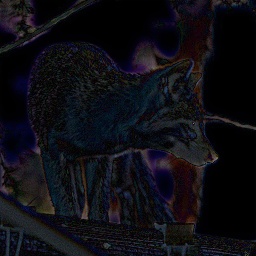} &
        \includegraphics[width=0.1\textwidth]{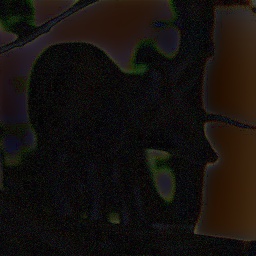} &
        \includegraphics[width=0.1\textwidth]{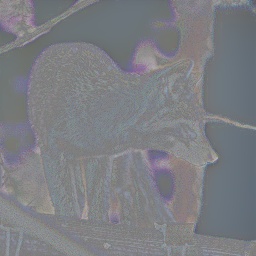} \\[3pt]
        
        % ===== row 2: Kandinsky3 =====
        \raisebox{0.04\textwidth}{\rotatebox[origin=c]{90}{\footnotesize kandinsky3}} &
        \includegraphics[width=0.1\textwidth]{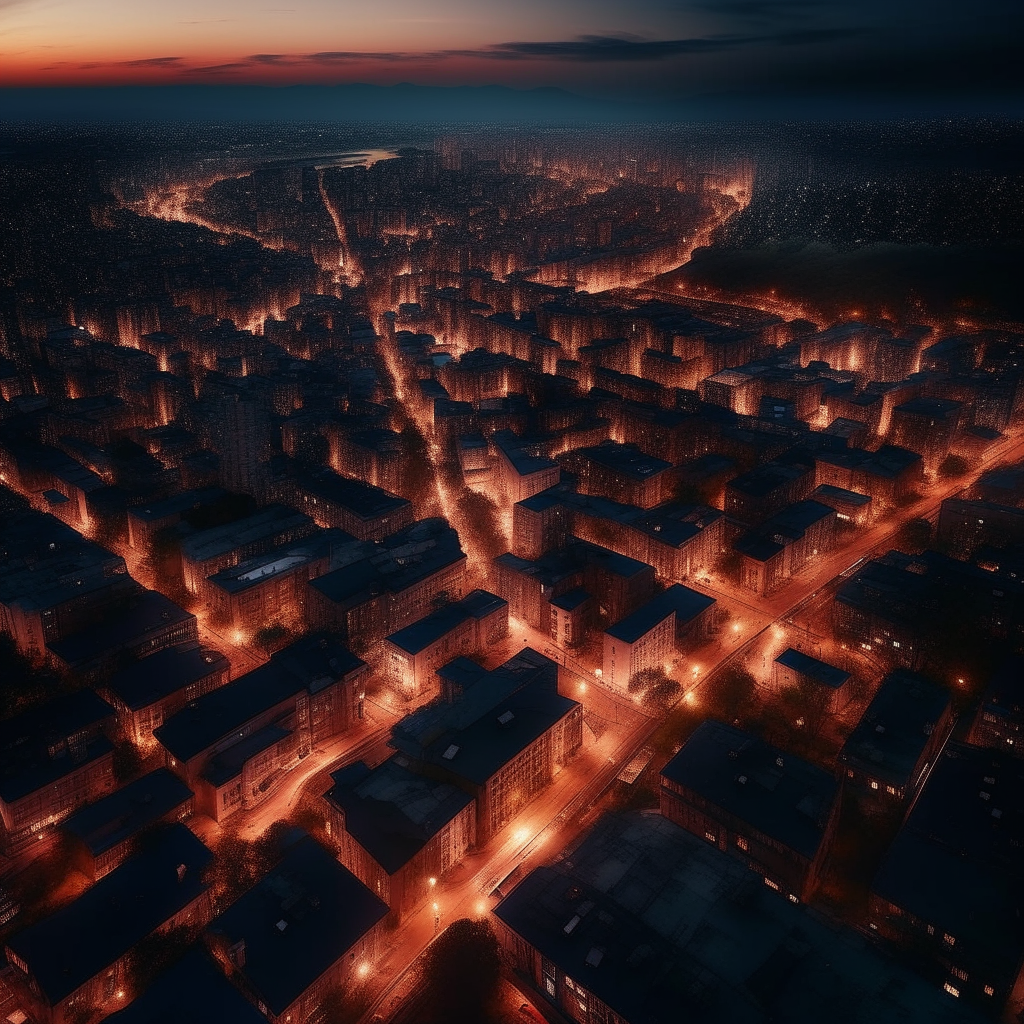} &
        \includegraphics[width=0.1\textwidth]{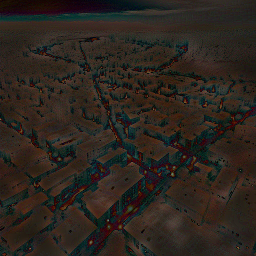} &
        \includegraphics[width=0.1\textwidth]{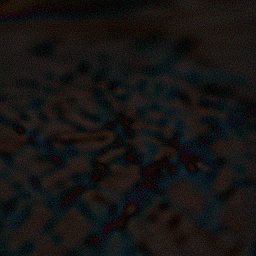} &
        \includegraphics[width=0.1\textwidth]{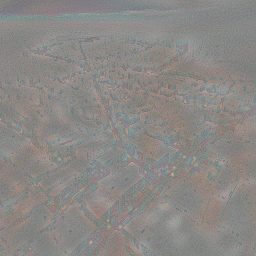} \\
        [3pt]
         % ===== row 3: SDXL =====
        \raisebox{0.04\textwidth}{\rotatebox[origin=c]{90}{\footnotesize SDXL}} &
        \includegraphics[width=0.1\textwidth]{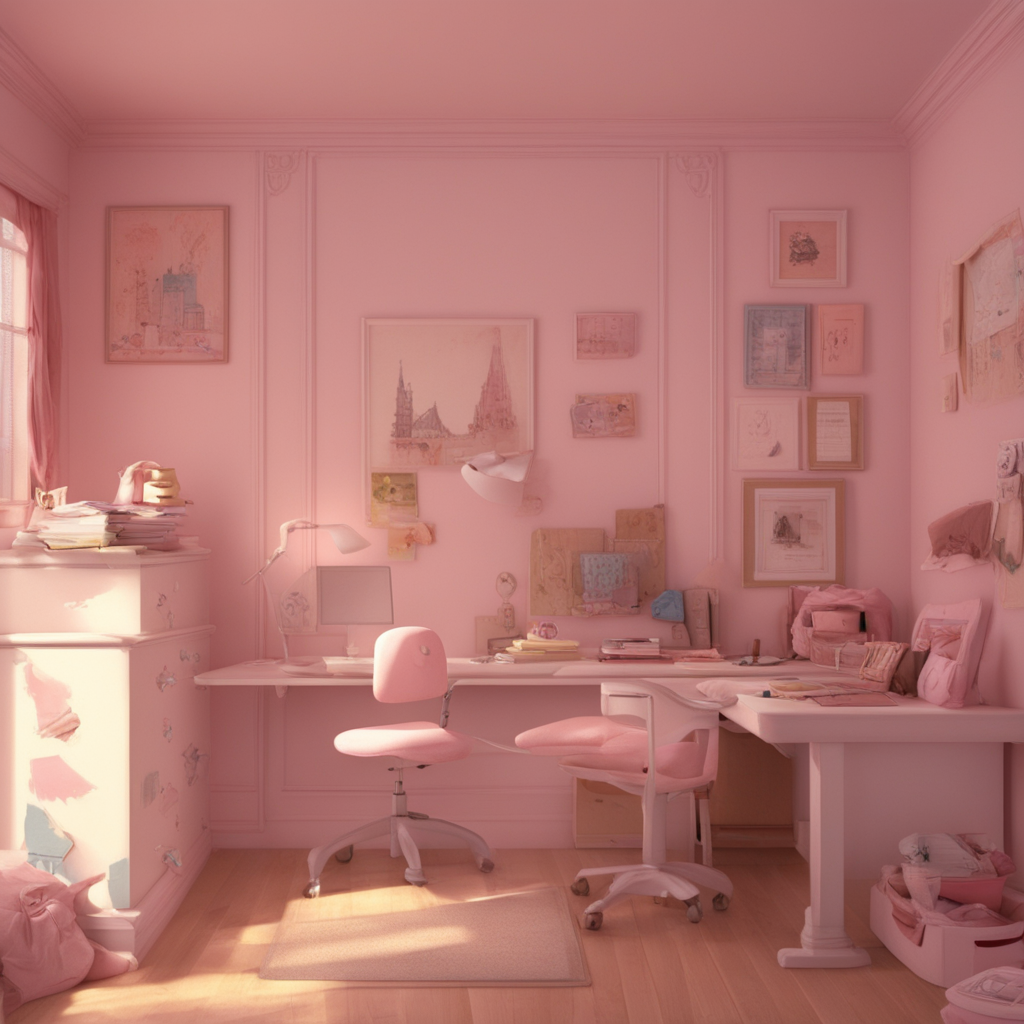} &
        \includegraphics[width=0.1\textwidth]{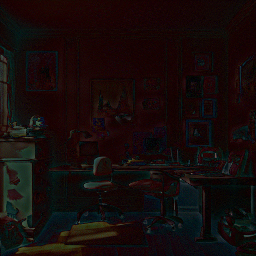} &
        \includegraphics[width=0.1\textwidth]{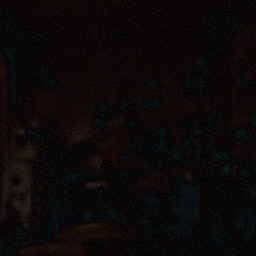} &
        \includegraphics[width=0.1\textwidth]{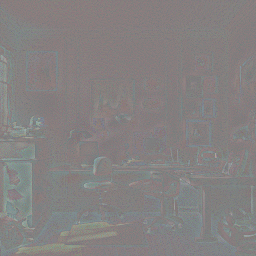} \\[3pt]
        
          % ===== row 4: Playground v2.5 =====
        \raisebox{0.04\textwidth}{\rotatebox[origin=c]{90}{\footnotesize playground-25}} &
        \includegraphics[width=0.1\textwidth]{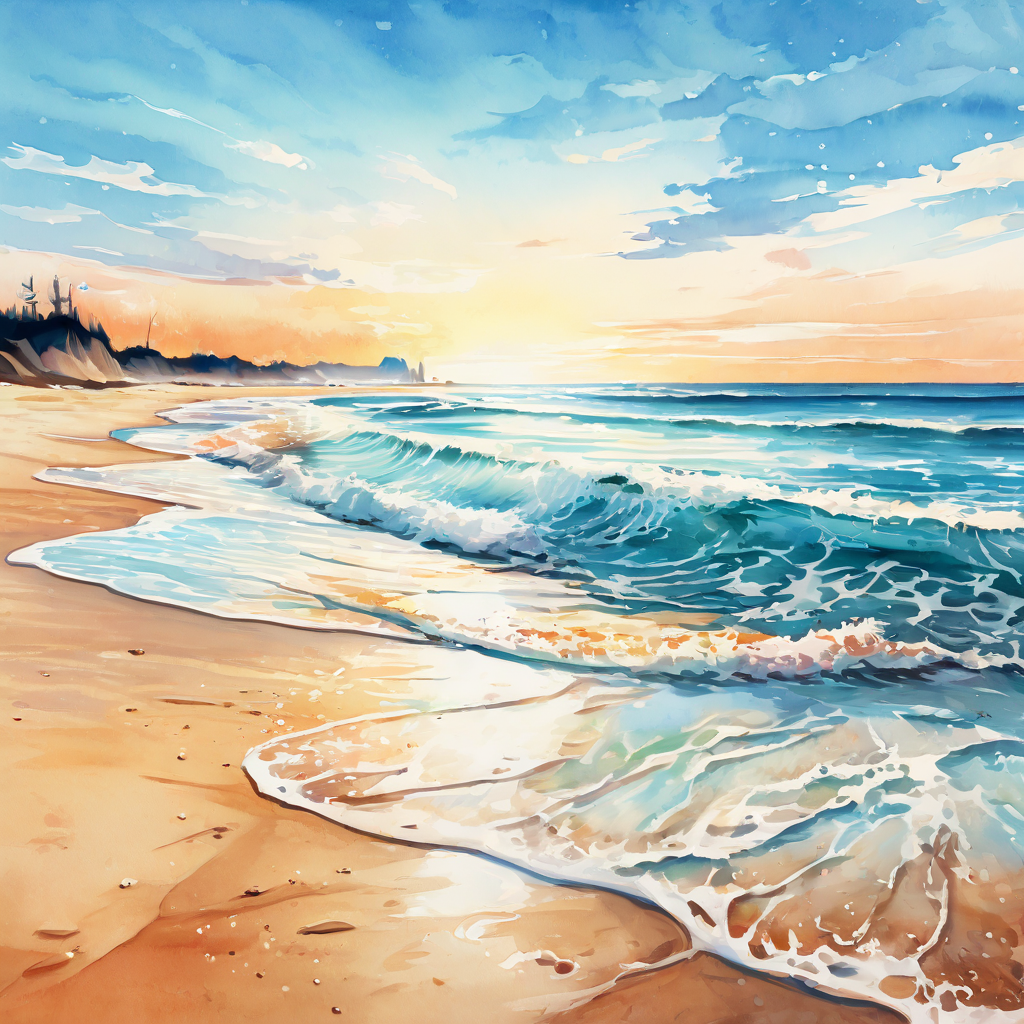} &
        \includegraphics[width=0.1\textwidth]{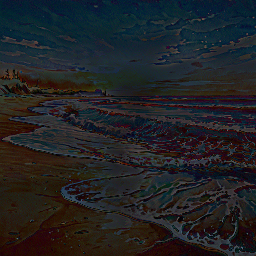} &
        \includegraphics[width=0.1\textwidth]{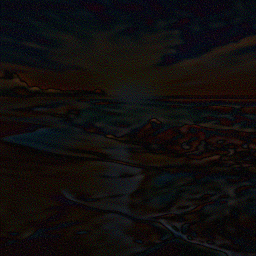} &
        \includegraphics[width=0.1\textwidth]{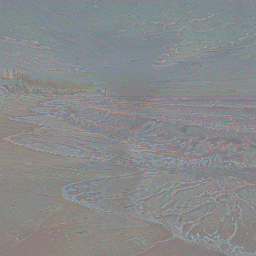} \\[3pt]

        % ===== row 5: LSUN-B w. ADM =====
        \raisebox{0.04\textwidth}{\rotatebox[origin=c]{90}{\footnotesize \shortstack{LSUN-B \\w.\ ADM}}} &
        \includegraphics[width=0.1\textwidth]{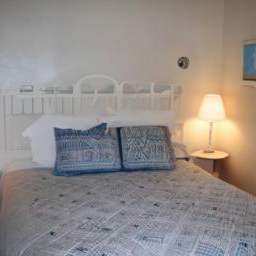} &
        \includegraphics[width=0.1\textwidth]{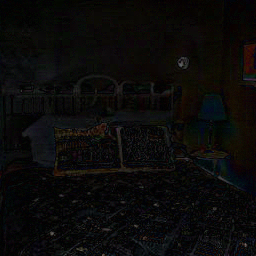} &
        \includegraphics[width=0.1\textwidth]{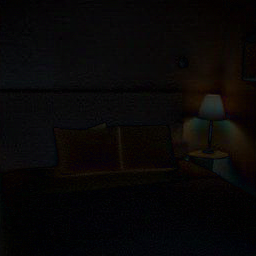} &
        \includegraphics[width=0.1\textwidth]{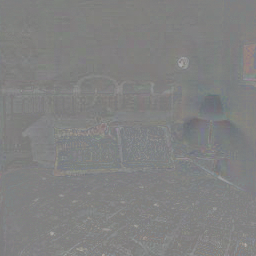} \\[3pt]

        % ===== row 6: ImageNet w. SDv1 =====
        \raisebox{0.04\textwidth}{\rotatebox[origin=c]{90}{\footnotesize \shortstack{ImageNet\\w.\ SDv1}}} &
        \includegraphics[width=0.1\textwidth]{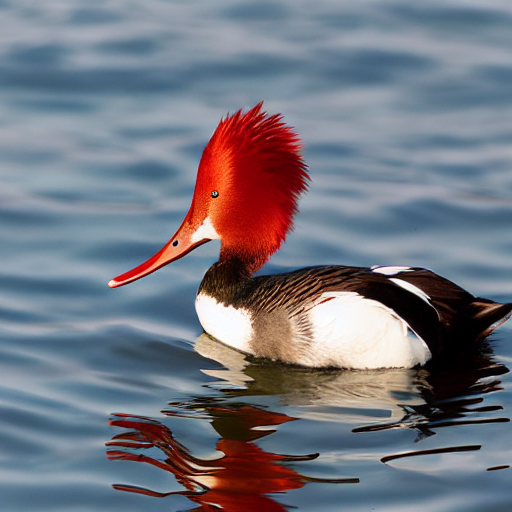} &
        \includegraphics[width=0.1\textwidth]{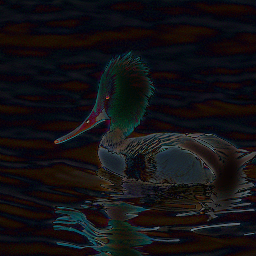} &
        \includegraphics[width=0.1\textwidth]{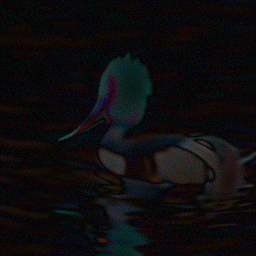} &
        \includegraphics[width=0.1\textwidth]{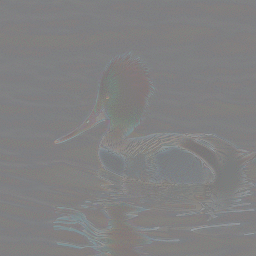} \\[3pt]
        & (a) $x$ 
        & (b) $\Delta(x)$ 
        & (c) $\Delta(x')$ 
        & (d) $\Delta^2(x)$ \\

    \end{tabular}

    \caption{Visualizations of (a) input image $x$; (b) its first-order reconstruction error $\Delta(x)$; (c) the reconstruction error of the reconstructed image $x'$; and (d) the second-order reconstruction error $\Delta^2(x)$. Compared to the first-order error $\Delta(x)$, the second-order error $\Delta^{2}(x)$ better distinguishes real from fake images: real images produce more stable and brighter responses, while fake images exhibit noticeably weaker signals, resulting in a larger gap between the two categories.}
    \label{fig:diff-maps}
\end{figure}

In this section, we conduct extensive numerical experiments to investigate the finite-sample performance of the proposed detector. We begin by describing the experimental setting in Section \ref{sec:experimentsetup}. We next report the results of our detector and compare its performance against existing state-of-the-art detectors in Section \ref{sec:compare} and Appendix Section \ref{sup:new_baseline}. Finally, we conduct a sensitivity analysis and an ablation study to examine the robustness of our detector and the contribution of the second-order difference in Sections \ref{sec:sensitivity} and \ref{sec:ablation}, respectively. 
\subsection{Experimental Setup}\label{sec:experimentsetup}
%\textbf{Datasets and evaluation metric.} 
%To evaluate the effectiveness of our proposed detection method, we use two datasets 
%under realistic conditions, we construct a self-collected dataset using several recent diffusion models. For real images, we randomly sample 10,000 images from LAION for training and reserve an additional 1,000 images for testing. For synthetic images, we use prompts collected from Midjourney and generate 10,000 training samples with open-source text-to-image models such as Kan3 and SDXL.
To comprehensively evaluate the effectiveness of each detector, we consider a variety of datasets and generative models. Our evaluation benchmark closely follows those used in DIRE \cite{Wang2023DIRE}, FIRE \cite{chu2025fire}, and FakeInversion \cite{cazenavette2024fakeinversion}. Specifically, we use two datasets -- ImageNet \cite{deng2009imagenet} and LAION \cite{schuhmann2022laion} -- to train the classifier for each detector, with one large-scale training set (40k real and 40k synthetic images) and one smaller set (10k real and 10k synthetic images). For ImageNet, we randomly sample 40,000 real images and 40,000 synthetic images generated by ADM \cite{dhariwal2021diffusion}. For LAION, we randomly sample 10,000 real images and generate 10,000 synthetic images using open-source text-to-image models such as Kandinsky 3.0 \cite{arkhipkin2023kandinsky} and SDXL \cite{podell2023sdxl}, with prompts collected from Midjourney. 

For testing, it is crucial to evaluate each detector on unseen datasets to assess how well it generalizes. Therefore, we use three test datasets -- LAION \cite{schuhmann2022laion}, ImageNet \cite{deng2009imagenet}, and LSUN-Bedroom \cite{yu2015lsun} (which is excluded from training) -- and generate synthetic images using a more diverse set of generative models than those used in training, including ADM \cite{dhariwal2021diffusion}, PNDM \cite{liu2022pseudo}, DDPM \cite{ho2020denoising}, iDDPM \cite{nichol2021improved}, SDv1 \cite{rombach2022high},  Segmind Vega \cite{gupta2024progressive}, Playground v2.5 \cite{li2024playground}, Stable Cascade \cite{pernias2023wurstchen}.  
Further details on dataset construction and data setup are reported to the Appendix Section \ref{sup:data}.

%\noindent
%\textbf{Implementation details.} 

We use the ADM model pretrained on LSUN-Bedroom to obtain the reconstructed image. After obtaining the two reconstructions, both the first- and second-order differences are fed into a ResNet-50 classifier for detection. All input images are resized to $256 \times 256$. 
For all baseline detectors, the classification threshold is fixed at the default level $c = 0.5$. Our method, however, uses two classifiers and labels an input as real only when both classifiers predict it as real. To allow for a fair comparison, we set the threshold of each classifier in DID to $1-\sqrt{0.5}\approx 0.29$, so that its overall false positive rate matches that of using a single classifier with a threshold of 0.5. 
%the joint decision rule is equivalent to a single-classifier setting of $0.5$.
Accuracy (ACC) is reported as the primary evaluation metric. All our experiments are run on NVIDIA H800 GPUs
; the computation and memory cost is reported in the Appendix Section \ref{sup:compute_cost}.
%We report Accuracy (ACC) and Average Precision (AP) as our primary evaluation metrics.
% More details and results are included in our supplementary materials.

%\noindent
%\textbf{Implementation details.}

\begin{table*}[!t]
\caption{Accuracy (\%) of different detectors when trained on \textbf{LAION \& Kan.3} (left block) and \textbf{LAION \& SDXL} (right block). The best results are highlighted in \textbf{bold}, while the second best results are \underline{underlined}. All averages are computed from the raw values and then rounded to two decimals. The last column of each block reports the improvement of DID over the second-best baseline.}
\label{tab:laion_kan3_sdxl_full}
\centering
\setlength{\tabcolsep}{2pt}
\renewcommand{\arraystretch}{1.05}
\begin{tabular}{ll| c c c c c c c c c c c c}
\toprule
\multicolumn{2}{c|}{\textbf{Train set \& Gen. model}} &
\multicolumn{5}{c}{\textbf{LAION \& Kan.3}} &
\multirow{2}{*}{\textbf{Imp.}} &
\multicolumn{5}{c}{\textbf{LAION \& SDXL}} &
\multirow{2}{*}{\textbf{Imp.}} \\
\cmidrule(lr){1-2}\cmidrule(lr){3-7}\cmidrule(lr){9-13}
\textbf{Eval set} &
\textbf{Gen. model} &
\textbf{DID} &
\textbf{DIRE} &
\textbf{LaRE$^2$} &
\textbf{AERO.} &
\textbf{UFD} &
&
\textbf{DID} &
\textbf{DIRE} &
\textbf{LaRE$^2$} &
\textbf{AERO.} &
\textbf{UFD} & \\
\midrule
\multirow{7}{*}{LSUN-B}
 & ADM            & \textbf{90.7} & \underline{88.1} & 50.3 & 50.0 & 50.8 & 21.85\% &
                    \textbf{95.0} & \underline{94.3} & 50.2 & 50.0 & 50.3 & 13.04\%\\
 & PNDM           & \textbf{78.2} & \underline{73.5} & 71.2 & 49.8 & 51.5 & 17.55\% &
                    \textbf{81.7} & \underline{74.6} & 51.6 & 49.8 & 50.9 & 27.95\%\\
 & DDPM           & \textbf{88.7} & \underline{85.2} & 56.5 & 56.4 & 51.5 & 24.05\% &
                    \textbf{95.1} & \underline{93.3} & 57.1 & 56.4 & 50.5 & 27.73\%\\
 & iDDPM          & \textbf{80.1} & \underline{77.5} & 50.0 & 49.8 & 54.2 & 11.33\% &
                    \textbf{89.0} & \underline{86.8} & 50.0 & 49.8 & 53.3 & 16.29\%\\
 & SDv1           & \underline{98.6} & 97.8 & \textbf{99.9} & 49.8 & 49.7 & - &
                    \underline{98.4} & 98.3& \textbf{99.7} & 49.8 & 50.5 & - \\
\cmidrule(lr){2-14}
  & Avg.          & \textbf{87.25}& \underline{84.41} & 65.56 & 51.13 & 51.52 & 18.23\% &
                    \textbf{91.83} & \underline{89.44} & 61.70 & 51.13 & 51.09 & 22.58\%\\
\midrule
\multirow{3}{*}{ImageNet}
 & ADM            & \textbf{99.1} & \underline{96.7} & 89.5 & 50.7 & 58.7 & 74.03\% &
                    \textbf{99.6} & \underline{96.3} & 93.3 & 50.7 & 61.1 & 89.65\%\\
 & SDv1           & \textbf{99.2} & \underline{97.7} & 94.4 & 35.0 & 65.4 & 65.33\% &
                    \textbf{99.1} & \underline{96.1} & 95.5 & 35.0 & 67.3 & 75.73\%\\
\cmidrule(lr){2-14}
 & Avg.           & \textbf{99.16}& \underline{97.16} & 91.91 & 42.83 & 62.03 & 70.46\% &
                    \textbf{99.34} & \underline{96.22} & 94.40 & 42.83 & 64.18 & 82.47\%\\
\midrule
\multirow{6}{*}{LAION}
 & Kan.3          & \textbf{100} & \textbf{100} & \textbf{100} & 58.7 & \underline{93.7} & - &
                    \textbf{99.8} & \underline{99.3} & 97.6 & 58.7 & 92.9 & 64.29\%\\
 & SDXL           & \textbf{100} & \underline{99.9} & 99.4 & 52.5 & 87.0 & 100\% &
                    \textbf{99.9} & \textbf{99.9} & \textbf{99.9} & 52.5 & \underline{92.3} & - \\
 & Vega           & \textbf{100} & \underline{99.4} & 95.5 & 51.0 & 79.9 & 100\% &
                    \textbf{99.9} & \underline{99.8} & 99.7 & 51.0 & 83.8 & 50.00\%\\
 & Playground v2.5& \textbf{100} & \textbf{100} & \underline{97.9} & 51.1 & 93.2 & - &
                    \textbf{99.9} & \textbf{99.9} & \underline{99.4} & 51.1 & 93.3 & - \\
 & Stable Cascade & \textbf{100} & \textbf{100} & \underline{99.6} & 64.7 & 95.7 & - &
                    \textbf{99.8} & \underline{99.5} & \underline{99.5} & 64.7 & 95.4 & 54.55\%\\
\cmidrule(lr){2-14}
 & Avg.           & \textbf{100} & \underline{99.84} & 98.45 & 55.57 & 89.87 & 100\% &
                    \textbf{99.84} & \underline{99.67} & 99.20 & 55.57 & 91.50 & 51.52\%\\
\midrule
\multicolumn{2}{l|}{Total average} &
\textbf{94.55}& \underline{92.96} & 83.66 & 51.60 & 69.25 & 22.52\% &
\textbf{96.42} & \underline{94.83} & 82.78 & 51.60 & 70.11 & 30.66\%\\
\bottomrule
\end{tabular}
\end{table*}

% \begin{table*}[t]
% \centering
% \small
% \begin{tabular}{l c c c c c c}
% \toprule
% \multicolumn{7}{c}{\textbf{Train: ImageNet + adm}} \\
% \midrule
% \textbf{Test} &
% \textbf{our/0.5/256} &
% \textbf{DIRE/256} &
% \textbf{FIRE} &
% \textbf{LaRe} &
% \textbf{AEROBLADE} &
% \textbf{UFD} \\
% \midrule
% lsun+adm            & 99.4 & 99.5 & 50.0 & 50.0 & 52.3 & 63.9 \\
% lsun+pndm           & 97.5 & 97.6 & 50.0 & 55.4 & 49.5 & 91.8 \\
% lsun+ddpm           & 99.5 & 99.8 & 50.0 & 56.6 & 56.7 & 76.5 \\
% lsun+iddpm          & 99.4 & 99.6 & 50.0 & 50.0 & 50.3 & 85.1 \\
% lsun+sdv1\_new      & 99.7 & 99.9 & 50.0 & 100.0 & 50.6 & 94.5 \\
% lsun (avg)          & 99.2 & 99.3 & 50.0 & 68.7 & 57.6 & 77.8 \\
% \midrule
% imagenet+adm        & 100.0 & 100.0 & 50.0 & 100.0 & 51.0 & 85.8 \\
% imagenet+sdv1       & 99.3  & 98.9  & 50.0 & 99.2  & 37.0 & 79.9 \\
% imagenet (avg)      & 99.6  & 99.4  & 50.0 & 99.6  & 44.0 & 82.8 \\
% \midrule
% laion+kandinsky3    & 99.3  & 99.3  & 50.0 & 64.5 & 66.7 & 72.6 \\
% laion+sdxl          & 99.4  & 99.4  & 50.0 & 66.4 & 55.5 & 69.5 \\
% laion+vega          & 99.5  & 99.5  & 50.0 & 66.4 & 52.2 & 68.6 \\
% laion+playground-25 & 99.8  & 99.8  & 50.0 & 66.0 & 53.1 & 82.8 \\
% laion+stable-cascade& 99.4  & 99.4  & 50.0 & 66.3 & 73.4 & 81.4 \\
% laion (avg)         & 99.5  & 99.5  & 50.0 & 65.9 & 60.1 & 75.0 \\
% \midrule
% whole\_average      & 99.4  & 99.4  & 50.0 & 72.4 & 56.5 & 77.5 \\
% \bottomrule
% \end{tabular}
% \caption{Accuracy (\%) of different detectors when trained on \textbf{ImageNet + adm}.}
% \label{tab:imagenet_adm_full}
% \end{table*}

\subsection{Comparision against Existing Baselines}\label{sec:compare}
We compare our proposal against the following five state-of-the-art algorithms, using their official open-source implementations for both training and inference. These baseline methods include four reconstruction-based algorithms -- (i) \textbf{DIRE} \cite{Wang2023DIRE}, (ii)  \textbf{LaRE$^2$} \cite{Luo2024LaRE}, and (iii) AEROBLADE \cite{ricker2024aeroblade} (denoted as \textbf{AERO.}) -- as well as (iv) a non–reconstruction-based approach, UniversalFakeDetect (denoted as \textbf{UFD}) \cite{ojha2024universalfakeimagedetectors}, which leverages a feature space not explicitly trained to distinguish real from synthetic images.
Further details for our method and all baselines are reported in the Appendix Section \ref{sup:exp}.

%1) DIRE \cite{Wang2023DIRE} exploits DDIM-based image-level reconstruction residuals as a discriminative clue. 2) FIRE \cite{chu2025fire} uses frequency-guided reconstruction error, 3) LaRe \cite{Luo2024LaRE} detects forgeries via latent-space representation consistency, 4) AEROBLADE(Aero.), a training-free strategy that measures reconstruction errors produced by the autoencoder component of latent diffusion models. 5) UniversalFakeDetect (UFD) \cite{ojha2024universalfakeimagedetectors} using an informative feature space not trained for real-vs-fake classification.

\paragraph{Performance with a Larger Training Dataset.}
% When trained on ImageNet+adm (Table~\ref{tab:imagenet_adm_full}), DID achieves accuracy that is largely comparable to DIRE across all evaluation domains, with essentially identical whole-average scores. Both methods clearly outperform FIRE and AEROBLADE. LaRe shows strong performance on a few specific test sets, especially \textit{lsun+if} and \textit{lsun+sdv1}, while UFD also performs well on a small number of in-domain settings. These gains, however, are isolated: both methods drop markedly on many other domains, including \textit{lsun+pndm}, \textit{lsun+ddpm}, and all LAION-based tests, leading to lower whole-average accuracy compared with DID and DIRE. Overall, on this benchmark, our detector achieves accuracy comparable to DIRE while exhibiting more stable behavior across different sampling variants.
When trained on the ImageNet dataset with 40k real images and 40k synthetic images generated by ADM, we report the accuracy of different detectors in Table~\ref{tab:imagenet_adm_full}. DID achieves performance very similar to DIRE. This is expected for two reasons. First, the training dataset contains a large number of samples. Second, we also use ADM for reconstruction, so the generative model used to produce the synthetic training images aligns with the one used for reconstruction. Under this large-sample and well-aligned setting, the first-order reconstruction error is already highly discriminative for distinguishing real from AI-generated images. As a result, both DID and DIRE achieve classification accuracy close to 100\%. In contrast, other methods such as AEROBLADE, LaRE$^2$ and UFD may perform competitively on certain datasets but are less robust across datasets, and their accuracy generally falls behind both DIRE and our DID.
%Since DID’s first-branch classifier dominates in this scenario, the overall performance naturally matches that of DIRE. 

\paragraph{Performance with a Smaller Training Dataset.}
%To further compare the detectors under a more challenging setting, we evaluate them using the "LAION $\&$ Kan3" and "LAION $\&$ SDXL" training configurations while reducing the training set to 10k real and 10k synthetic images. 
We report the accuracy of various detectors trained on the LAION dataset in Table~\ref{tab:laion_kan3_sdxl_full}. This represents a more challenging setting for two reasons: (i) the sample size is smaller, with only 10k real and 10k synthetic images; (ii) the generators used to create the synthetic training datasets (Kan3 and SDXL) differ from the ADM model used for reconstruction. In such cases, the first-order reconstruction discrepancy alone is no longer sufficiently reliable. Consequently, DID benefits from its second-order reconstruction error and outperforms DIRE in almost all settings, particularly when evaluated on LSUN-B and ImageNet, where the test images differ from the training dataset LAION. Across these cases, our total improvement ranges from 20–30\%. This demonstrates that our second-order reconstruction error plays a crucial role in more complex tasks, especially when the sample size is smaller or when the generator differs from the reconstruction model. Other baseline detectors, such as LaRE$^2$ and UFD, perform well when the training and testing datasets are the same, but their performance drops significantly when the datasets differ. This demonstrates their limited generalization ability. AERO. also fails in most cases.

\subsection{Sensitivity Analysis}\label{sec:sensitivity}

\begin{table*}[t]
\caption{Detection accuracy (\%) on GAN-generated images under different training settings. Best in \textbf{bold}, second best \underline{underlined}. All averages are computed from the raw values and then rounded to two decimals. The last column reports the improvement of DID over the second-best baseline.}
\label{tab:gan_acc_only}
\centering
\renewcommand{\arraystretch}{1}
\begin{tabular}{l l l| c c c c c c }
\toprule
\textbf{Train set \& Gen. model} & \textbf{Eval set} & \textbf{Gen. model} & \multicolumn{5}{c}{\textbf{Accuracy (\%)}} & \textbf{Imp.} \\
\cmidrule(lr){4-8}
 &  &  &
\textbf{DID} & \textbf{DIRE}  & \textbf{LaRE$^2$} &
\textbf{AERO.} & \textbf{UFD} & \\
\midrule
\multirow{3}{*}{LAION \& Kan.3}
& \multirow{3}{*}{LSUN-B}
& StyleGAN          & \textbf{95.8} & \underline{94.3} & 50.4 & 49.8 & 50.8 & 26.32\% \\
& & ProjectedGAN     & \textbf{93.4} & \underline{91.0} & 51.2 & 49.8 & 50.7 & 26.52\% \\
& & Diff-ProjectedGAN& \textbf{92.7} & \underline{90.2} & 51.9 & 49.8 & 51.1 & 25.51\% \\
\cmidrule(lr){3-9}
& & Avg.             & \textbf{93.95} & \underline{91.82} & 51.12 & 49.75 & 50.85 & 26.07\% \\
\midrule
\multirow{3}{*}{LAION \& SDXL}
& \multirow{3}{*}{LSUN-B}
& StyleGAN          & \textbf{96.8} & \underline{96.3} & 50.7 & 49.8 & 50.9 & 12.16\% \\
& & ProjectedGAN     & \textbf{94.8} & \underline{93.8} & 52.8 & 49.8 & 50.9 & 16.13\% \\
& & Diff-ProjectedGAN& \textbf{93.1} & \underline{92.1} & 53.6 & 49.8 & 50.9 & 12.66\% \\
\cmidrule(lr){3-9}
& & Avg.             & \textbf{94.88} & \underline{94.07} & 52.37 & 49.75 & 50.88 & 13.76\% \\
\midrule
\multicolumn{3}{l|}{Total average} &
\textbf{94.42} & \underline{92.94} & 51.74 & 49.75 & 50.87 & 20.90\% \\
\bottomrule
\end{tabular}
\end{table*}

In Section \ref{sec:compare}, all synthetic images were generated using diffusion models. In this section, we further evaluate each detector’s generalization ability by testing it on images produced by generative adversarial networks (GANs). We consider three GAN models, such as StyleGAN, ProjectedGAN, and Diff-ProjectedGAN, and report the ACC of each detector in Table~\ref{tab:gan_acc_only}. Although our model is trained exclusively on diffusion-generated data from “Kandinsky3’’ or “SDXL,’’ it consistently achieves the highest accuracy across all three GAN variants. In every setting, our method outperforms DIRE, with improvements of up to {20.90\%}. The performance gap becomes even more pronounced when compared with LaRE$^2$, and AEROBLADE, all of which fail to generalize to these GAN-generated distributions. UniversalFakeDetect also struggles in this scenario. These results demonstrate that our detector -- despite being trained solely on diffusion-generated images -- remains robust to synthetic images produced by substantially different generative mechanisms. 
We also conduct a complementary sensitivity analysis on image formats (JPEG vs. PNG); details are provided in the Appendix Section \ref{sup:sensitive}.
%captures a model-agnostic signal that transfers reliably to GAN-generated content despite substantial differences in generative architecture and sampling dynamics.

\subsection{Ablation Study}\label{sec:ablation}
In this section, we conduct an ablation study to compare our DID with a variant (denoted as $\Delta^2$) that uses only the second-order reconstruction error and excludes the first-order term. For completeness, we also include DIRE as a reference. To facilitate the computation, we train a lightweight model using only 10\% of the LAION training data and report the results in Table~\ref{tab:kan3_eval_acc}.

% we conduct several ablation studies 
%to evaluate the contribution of each component in our dual detection framework
% .
% \subsubsection{Effect of First- and Second-order Discrepancies}
% To analyze the role of each reconstruction discrepancy
%, we compare three settings of our framework: (i) using only the first-order error $\Delta$ (equivalent to the DIRE setting), (ii) using only the second-order error $\Delta^2$, and (iii) our full DID detector. For this analysis, we run all three settings on a 10\% subset of the data and adopt an identical training and evaluation protocol across them. This provides a lightweight setting that is easy to reproduce while preserving the qualitative conclusions observed on the full dataset.

%As shown in Table~\ref{tab:kan3_eval_acc}, 
We make several observations. First, in most settings, $\Delta^2$ outperforms DIRE, demonstrating that second-order differences are more effective than first-order differences at capturing subtle discrepancies between real and synthetic images. Second, in settings where LSUN is used as the evaluation dataset and synthetic images are generated by ADM, DIRE outperforms $\Delta^2$. A closer inspection reveals that this corresponds to a simple scenario in which the synthetic images are produced by the same model used for reconstruction. This is consistent with our analysis in Section~\ref{sec:method}, which shows that first-order differences are sufficient in simpler settings. Finally, DID outperforms or performs comparably to both $\Delta^2$ and DIRE across all settings, illustrating the advantages of leveraging both first- and second-order differences for classification.
Additional ablations on the fusion strategy are reported in the Appendix Section \ref{sup:ablation_fusion}.

\begin{table}[t]
\caption{Accuracy (\%) on various evaluation datasets when trained on LAION \& Kan3.}
\label{tab:kan3_eval_acc}
\centering
\renewcommand{\arraystretch}{1}
\begin{tabular}{llccc}
\toprule
\textbf{Eval set} & \textbf{Gen. model} & \multicolumn{3}{c}{\textbf{Accuracy (\%)}} \\
\cmidrule(lr){3-5}
 &  & \textbf{DID} & \boldmath{$\Delta^2$} & \textbf{DIRE} \\
\midrule
\multirow{3}{*}{LSUN-B} 
  & ADM  & 86.5 & 69.0 & 84.0 \\
  & PNDM & 83.0 & 81.5 & 67.5 \\
  & DDPM & 82.0 & 78.0 & 71.5 \\
\midrule
\multirow{2}{*}{ImageNet}
  & ADM  & 98.9 & 98.4 & 91.0 \\
  & SDv1 & 98.9 & 98.7 & 93.5 \\
\midrule
\multirow{3}{*}{LAION}
  & Kan.3 & 99.9 & 100 & 99.9 \\
  & SDXL & 99.9 & 99.9  & 99.2 \\
  & Vega & 99.9 & 99.9  & 98.8 \\
\bottomrule
\end{tabular}

\end{table}

\section{Discussion}
This paper studies the detection of AI-generated images. A popular approach is reconstruction-based, which relies on the first-order difference between an input image and its reconstruction as the basis for classification. Although these algorithms perform well when real and AI-generated images differ substantially, they often fail when the two appear highly similar. To address this challenge, we propose a difference-in-difference approach that utilizes a second-order difference. Our analysis in Section \ref{sec:DID} shows that the second-order difference effectively removes perturbation noise and uncovers subtle discrepancies between real and AI-generated images. We then integrate both first- and second-order differences into our classifier to simultaneously handle both settings. Extensive experiments demonstrate that our detector outperforms state-of-the-art algorithms (Section \ref{sec:compare}); an sensitivity analysis shows that its performance is robust across different generative models (Section \ref{sec:sensitivity}); and an ablation study confirms the critical role of the second-order difference (Section \ref{sec:ablation}). 

We focus on using second-order differences for detection, but our idea naturally extends to even higher-order differences. For example, one may define a third-order difference as
\begin{equation}
    \Delta^3(x) = |\Delta^2(x) - \Delta^2(x')|,
\end{equation}
for an imput image $x$ and its reconstruction $x'$. In principle, one could incorporate all higher-order differences into the classifier to capture weak signals. However, computing a $K$th-order difference requires reconstructing the image $K$ times, which significantly increases the computation. Thus, there is an inherent trade-off: higher-order differences may uncover more subtle discrepancies between real and AI-generated images, but at the cost of reduced computational efficiency. Determining the optimal order that balances detection performance and computational cost remains an open question for our future work. 

Finally, we note that although DID is proposed for detecting AI-generated images, the same principle can be readily generalized to the detection of LLM-generated text \citep[see e.g.,][]{mitchell2023detectgpt,baofast,zhouadadetectgpt2025,zhou2026detecting}.

\section*{Acknowledgements}
Qi, Zhou, and Yang's research was supported by the National Natural Science Foundation of China (NSFC) under Grant No.~12271286.

{
    \small
    \bibliographystyle{ieeenat_fullname}
    \bibliography{main}
}

\clearpage
\setcounter{page}{1}
\maketitlesupplementary

The Supplementary Material is organized as follows. Section~\ref{sup:data} describes how the training and test sets are constructed and details the generative models we use. Section~\ref{sup:exp} details the training and evaluation pipelines for our model and all baselines.  Section~\ref{sup:visualization} shows example visualizations from the various datasets used to evaluate DIRE and DID.

\section{More Details of Data}
\label{sup:data}

\subsection{Training Data}

We train models seperately on each training sets for all baseline methods and our method. Our training data includes the following three training sets: \textbf{ImageNet \& ablated diffusion model}, \textbf{LAION \& Kandinsky~3}, and \textbf{LAION \& Stable Diffusion XL}.
\\
\begin{itemize}[leftmargin=*]
    \setlength{\itemsep}{1.5em} 
    \item \textbf{ImageNet \& ablated diffusion model.}
    This training set contains 40k real images from the 1k classes of ImageNet~\cite{deng2009imagenet} and 40k fake images generated by an ablated diffusion model~\cite{dhariwal2021diffusion} constructed and released by the authors of DIRE~\cite{Wang2023DIRE}. 
    We access the data through the official repository at \href{https://github.com/ZhendongWang6/DIRE.git}{this link}.

    \item \textbf{LAION \& Kandinsky~3.}
    This training set consists of 10k real images randomly sampled from LAION~\cite{schuhmann2022laion} and 10k fake images generated with Kandinsky~3~\cite{arkhipkin2023kandinsky} using prompts randomly drawn from the \href{https://huggingface.co/datasets/wanng/midjourney-v5-202304-clean}{Midjourney} prompt set. The detailed generation process is described in Section~\ref{sup:test}.
    
    \item \textbf{LAION \& Stable Diffusion XL.}
    This training set likewise consists of 10k real images randomly sampled from LAION~\cite{schuhmann2022laion} and 10k fake images generated with Stable Diffusion XL (SDXL)~\cite{podell2023sdxl}, using the same pool of \href{https://huggingface.co/datasets/wanng/midjourney-v5-202304-clean}{Midjourney} prompts. The detailed generation process is described in Section~\ref{sup:test}.

\end{itemize}

\subsection{Evaluation Data}
\label{sup:test}
Our evaluation includes both in-distribution and out-of-distribution datasets. For each dataset, the real images are combined with fake images from different sources.
\\ \\ 
\textbf{Real images.}
For each evaluation set, the real images are drawn respectively from three sources: 5k images from ~\cite{deng2009imagenet}, 1k images from LAION~\cite{schuhmann2022laion}, and 1k images from LSUN-B~\cite{yu2015lsun}. 
\\ \\
\textbf{Fake images.}
In ImageNet and LSUN-B based evaluation sets, we directly use all the test samples that the DIRE~\cite{Wang2023DIRE} authors provide in the official datasets at \href{https://github.com/ZhendongWang6/DIRE.git}{this link}. In the LAION-based evaluation sets, we conidered a more challenging setting. We first randomly sample 1k prompts from \href{https://huggingface.co/datasets/wanng/midjourney-v5-202304-clean}{Midjourney} and then generate images using several state-of-the-art text-to-image models on Hugging Face. 
\\ \\
\begin{itemize}[leftmargin=*]
\setlength{\itemsep}{1.5em} 
    \item \textbf{Kandinsky~3.}
    For Kandinsky~3 images, we use the Kandinsky~3 model~\cite{arkhipkin2023kandinsky} from Hugging Face (\href{https://huggingface.co/kandinsky-community/kandinsky-3}{\textit{kandinsky-community/kandinsky-3}}), with the default generation parameters.
    \item \textbf{Stable Diffusion XL.}
    For SDXL images, we follow the standard two-stage Stable Diffusion XL pipeline~\cite{podell2023sdxl} from Hugging Face: we first generate an initial image with the base model (\href{https://huggingface.co/stabilityai/stable-diffusion-xl-base-1.0}{\textit{stabilityai/stable-diffusion-xl-base-1.0}}) and then refine it with the refiner model (\href{https://huggingface.co/stabilityai/stable-diffusion-xl-refiner-1.0}{\textit{stabilityai/stable-diffusion-xl-refiner-1.0}}), while keeping all other parameters at their default values.
    \item \textbf{Segmind Vega.}
    For Segmind Vega images, we use the Segmind Vega model~\cite{gupta2024progressive} from Hugging Face (\href{https://huggingface.co/segmind/Segmind-Vega}{\textit{segmind/Segmind-Vega}}), again with the default generation parameters.
    \item \textbf{Playground~v2.5.}
    For Playground~v2.5 images, we use the Playground~v2.5 model~\cite{li2024playground} from Hugging Face  (\href{https://huggingface.co/playgroundai/playground-v2.5-1024px-aesthetic}{\textit{playgroundai/playground-v2.5-1024px-aesthetic}}), with the default parameters in their usage example.
    \item \textbf{Stable Cascade.}
    For Stable Cascade images, we follow the official two-stage Stable Cascade~\cite{pernias2023wurstchen} setup from Hugging Face: we first sample image latents using the StableCascadePriorPipeline (\href{https://huggingface.co/stabilityai/stable-cascade-prior}{\textit{stabilityai/stable-cascade-prior}}), and then decode these latents into RGB images using the StableCascadeDecoderPipeline (\href{https://huggingface.co/stabilityai/stable-cascade}{\textit{stabilityai/stable-cascade}}), keeping all other parameters at their default values.
\end{itemize}

\subsection{Validation Data}
The validation set follows the same construction pattern as the training data and contains 2k samples from \textbf{LAION \& Kandinsky 3}, 2k from \textbf{LAION \& Stable Diffusion XL}, and 10k from \textbf{ImageNet \& the ablated diffusion model}, respectively, for each training. It is used solely for early stopping to select the best checkpoint. As several baselines are originally implemented with different early stopping strategies in their public codebases, we retain their strategy primarily to guarantee that the baseline models reach their best performance, and we apply the same protocol as DIRE~\cite{Wang2023DIRE} to our method for a strictly fair comparison on the test sets.

\section{More Details of Experiment Implementation}
\label{sup:exp}
% We make three additional remarks:
%  {\color{red}Note that for the first dataset, the generative model used to produce fake images is the same as the reconstruction model in our method (ADM), whereas for the latter two datasets, the generation models differ from our reconstruction model. }

\begin{table*}[t]
\caption{Accuracy (\%) of DID trained on the PNG-only variant. 
Left block: trained on \textbf{LAION \& Kan.3}. 
Right block: trained on \textbf{LAION \& SDXL}. 
Models are evaluated on PNG, JPEG~95, and JPEG~75 versions of each test set. 
All averages are computed from raw values and then rounded to two decimals.}
\label{tab:format_sensitivity}
\centering
\setlength{\tabcolsep}{8pt}
\renewcommand{\arraystretch}{1.05}
\newcolumntype{C}{>{\centering\arraybackslash}p{1.3cm}}
\begin{tabular}{l|CCC|CCC}
\toprule
\multirow{2}{*}{\textbf{Gen. model}} &
\multicolumn{3}{c|}{\textbf{LAION \& Kan.3} (PNG-only training)} &
\multicolumn{3}{c}{\textbf{LAION \& SDXL}} \\
\cmidrule(lr){2-4}\cmidrule(lr){5-7}
& \textbf{PNG} & \textbf{JPG95} & \textbf{JPG75} &
  \textbf{PNG} & \textbf{JPG95} & \textbf{JPG75} \\
\midrule
Kan.3            & 97.2 & 98.9 & 95.1 & 90.6 & 85.4 & 70.4 \\
SDXL             & 93.2 & 88.9 & 76.3 & 97.3 & 97.1 & 85.6 \\
Vega             & 89.1 & 84.3 & 72.3 & 95.2 & 93.0 & 76.3 \\
Playground v2.5  & 94.9 & 92.9 & 82.7 & 91.6 & 88.0 & 73.4 \\
Stable Cascade   & 96.2 & 95.6 & 89.4 & 94.4 & 87.2 & 71.7 \\
\midrule
Avg. & \textbf{94.09} & \textbf{92.09} & \textbf{83.15} &
                   \textbf{93.81} & \textbf{90.13} & \textbf{75.46} \\
\bottomrule
\end{tabular}
\end{table*}

\subsection{Reconstruction Module}

This section provides implementation details of the reconstruction module used in the $\Delta$ and $\Delta'$ branches.  
The reconstruction operator itself follows the standard inversion--reconstruction pipeline described in the main paper (Section~\ref{sec:method}).  
Here, we focus solely on the practical settings required for reproducibility, rather than repeating the theoretical formulation.

\paragraph{Inversion and First Reconstruction}

Each input image is resized or cropped to $256\times256$ and normalized according to the pretrained diffusion model.  
We apply deterministic DDIM inversion to obtain the corresponding latent noise representation.  
During reconstruction, we use DDIM sampling with \textbf{20 steps} (\texttt{timestep\_respacing="ddim20"}) and a fixed \textbf{$\eta=0$}.  
A single sampling pass produces the first reconstructed image $x'_0$, from which we compute the element-wise absolute reconstruction error $\Delta = |x_0 - x'_0|$.

\paragraph{Second Reconstruction and DiffRecon Features}

Using $x'_0$ as the new input, we apply the same DDIM-20 inversion--reconstruction procedure to obtain a second reconstructed image $x''_0$.  
The second-order reconstruction error is defined as $\Delta' = |x'_0 - x''_0|$.  
We then compute the \textit{DiffRecon} feature as $\Delta^2 = \Delta - \Delta'$.  
The resulting error maps are mapped to the standard 8-bit intensity range $[0,255]$ and saved as images, which are subsequently used as inputs to the downstream classifier.

\paragraph{Fixed Inference Without Parameter Updates}

The reconstruction module operates strictly in inference mode, and all U-Net and diffusion-scheduler parameters remain frozen.  
We use publicly released ADM checkpoints and adopt a fixed DDIM-20 configuration ($\eta=0$).  
Except for the image resolution and sampling settings, the reconstruction module introduces \textbf{no additional hyperparameters}.

\paragraph{Dataset-Specific ADM Checkpoints}

All reconstructions are performed with fixed, publicly released ADM checkpoints.
LSUN real and fake images are processed using the ADM model trained on LSUN.
All ImageNet and LAION images—including their synthetic variants—are reconstructed using the unconditional ImageNet $256\times256$ ADM model.
No fine-tuning or parameter adaptation is applied; only the appropriate dataset-specific checkpoint is selected, while all other inversion and sampling configurations remain unchanged.

\subsection{Training Details}

\paragraph{DID ($\Delta$, $\Delta^2$)}
The two branches of DID are trained independently using the $\Delta$ and $\Delta^2$ features generated offline by the reconstruction module.  
During training, no gradients are back-propagated through the reconstruction process or the diffusion model; only the classifier parameters are updated.

We adopt a ResNet-50 classifier initialized with ImageNet-pretrained weights.  
All input images are first resized to $256\times256$.  
During training, we apply light augmentation consisting of random cropping to $224\times224$ and random horizontal flipping, while validation and testing use a center crop to $224\times224$.  
All inputs are normalized.

The classifier is optimized using Adam (default parameters $\beta_1 = 0.9$, $\beta_2 = 0.999$) with an initial learning rate of $10^{-4}$.  
Binary cross-entropy is used as the loss function.  
A unified batch size of 256 is used for all methods.  
Training runs for up to 100 epochs with early stopping based on validation accuracy, and the best validation checkpoint is used for all reported results.

\paragraph{DIRE}
DIRE is trained in the same way as DID, using the $\Delta$ feature as input and following identical preprocessing and classifier settings in their \href{https://github.com/ZhendongWang6/DIRE}{official repository}.

\paragraph{LaRE$^2$}
We reproduce LaRE$^2$ using the \href{https://github.com/luo3300612/lare}{official implementation} released by the authors and employ the provided \texttt{CLipClassifierWMapV6} as the classifier.  
The model uses CLIP RN50 as the visual backbone with official pretrained weights.

Before being fed into the network, input images are normalized and padded--center-cropped to $224\times224$.  
Random cropping is applied during training, while validation and testing use a center crop.  
The classifier is optimized with Adam ($\beta_1 = 0.9$, $\beta_2 = 0.999$) using a learning rate of $10^{-4}$ and a batch size of 256.  
The learning rate is adaptively reduced using \texttt{ReduceLROnPlateau} based on validation accuracy, with a minimum learning rate of $10^{-7}$.

\paragraph{AEROBLADE (AERO.)}
AEROBLADE is used as a training--free baseline.  
The \href{https://github.com/jonasricker/aeroblade}{official implementation} computes reconstruction errors with the pretrained diffusion autoencoder and evaluates them using ROC--based metrics such as AUC, without introducing any learning or threshold selection.  
We follow this setting and keep the autoencoder fixed throughout.

For consistent evaluation across all methods, we introduce a lightweight calibration step:  
reconstruction scores are computed on the training split and a fixed operating threshold is chosen as the $95$th percentile of the training scores.  
This threshold is applied to the test split to compute the reported ACC.  
No parameter optimization is involved, and AEROBLADE remains fully training--free.

\paragraph{UniversalFakeDetect (UFD)}
UniversalFakeDetect is used as a supervised baseline.  
Following the \href{https://github.com/WisconsinAIVision/UniversalFakeDetect}{official implementation} and released configuration, we use a pretrained CLIP ViT-L/14 backbone together with a linear binary classifier for distinguishing real and generated images.  
The backbone is kept frozen during training, and only the classification head is updated on the training split.

All images are resized to $256\times256$, cropped to $224\times224$, augmented with random cropping and horizontal flipping during training, and normalized.  
The classifier head is trained with BCE-with-logits loss using the AdamW optimizer $(\beta_1=0.9,\beta_2=0.999)$, a batch size of 256, and an initial learning rate of $10^{-4}$.  
Validation accuracy is used for model selection, and the best checkpoint is reported on the test split.

\section{Sensitivity to image formats.}
\label{sup:sensitive}

In the main experiments, real images are provided in JPEG format while generated images are stored as PNG files, following the original DIRE setup. To verify that DID does not rely on such JPEG/PNG container differences, we conduct an additional sensitivity analysis.

We rebuild the training set so that all real images come from their official PNG releases (the fake images are already PNG), making the entire training set natively PNG. We then train DID on this PNG-only data and evaluate the resulting model on three test variants: native PNG, JPEG~95, and JPEG~75, all derived from the same PNG originals.

As shown in Table~\ref{tab:format_sensitivity}, DID maintains consistently high performance across all formats. This demonstrates that DID does not depend on JPEG/PNG encoding artifacts and instead captures intrinsic differences between real and generated images.

\section{New baselines and datasets}
\label{sup:new_baseline}

We further compare DID with two newly added baselines, \textit{Effort}~\cite{yan2025orthogonal} and \textit{DDA}~\cite{chen2025dual}, under two training setups: training all models on (i) \textit{GenImage}~\cite{zhu2023genimage} (Table~\ref{sup:tab:acc_genimage_train_singlecol}) and (ii) \textit{ImageNet} (Table~\ref{sup:tab:acc_imagenet_train_singlecol}). We then evaluate these models on \textit{GenImage}~\cite{zhu2023genimage}, \textit{GenDet}~\cite{zhu2023gendet}, and our two original test datasets, \textit{LAION} and \textit{ImageNet}. In addition to the original ACC metric, we also report two newly added metrics, ROC-AUC and FPR@TPR95. Across both tables, DID consistently outperforms all baselines.

\begin{table*}[t]
\caption{Acc (\%) / AUROC / FPR@TPR95 (\%) of DID (with two thresholds) and two additional baselines (\textit{Effort} and \textit{DDA}). All detectors are trained on \textit{GenImage} and evaluated on \textit{GenImage}, \textit{GenDet}, and our two original test datasets. To avoid excessive 1.00 values, we report AUROC with three decimal places.}
%Only averages are reported on the existing datasets to save space.}
\label{sup:tab:acc_genimage_train_singlecol}
\centering
\setlength{\tabcolsep}{8pt}
\renewcommand{\arraystretch}{1.05}
\newcolumntype{C}{>{\centering\arraybackslash}p{1.5cm}}
\begin{tabular}{l l|c c c c @{}}
\toprule
 \textbf{Eval set}&\textbf{Gen. model} &
\textbf{DID$_{0.29}$} & \textbf{DID$_{0.5}$} & \textbf{Effort} & \textbf{DDA} \\
 \midrule
\multirow{9}{*}{GenImage}
 &ADM
 & \underline{99.90}/\textbf{1.000}/\textbf{0.00}
 & \textbf{100.00}/\textbf{1.000}/\textbf{0.00}
 & 98.55/\underline{0.999}/0.40
 & 99.30/\textbf{1.000}/\underline{0.20} \\
 &BigGAN
 & \underline{99.90}/\textbf{1.000}/\textbf{0.00}
 & \textbf{99.95}/\textbf{1.000}/\textbf{0.00}
 & 95.45/0.995/2.60
 & 98.70/\underline{0.999}/\underline{0.50} \\
 &GLIDE
 & \underline{99.90}/\textbf{1.000}/\textbf{0.00}
 & \textbf{100.00}/\textbf{1.000}/\textbf{0.00}
 & 98.35/\underline{0.999}/\underline{0.20}
 & 99.20/\textbf{1.000}/\underline{0.20} \\
 &Midjourney
 & \textbf{99.70}/\textbf{1.000}/\textbf{0.00}
 & \textbf{99.70}/\textbf{1.000}/\textbf{0.00}
 & \underline{57.25}/0.744/80.00
 & 54.95/\underline{0.845}/\underline{56.30} \\
 &SD-v1.4
 & \underline{99.45}/\textbf{1.000}/\textbf{0.00}
 & \textbf{99.50}/\textbf{1.000}/\textbf{0.00}
 & 86.70/\underline{0.975}/\underline{13.00}
 & 72.81/0.965/15.60 \\
 &SD-v1.5
 & \textbf{99.50}/\textbf{1.000}/\textbf{0.00}
 & \underline{99.40}/\textbf{1.000}/\textbf{0.00}
 & 83.95/\underline{0.975}/\underline{11.30}
 & 72.65/0.962/16.50 \\
 &VQDM
 & \textbf{99.90}/\textbf{1.000}/\textbf{0.00}
 & \textbf{99.90}/\textbf{1.000}/\textbf{0.00}
 & 97.00/\underline{0.997}/1.80
 & \underline{99.50}/\textbf{1.000}/\underline{0.30} \\
 &Wukong
 & \textbf{99.20}/\textbf{1.000}/\textbf{0.00}
 & \underline{99.15}/\textbf{1.000}/\textbf{0.00}
 & 83.05/\underline{0.969}/15.60
 & 74.50/0.968/\underline{11.00} \\
 \cmidrule(lr){2-6}
 & Avg.
 & \underline{99.68}/\textbf{1.000}/\textbf{0.00}
 & \textbf{99.70}/\textbf{1.000}/\textbf{0.00}
 & 87.54/0.957/15.61
 & 83.95/\underline{0.967}/\underline{12.33}\\
 \midrule
 GenDet
 &BigGAN
 & \textbf{99.95}/\textbf{1.000}/\textbf{0.00}
 & \textbf{99.95}/\textbf{1.000}/\textbf{0.00}
 & 98.55/\textbf{1.000}/\underline{0.10}
 & \underline{99.20}/\textbf{1.000}/\textbf{0.00} \\
 &CRN
 & \textbf{99.95}/\textbf{1.000}/\textbf{0.00}
 & \textbf{99.95}/\textbf{1.000}/\textbf{0.00}
 & 98.15/0.998/0.60
 & \underline{98.45}/\underline{0.999}/\underline{0.40} \\
 &DeepFake
 & \textbf{99.65}/\textbf{1.000}/\textbf{0.00}
 & \underline{99.45}/\textbf{1.000}/\textbf{0.00}
 & 96.35/\underline{0.995}/2.50
 & 99.20/\textbf{1.000}/\textbf{0.00} \\
 &GauGAN
 & \textbf{99.95}/\textbf{1.000}/\textbf{0.00}
 & \textbf{99.95}/\textbf{1.000}/\textbf{0.00}
 & 97.45/\underline{0.997}/0.90
 & \underline{99.35}/\textbf{1.000}/\textbf{0.00} \\
 &IMLE
 & \textbf{99.95}/\textbf{1.000}/\textbf{0.00}
 & \textbf{99.95}/\textbf{1.000}/\textbf{0.00}
 & \underline{98.70}/\textbf{1.000}/\underline{0.10}
 & 97.70/\underline{0.998}/1.00 \\
 &StarGAN
 & \textbf{99.95}/\textbf{1.000}/\textbf{0.00}
 & \textbf{99.95}/\textbf{1.000}/\textbf{0.00}
 & \underline{98.50}/\underline{0.999}/0.20
 & 99.35/\textbf{1.000}/\textbf{0.00} \\
 \cmidrule(lr){2-6}
 & Avg.
 & \textbf{99.90}/\textbf{1.000}/\textbf{0.00}
 & \underline{99.87}/\textbf{1.000}/\textbf{0.00}
 & 97.95/0.998/0.73
 & 98.88/\underline{0.999}/\underline{0.23} \\
 \midrule
 LAION 
 & Kan.3
 & \textbf{99.60}/\textbf{1.000}/\textbf{0.00}
 & \textbf{99.60}/\textbf{1.000}/\textbf{0.00}
 & \underline{63.85}/\underline{0.845}/61.00
 & 57.10/0.832/\underline{48.20} \\
 & SDXL
 & \textbf{99.60}/\textbf{0.999}/\textbf{0.00}
 & \underline{99.50}/\textbf{0.999}/\textbf{0.00}
 & 66.85/0.870/59.20
 & 67.70/\underline{0.907}/\underline{31.50} \\
 & Vega
 & \textbf{98.90}/\textbf{0.999}/\textbf{0.00}
 & \underline{98.75}/\textbf{0.999}/\textbf{0.00}
 & 57.50/0.753/78.10
 & 76.90/\underline{0.943}/\underline{23.30} \\
 & Playground v2.5
 & \textbf{99.60}/\textbf{1.000}/\textbf{0.00}
 & \underline{99.55}/\textbf{1.000}/\textbf{0.00}
 & 67.05/0.865/62.70
 & 81.55/\underline{0.959}/\underline{17.70} \\
 & Stable Cascade
 & \textbf{99.65}/\textbf{1.000}/\textbf{0.00}
 & \underline{99.55}/\textbf{1.000}/\textbf{0.00}
 & 81.20/\underline{0.960}/\underline{20.00}
 & 69.20/0.920/28.00 \\
\cmidrule(lr){2-6}
 & Avg.
 & \textbf{99.47}/\textbf{0.999}/\textbf{0.00}
 & \underline{99.39}/\textbf{0.999}/\textbf{0.00}
 & 67.29/0.859/56.20
 & 70.49/\underline{0.912}/\underline{29.74} \\
\midrule
 ImageNet
 & ADM
 & \textbf{99.65}/\textbf{1.000}/\textbf{0.00}
 & \textbf{99.65}/\textbf{1.000}/\textbf{0.00}
 & 97.20/\underline{0.996}/1.20
 & \underline{99.05}/\textbf{1.000}/\underline{0.20} \\
 & SDv1
 & \textbf{99.35}/\textbf{1.000}/\textbf{0.00}
 & \underline{98.70}/\textbf{1.000}/\textbf{0.00}
 & 84.70/0.963/18.10
 & 71.05/\underline{0.967}/\underline{12.10} \\
 \cmidrule(lr){2-6}
 & Avg.
 & \textbf{99.50}/\textbf{1.000}/\textbf{0.00}
 & \underline{99.17}/\textbf{1.000}/\textbf{0.00}
 & 90.95/0.979/9.65
 & 85.05/\underline{0.983}/\underline{6.15} \\
\midrule
\multicolumn{2}{l|}{Total average}
 & \textbf{99.68}/\textbf{1.000}/\textbf{0.00}
 & \underline{99.62}/\textbf{1.000}/\textbf{0.00}
 & 86.02/0.947/20.46
 & 85.11/\underline{0.965}/\underline{12.43} \\
\bottomrule
\end{tabular}
\end{table*}

\begin{table*}[t]
\caption{Acc (\%) / AUROC / FPR@TPR95 (\%) of DID (with two thresholds) and two additional baselines (\textit{Effort} and \textit{DDA}). All detectors are trained on \textit{ImageNet} and evaluated on \textit{GenImage}, \textit{GenDet}, and our two original test datasets, \textit{LAION} and \textit{ImageNet}. To avoid excessive 1.00 values, we report AUROC with three decimal places.}
\label{sup:tab:acc_imagenet_train_singlecol}
\centering
\setlength{\tabcolsep}{8pt}
\renewcommand{\arraystretch}{1.05}
\newcolumntype{C}{>{\centering\arraybackslash}p{1.5cm}}
\begin{tabular}{l l|c c c c @{}}
\toprule
\textbf{Eval set} & \textbf{Gen. model} &
\textbf{DID$_{0.29}$} & \textbf{DID$_{0.5}$} & \textbf{Effort} & \textbf{DDA} \\
\midrule
\multirow{9}{*}{GenImage}
& ADM
& \underline{99.20}/\textbf{1.000}/\textbf{0.00}
& \textbf{99.50}/\textbf{1.000}/\textbf{0.00}
& 86.30/\underline{0.986}/6.70
& 98.30/\underline{0.999}/\underline{0.10} \\
& BigGAN
& \underline{99.10}/\textbf{1.000}/\textbf{0.00}
& \textbf{99.40}/\textbf{1.000}/\textbf{0.00}
& 80.15/0.878/\underline{39.00}
& 98.50/\underline{1.000}/\textbf{0.00} \\
& GLIDE
& \underline{99.20}/\textbf{1.000}/\textbf{0.00}
& \textbf{99.50}/\textbf{1.000}/\textbf{0.00}
& 85.85/0.973/12.80
& 97.50/\underline{0.998}/\underline{0.90} \\
& Midjourney
& \underline{99.20}/\textbf{1.000}/\textbf{0.00}
& \textbf{99.50}/\textbf{1.000}/\textbf{0.00}
& 57.00/0.597/\underline{90.20}
& 56.15/\underline{0.810}/\underline{58.80} \\
& SD-v1.4
& \underline{99.20}/\textbf{1.000}/\textbf{0.00}
& \textbf{99.50}/\textbf{1.000}/\textbf{0.00}
& 85.30/0.958/19.10
& 82.91/\underline{0.969}/\underline{13.30} \\
& SD-v1.5
& \underline{99.20}/\textbf{1.000}/\textbf{0.00}
& \textbf{99.50}/\textbf{1.000}/\textbf{0.00}
& 84.75/0.950/22.10
& 83.60/\underline{0.967}/\underline{17.00} \\
& VQDM
& \underline{99.20}/\textbf{1.000}/\textbf{0.00}
& \textbf{99.50}/\textbf{1.000}/\textbf{0.00}
& 86.45/0.991/\underline{5.40}
& 98.55/\underline{1.000}/\textbf{0.00} \\
& Wukong
& \underline{99.20}/\textbf{1.000}/\textbf{0.00}
& \textbf{99.45}/\textbf{1.000}/\textbf{0.00}
& 84.40/0.949/22.80
& 84.10/\underline{0.973}/\underline{12.70} \\
\cmidrule(lr){2-6}
& Avg.
& \underline{99.19}/\textbf{1.000}/\textbf{0.00}
& \textbf{99.48}/\textbf{1.000}/\textbf{0.00}
& 81.28/0.910/27.26
& 87.45/\underline{0.964}/\underline{12.85} \\
\midrule
GenDet
& BigGAN
& \underline{99.85}/\textbf{1.000}/\textbf{0.00}
& \textbf{99.90}/\textbf{1.000}/\textbf{0.00}
& 99.70/\underline{1.000}/\textbf{0.00}
& 99.45/1.000/\textbf{0.00} \\
& CRN
& \underline{99.85}/\textbf{1.000}/\textbf{0.00}
& \textbf{99.90}/\textbf{1.000}/\textbf{0.00}
& 98.80/\underline{1.000}/0.30
& 99.35/0.999/\underline{0.10} \\
& DeepFake
& \textbf{99.50}/\textbf{1.000}/\underline{0.10}
& 99.20/\textbf{1.000}/\underline{0.10}
& 96.75/0.997/0.80
& \underline{99.30}/\underline{1.000}/\textbf{0.00} \\
& GauGAN
& \underline{99.85}/\textbf{1.000}/\textbf{0.00}
& \textbf{99.90}/\textbf{1.000}/\textbf{0.00}
& 99.65/\underline{1.000}/\textbf{0.00}
& 99.60/\textbf{1.000}/\textbf{0.00} \\
& IMLE
& \underline{99.85}/\textbf{1.000}/\textbf{0.00}
& \textbf{99.90}/\textbf{1.000}/\textbf{0.00}
& 99.75/\underline{1.000}/\textbf{0.00}
& 99.40/1.000/\underline{0.10} \\
& StarGAN
& \underline{99.85}/\textbf{1.000}/\textbf{0.00}
& \textbf{99.90}/\textbf{1.000}/\textbf{0.00}
& 99.65/1.000/\textbf{0.00}
& 99.60/\underline{1.000}/\textbf{0.00} \\
\cmidrule(lr){2-6}
& Avg.
& \textbf{99.79}/\textbf{1.000}/\textbf{0.02}
& \underline{99.78}/\textbf{1.000}/\textbf{0.02}
& 99.05/0.999/0.18
& 99.45/\underline{1.000}/\underline{0.03} \\
\midrule
LAION
& Kan.3
& \underline{99.30}/\textbf{1.000}/\textbf{0.00}
& \textbf{99.55}/\textbf{1.000}/\textbf{0.00}
& 80.75/\underline{0.934}/\underline{31.30}
& 59.05/0.826/47.70 \\
& Playground v2.5
& \underline{99.30}/\textbf{1.000}/\textbf{0.00}
& \textbf{99.60}/\textbf{1.000}/\textbf{0.00}
& 83.95/0.941/28.60
& 87.00/\underline{0.962}/\underline{14.80} \\
& SDXL
& \underline{99.25}/\textbf{0.999}/\textbf{0.00}
& \textbf{99.55}/\textbf{0.999}/\textbf{0.00}
& 85.00/\underline{0.953}/\underline{21.30}
& 76.30/0.911/35.20 \\
& Stable Cascade
& \underline{99.35}/\textbf{1.000}/\textbf{0.00}
& \textbf{99.65}/\textbf{1.000}/\textbf{0.00}
& 90.60/\underline{0.976}/\underline{12.60}
& 75.50/0.928/25.40 \\
& Vega
& \underline{99.25}/\textbf{1.000}/\textbf{0.00}
& \textbf{99.55}/\textbf{1.000}/\textbf{0.00}
& 79.85/0.922/34.30
& 80.80/\underline{0.944}/\underline{22.30} \\
\cmidrule(lr){2-6}
& Avg.
& \underline{99.29}/\textbf{1.000}/\textbf{0.00}
& \textbf{99.58}/\textbf{1.000}/\textbf{0.00}
& 84.03/\underline{0.945}/\underline{25.62}
& 75.73/0.914/29.08 \\
\midrule
ImageNet
& ADM
& \underline{99.85}/\textbf{1.000}/\textbf{0.00}
& \textbf{99.90}/\textbf{1.000}/\textbf{0.00}
& 99.40/0.999/\textbf{0.00}
& 99.55/\underline{1.000}/\textbf{0.00} \\
& SDv1
& \underline{99.85}/\textbf{1.000}/\textbf{0.00}
& \textbf{99.90}/\textbf{1.000}/\textbf{0.00}
& 98.10/\underline{0.999}/\underline{0.30}
& 83.45/0.987/6.40 \\
\cmidrule(lr){2-6}
& Avg.
& \underline{99.85}/\textbf{1.000}/\textbf{0.00}
& \textbf{99.90}/\textbf{1.000}/\textbf{0.00}
& 98.75/\underline{1.000}/\underline{0.15}
& 91.50/0.993/3.20 \\
\midrule
\multicolumn{2}{l|}{Total average}
& \underline{99.45}/\textbf{1.000}/\textbf{0.00}
& \textbf{99.63}/\textbf{1.000}/\textbf{0.00}
& 88.67/0.953/16.55
& 88.47/\underline{0.965}/\underline{12.13} \\
\bottomrule
\end{tabular}
\end{table*}

\begin{table}[th]
\centering
\small
\caption{\small Per-image inference time (seconds) and memory usage (MB) for LaRE2, DIRE, and DID.}\label{tab:computational_cost}
\setlength{\tabcolsep}{8pt}
\renewcommand{\arraystretch}{1.05}
\begin{tabular}{@{}lcccc@{}}
\toprule
\multirow{2}{*}{Method} & Reconstruction  & Testing \\
 & Time (Memory)  & Time  (Memory) \\
\midrule
LaRE2 & 0.09 (33.34)   & 0.66 (9.74) \\
DIRE  & 1.01 (396.48)  & 0.34 (21.77) \\
DID   & 2.01 (398.39)  & 0.45 (22.55) \\
\bottomrule
\end{tabular}
\end{table}

\section{Computation and Memory Cost}
\label{sup:compute_cost}

Table~\ref{tab:computational_cost} reports the inference cost on a single NVIDIA A800 GPU.
DID takes 2.46 seconds per image, including both reconstruction and testing.
For comparison, DIRE and LaRE take 1.35s and 0.75s per image, respectively.
Although DID is slower, it provides a clear accuracy gain (Table~1 in the main text).
DID’s peak GPU memory is similar to DIRE (Table~\ref{tab:computational_cost}),
so it can run on a single GPU.

\begin{table}[th]
\caption{\small Acc(\%) / AUROC / FPR@TPR95(\%) of DID$_{0.29}$ and a variant that fits both 1st- and 2nd-order signals into a single classifier (DID$_{SC}$). }
\label{reb:tab:DID_single_NN}
\centering
\small
\setlength{\tabcolsep}{8pt}
\renewcommand{\arraystretch}{1.05}
\begin{tabular}{@{}l l cc@{}}
\toprule
\textbf{Train set} & \textbf{Eval set} & \textbf{LAION} & \textbf{ImageNet} \\
\midrule
\multirow{2}{*}{\textbf{ImageNet}} & DID$_{0.29}$ & 99.29/1.00/0.00 & 99.85/1.00/0.00 \\
                         & DID$_{SC}$   & 99.03/1.00/0.00 & 99.45/1.00/0.00 \\
\bottomrule
\end{tabular}
\end{table}

\section{Ablation on Fusion Strategy}
\label{sup:ablation_fusion}

Our method uses two separate classifiers to process 1st- and 2nd-order signals, and combines their outputs with an additive gate.
This section studies the fusion strategy.
Specifically, we compare our score-level fusion (two classifiers with an additive gate) with a feature-level variant that concatenates
1st- and 2nd-order features and feeds them into a single classifier.

As shown in Table~\ref{reb:tab:DID_single_NN}, the single-classifier variant achieves performance comparable to our main design.
We keep the two-classifier design mainly for simplicity and interpretability.
Since 1st- and 2nd-order reconstruction errors capture different signals, separating them makes the contribution of each signal easier to analyze.
The two-classifier design is also more memory-efficient, as it uses fewer parameters than the single-classifier variant.

\section{More Visualization About DID}
\label{sup:visualization}
\begin{figure*}[t]
    \centering
    \setlength{\tabcolsep}{1pt}
    \renewcommand{\arraystretch}{1}
    \begin{tabular}{c@{\hspace{2pt}}cccc cccc}
        % -------- real ----------
        \raisebox{0.05\textwidth}{\rotatebox[origin=c]{90}{Real}} &
        \includegraphics[width=0.12\textwidth]{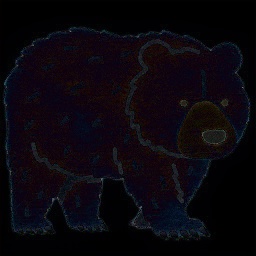} &
        \includegraphics[width=0.12\textwidth]{sup_real/017_dire.jpg} &
        \includegraphics[width=0.12\textwidth]{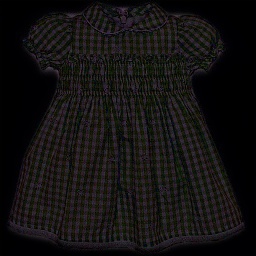} &
        \includegraphics[width=0.12\textwidth]{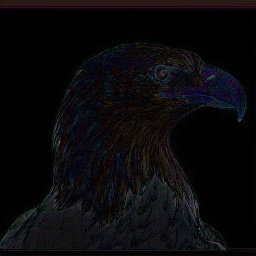} &
        \includegraphics[width=0.12\textwidth]{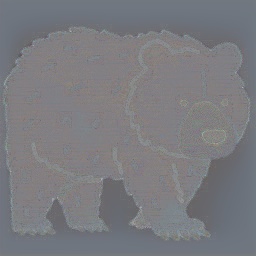} &
        \includegraphics[width=0.12\textwidth]{sup_real/017_diff.jpg} &
        \includegraphics[width=0.12\textwidth]{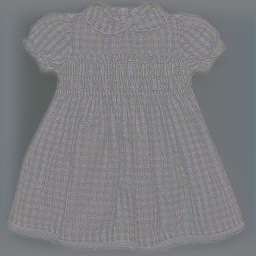} &
        \includegraphics[width=0.12\textwidth]{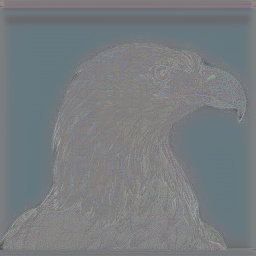} \\
        
        % -------- adm ----------
        \raisebox{0.05\textwidth}{\rotatebox[origin=c]{90}{ADM}} &
        \includegraphics[width=0.12\textwidth]{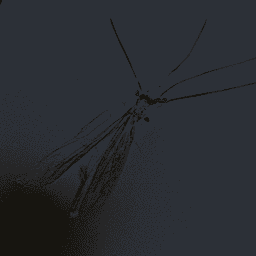} &
        \includegraphics[width=0.12\textwidth]{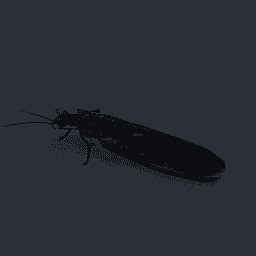} &
        \includegraphics[width=0.12\textwidth]{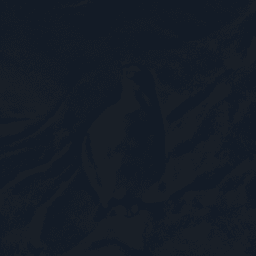} &
        \includegraphics[width=0.12\textwidth]{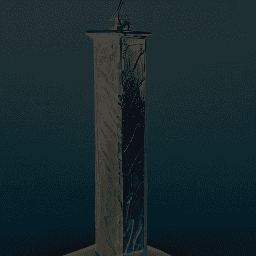} &
        \includegraphics[width=0.12\textwidth]{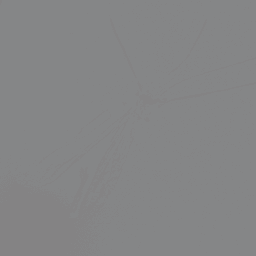} &
        \includegraphics[width=0.12\textwidth]{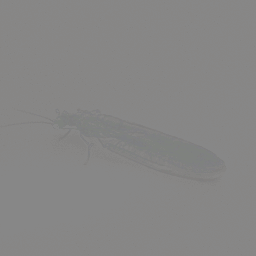} &
        \includegraphics[width=0.12\textwidth]{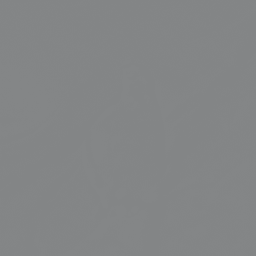} &
        \includegraphics[width=0.12\textwidth]{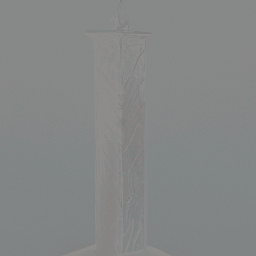} \\
        
        % -------- kan3 ----------
        \raisebox{0.05\textwidth}{\rotatebox[origin=c]{90}{Kan.3}} &
        \includegraphics[width=0.12\textwidth]{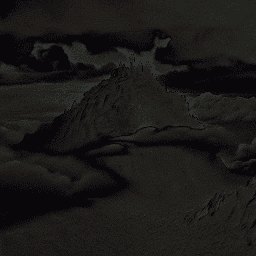} &
        \includegraphics[width=0.12\textwidth]{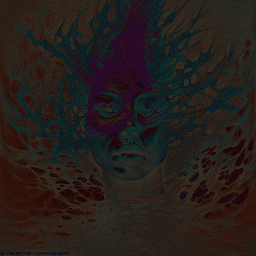} &
        \includegraphics[width=0.12\textwidth]{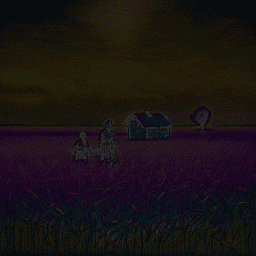} &
        \includegraphics[width=0.12\textwidth]{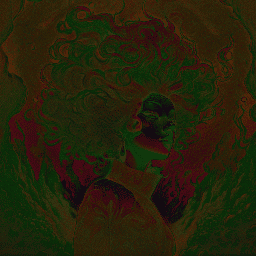} &
        \includegraphics[width=0.12\textwidth]{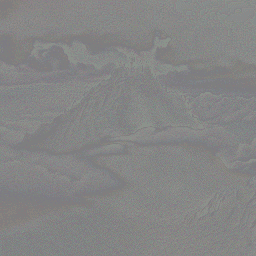} &
        \includegraphics[width=0.12\textwidth]{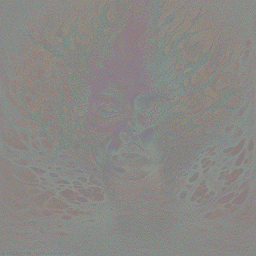} &
        \includegraphics[width=0.12\textwidth]{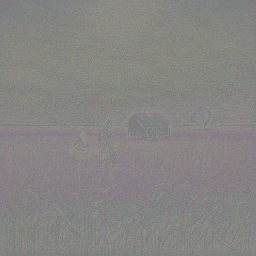} &
        \includegraphics[width=0.12\textwidth]{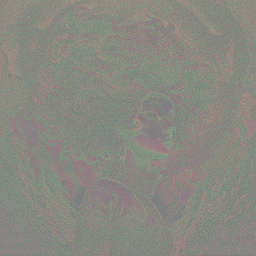} \\
        
        % -------- vega ----------
        \raisebox{0.05\textwidth}{\rotatebox[origin=c]{90}{Vega}} &
        \includegraphics[width=0.12\textwidth]{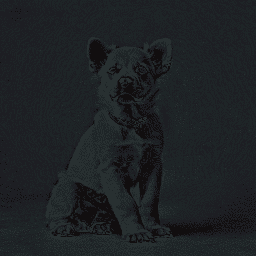} &
        \includegraphics[width=0.12\textwidth]{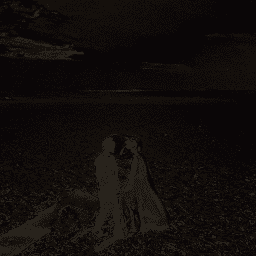} &
        \includegraphics[width=0.12\textwidth]{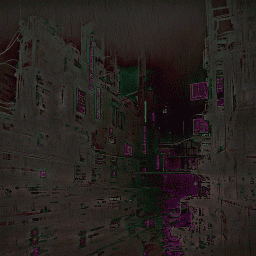} &
        \includegraphics[width=0.12\textwidth]{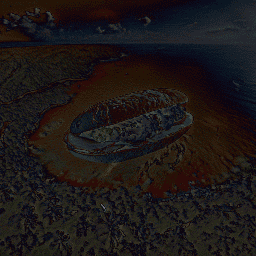} &
        \includegraphics[width=0.12\textwidth]{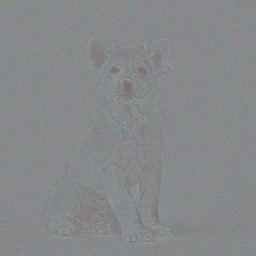} &
        \includegraphics[width=0.12\textwidth]{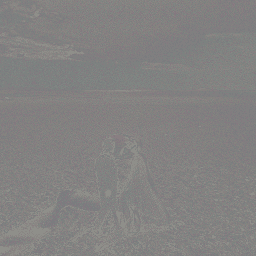} &
        \includegraphics[width=0.12\textwidth]{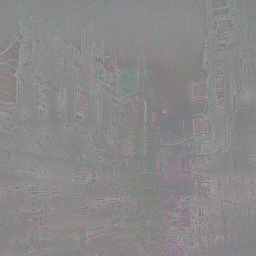} &
        \includegraphics[width=0.12\textwidth]{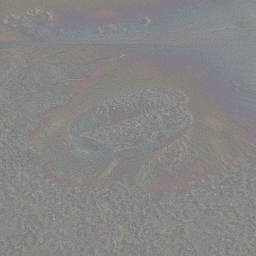} \\
        
        % -------- sdxl ----------
        \raisebox{0.05\textwidth}{\rotatebox[origin=c]{90}{SDXL}} &
        \includegraphics[width=0.12\textwidth]{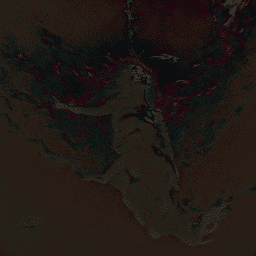} &
        \includegraphics[width=0.12\textwidth]{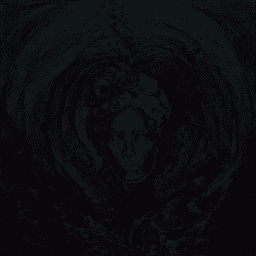} &
        \includegraphics[width=0.12\textwidth]{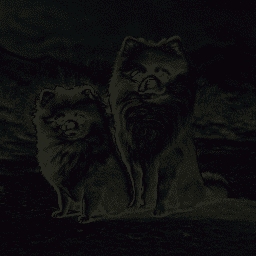} &
        \includegraphics[width=0.12\textwidth]{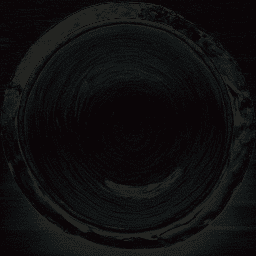} &
        \includegraphics[width=0.12\textwidth]{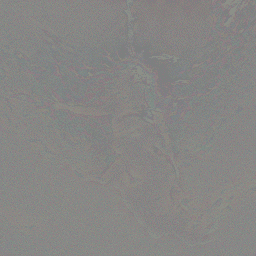} &
        \includegraphics[width=0.12\textwidth]{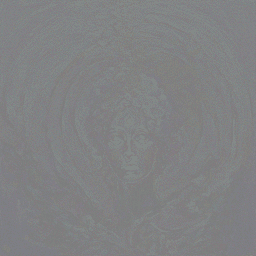} &
        \includegraphics[width=0.12\textwidth]{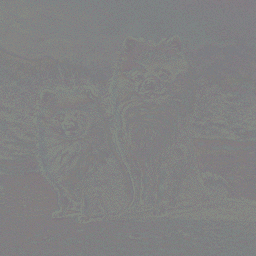} &
        \includegraphics[width=0.12\textwidth]{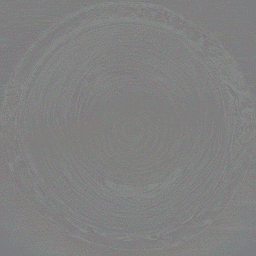} \\
        
        % -------- playground ----------
        \raisebox{0.055\textwidth}{\rotatebox[origin=c]{90}{Playground-2.5}} &
        \includegraphics[width=0.12\textwidth]{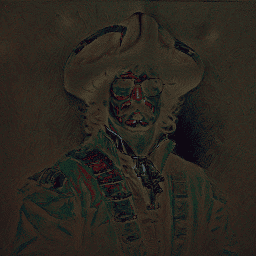} &
        \includegraphics[width=0.12\textwidth]{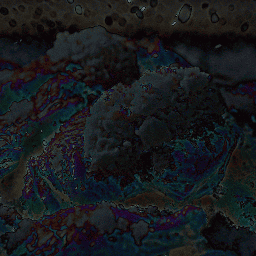} &
        \includegraphics[width=0.12\textwidth]{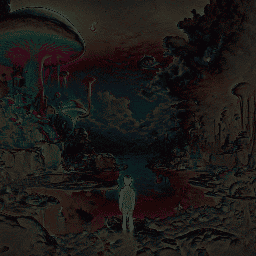} &
        \includegraphics[width=0.12\textwidth]{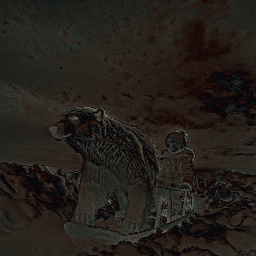} &
        \includegraphics[width=0.12\textwidth]{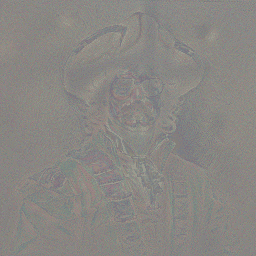} &
        \includegraphics[width=0.12\textwidth]{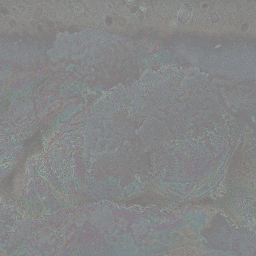} &
        \includegraphics[width=0.12\textwidth]{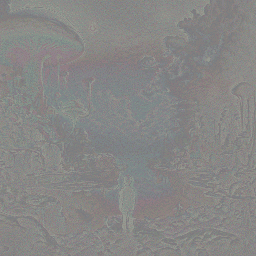} &
        \includegraphics[width=0.12\textwidth]{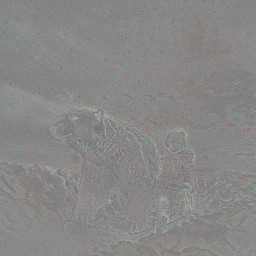} \\
        
        % -------- stable ----------
        \raisebox{0.05\textwidth}{\rotatebox[origin=c]{90}{Stable Cascade}} &
        \includegraphics[width=0.12\textwidth]{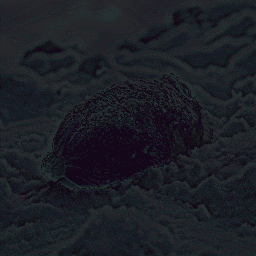} &
        \includegraphics[width=0.12\textwidth]{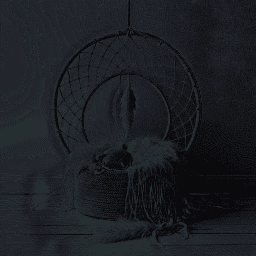} &
        \includegraphics[width=0.12\textwidth]{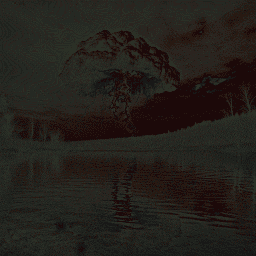} &
        \includegraphics[width=0.12\textwidth]{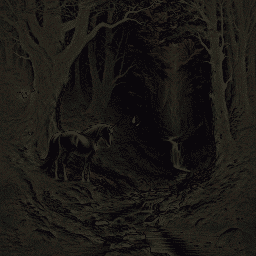} &
        \includegraphics[width=0.12\textwidth]{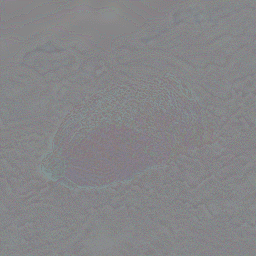} &
        \includegraphics[width=0.12\textwidth]{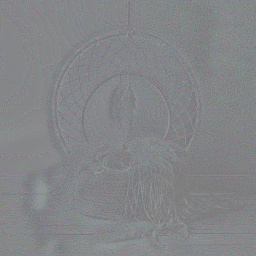} &
        \includegraphics[width=0.12\textwidth]{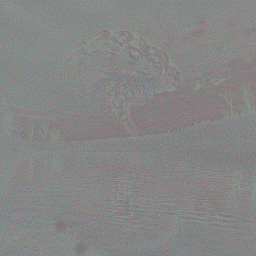} &
        \includegraphics[width=0.12\textwidth]{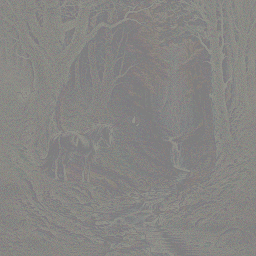} \\
        % -------- braces at bottom ----------
        & \multicolumn{4}{c}{$\underbrace{\rule{0.48\textwidth}{0pt}}_{\mathlarger{\bm{\Delta}(x)}}$} &
          \multicolumn{4}{c}{$\underbrace{\rule{0.48\textwidth}{0pt}}_{\mathlarger{\bm{\Delta}^2(x)}}$} \\

    \end{tabular}
    \caption{
\textbf{Comparison of first--order and second--order residuals across real and generated images.}
The DIRE framework \cite{Wang2023DIRE} shows that real images typically produce sharper and more structured first-order residuals $\Delta x$, whereas fake images tend to yield weaker patterns. However, in broader evaluation across diverse and strong generative models, first-order residuals are not always stable when the real and synthetic distributions are close. In more challenging cases, fake images can even produce stronger residual signals than real images, leading to potential misclassification. In contrast, our proposed second-order residual module $\Delta^2 x$ provides more consistent separability between real and fake samples by taking the difference between the reconstruction error of the input image and that of its reconstructed version.
}
    \label{fig:part1}
\end{figure*}

\begin{figure*}[t]
    \centering
    \setlength{\tabcolsep}{1pt}
    \renewcommand{\arraystretch}{0.1}
    \begin{tabular}{c@{\hspace{2pt}}cccccc}
        \raisebox{0.07\textwidth}{\rotatebox[origin=c]{0}{$x$}} &
        \includegraphics[width=0.15\textwidth]{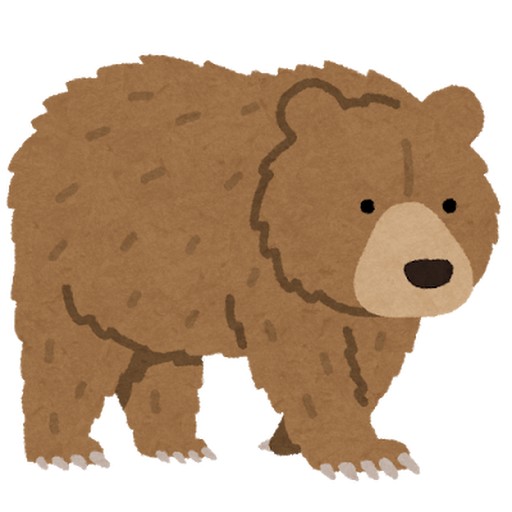} &
        \includegraphics[width=0.15\textwidth]{sup_real/017_real.jpg} &
        \includegraphics[width=0.15\textwidth]{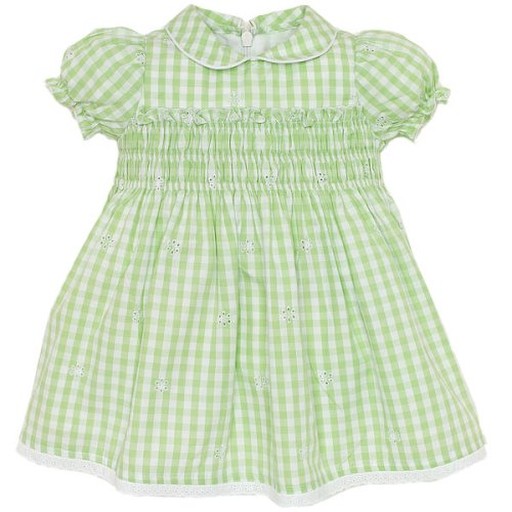} &
        \includegraphics[width=0.15\textwidth]{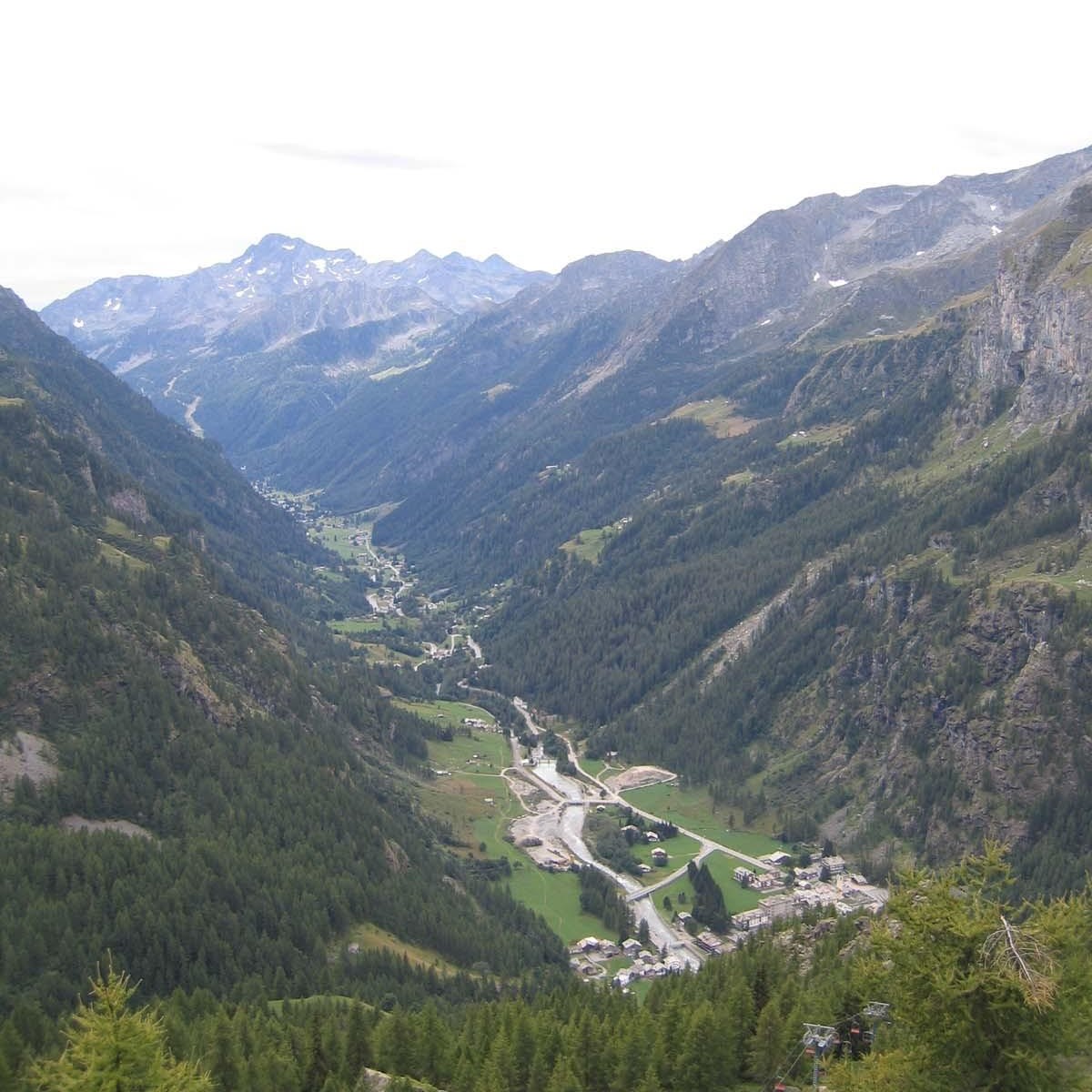} &
        \includegraphics[width=0.15\textwidth]{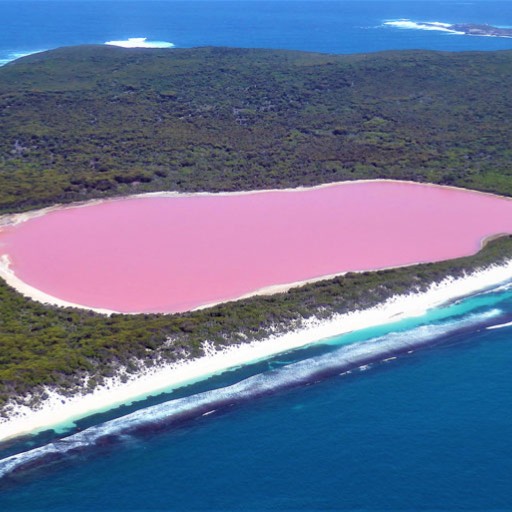} &
        \includegraphics[width=0.15\textwidth]{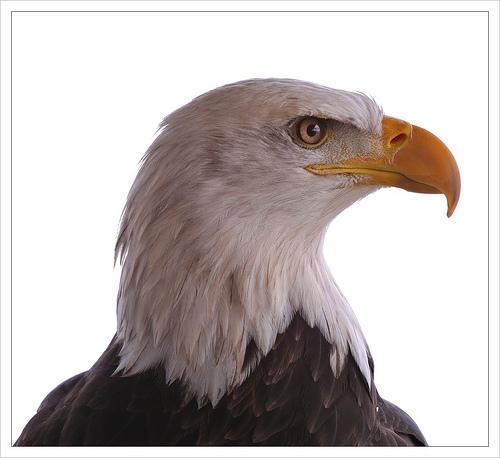} \\
        
        \raisebox{0.07\textwidth}{\rotatebox[origin=c]{0}{$x'$}} &
        \includegraphics[width=0.15\textwidth]{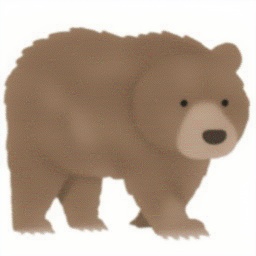} &
        \includegraphics[width=0.15\textwidth]{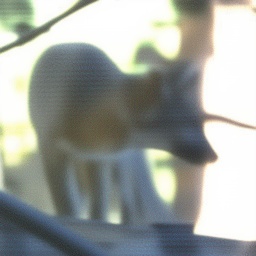} &
        \includegraphics[width=0.15\textwidth]{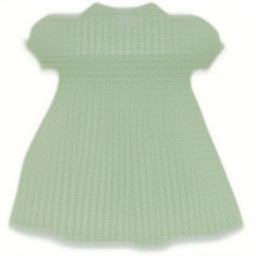} &
        \includegraphics[width=0.15\textwidth]{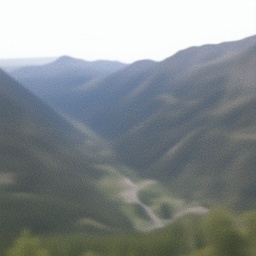} &
        \includegraphics[width=0.15\textwidth]{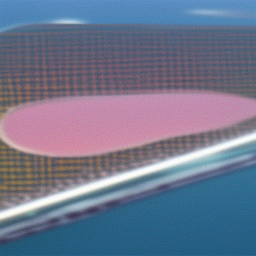} &
        \includegraphics[width=0.15\textwidth]{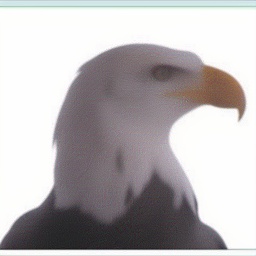} \\
        
        \raisebox{0.07\textwidth}{\rotatebox[origin=c]{0}{$\Delta x$}} &
        \includegraphics[width=0.15\textwidth]{sup_real/002_dire.jpg} &
        \includegraphics[width=0.15\textwidth]{sup_real/017_dire.jpg} &
        \includegraphics[width=0.15\textwidth]{sup_real/034_dire.jpg} &
        \includegraphics[width=0.15\textwidth]{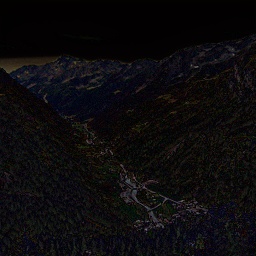} &
        \includegraphics[width=0.15\textwidth]{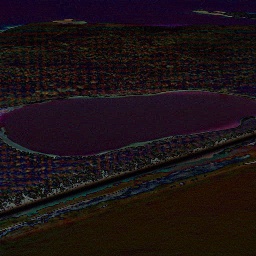} &
        \includegraphics[width=0.15\textwidth]{sup_real/098_dire.jpg} \\
        
        \raisebox{0.07\textwidth}{\rotatebox[origin=c]{0}{$x''$}} &
        \includegraphics[width=0.15\textwidth]{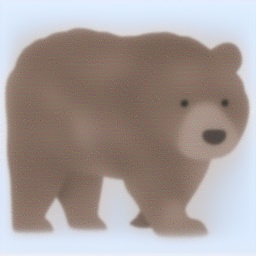} &
        \includegraphics[width=0.15\textwidth]{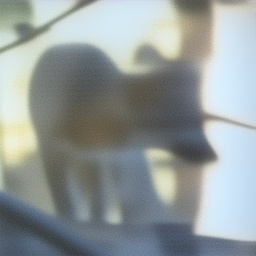} &
        \includegraphics[width=0.15\textwidth]{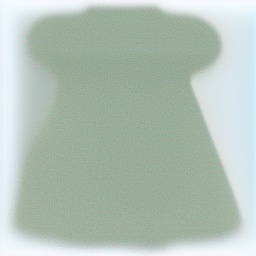} &
        \includegraphics[width=0.15\textwidth]{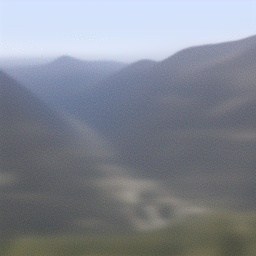} &
        \includegraphics[width=0.15\textwidth]{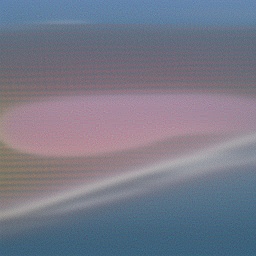} &
        \includegraphics[width=0.15\textwidth]{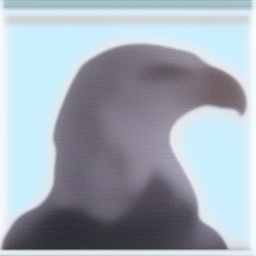} \\
        
        \raisebox{0.07\textwidth}{\rotatebox[origin=c]{0}{$\Delta x'$}} &
        \includegraphics[width=0.15\textwidth]{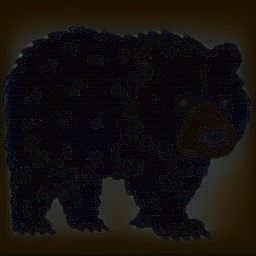} &
        \includegraphics[width=0.15\textwidth]{sup_real/017_dire2.jpg} &
        \includegraphics[width=0.15\textwidth]{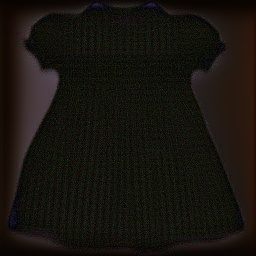} &
        \includegraphics[width=0.15\textwidth]{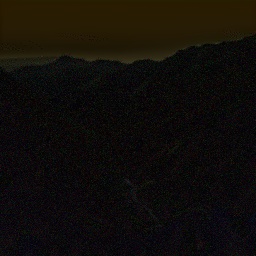} &
        \includegraphics[width=0.15\textwidth]{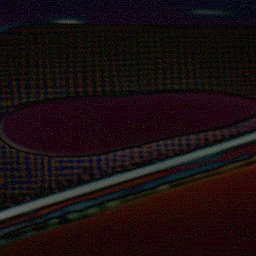} &
        \includegraphics[width=0.15\textwidth]{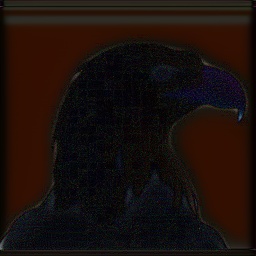} \\
        
        \raisebox{0.07\textwidth}{\rotatebox[origin=c]{0}{$\Delta^2 x$}} &
        \includegraphics[width=0.15\textwidth]{sup_real/002_diff.jpg} &
        \includegraphics[width=0.15\textwidth]{sup_real/017_diff.jpg} &
        \includegraphics[width=0.15\textwidth]{sup_real/034_diff.jpg} &
        \includegraphics[width=0.15\textwidth]{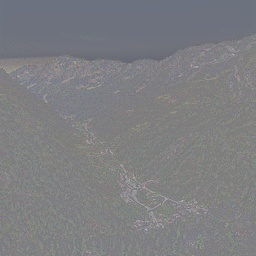} &
        \includegraphics[width=0.15\textwidth]{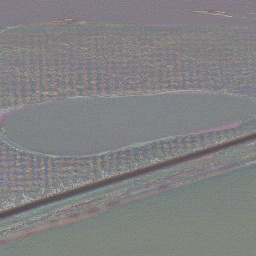} &
        \includegraphics[width=0.15\textwidth]{sup_real/098_diff.jpg} \\
    \end{tabular}
    \caption{
    \textbf{Visualization of real images across the DID pipeline.}
    We show the original image $x$, the first reconstruction $x'$, the second reconstruction $x''$, and their corresponding residual maps $\Delta x$, $\Delta x'$, and $\Delta^2 x$. 
    }
    \label{fig:sup_real_matrix}
\end{figure*}

\begin{figure*}[t]
    \centering
    \setlength{\tabcolsep}{1pt}
    \renewcommand{\arraystretch}{0.1}
    \begin{tabular}{c@{\hspace{2pt}}cccccc}
        % x (real)
        \raisebox{0.07\textwidth}{\rotatebox[origin=c]{0}{$x$}} &
        \includegraphics[width=0.15\textwidth]{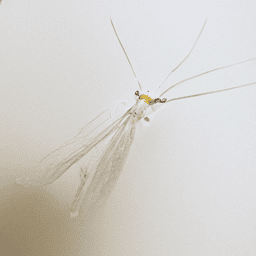} &
        \includegraphics[width=0.15\textwidth]{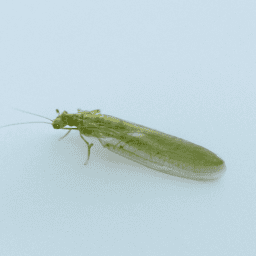} &
        \includegraphics[width=0.15\textwidth]{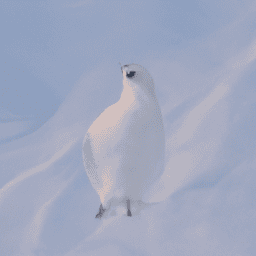} &
        \includegraphics[width=0.15\textwidth]{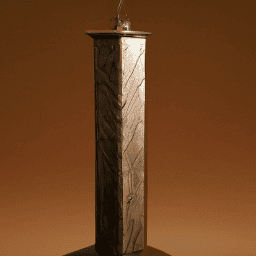} &
        \includegraphics[width=0.15\textwidth]{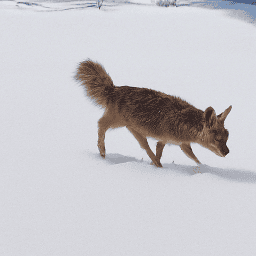} &
        \includegraphics[width=0.15\textwidth]{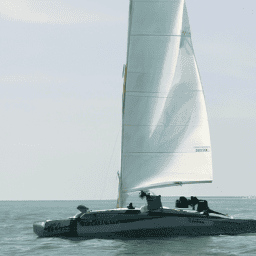} \\
        
        % x' (rescons)
        \raisebox{0.06\textwidth}{\rotatebox[origin=c]{0}{$x'$}} &
        \includegraphics[width=0.15\textwidth]{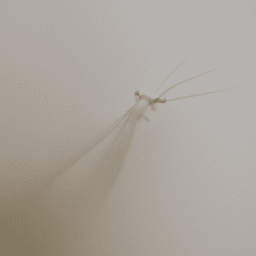} &
        \includegraphics[width=0.15\textwidth]{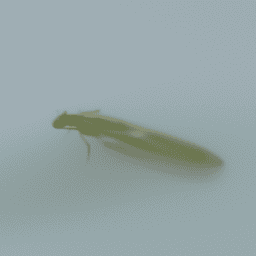} &
        \includegraphics[width=0.15\textwidth]{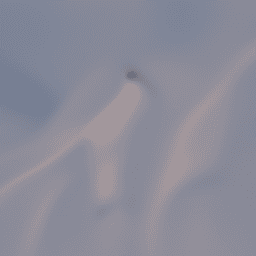} &
        \includegraphics[width=0.15\textwidth]{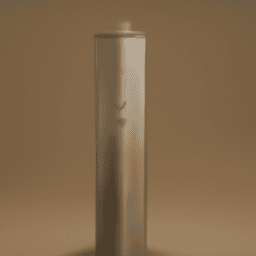} &
        \includegraphics[width=0.15\textwidth]{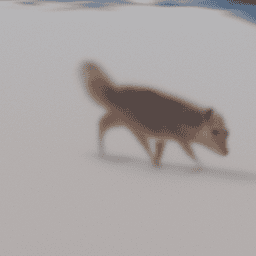} &
        \includegraphics[width=0.15\textwidth]{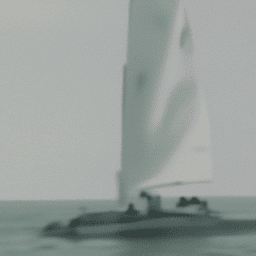} \\

        \raisebox{0.07\textwidth}{\rotatebox[origin=c]{0}{$\Delta x$}} &
        \includegraphics[width=0.15\textwidth]{sup_adm/001_dire.png} &
        \includegraphics[width=0.15\textwidth]{sup_adm/007_dire.png} &
        \includegraphics[width=0.15\textwidth]{sup_adm/009_dire.png} &
        \includegraphics[width=0.15\textwidth]{sup_adm/011_dire.png} &
        \includegraphics[width=0.15\textwidth]{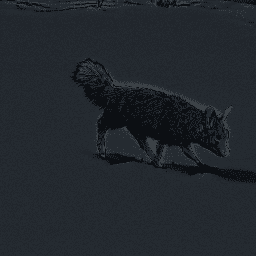} &
        \includegraphics[width=0.15\textwidth]{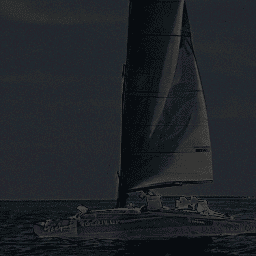} \\
        
        % x'' (rescons2)
        \raisebox{0.07\textwidth}{\rotatebox[origin=c]{0}{$x''$}} &
        \includegraphics[width=0.15\textwidth]{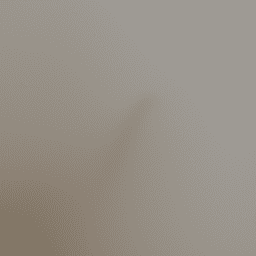} &
        \includegraphics[width=0.15\textwidth]{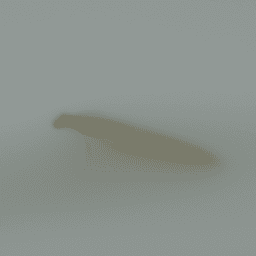} &
        \includegraphics[width=0.15\textwidth]{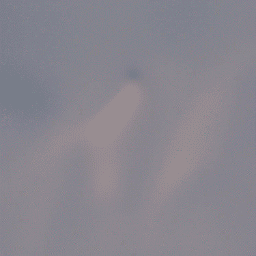} &
        \includegraphics[width=0.15\textwidth]{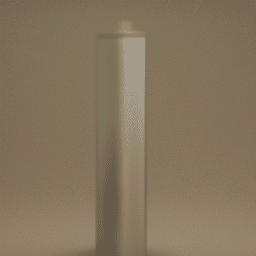} &
        \includegraphics[width=0.15\textwidth]{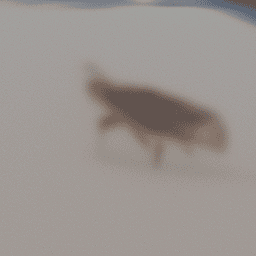} &
        \includegraphics[width=0.15\textwidth]{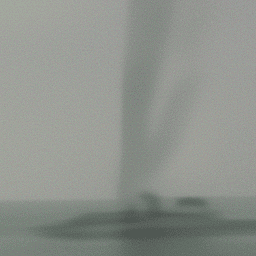} \\
        
        % Δx' (dire2)
        \raisebox{0.07\textwidth}{\rotatebox[origin=c]{0}{$\Delta x'$}} &
        \includegraphics[width=0.15\textwidth]{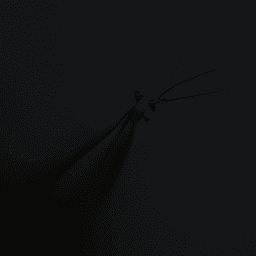} &
        \includegraphics[width=0.15\textwidth]{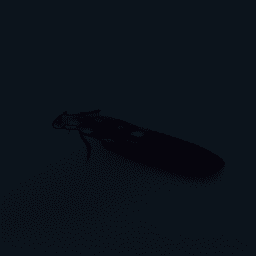} &
        \includegraphics[width=0.15\textwidth]{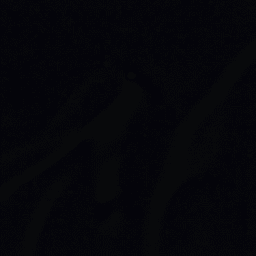} &
        \includegraphics[width=0.15\textwidth]{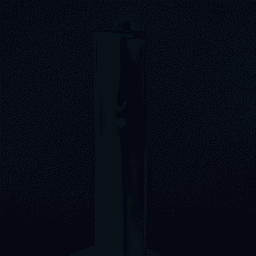} &
        \includegraphics[width=0.15\textwidth]{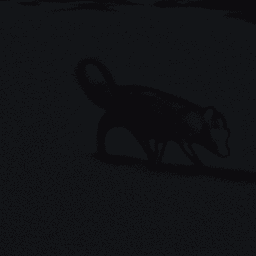} &
        \includegraphics[width=0.15\textwidth]{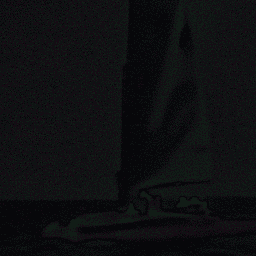} \\
        
        % Δ^2 x (diff)
        \raisebox{0.07\textwidth}{\rotatebox[origin=c]{0}{$\Delta^2 x$}} &
        \includegraphics[width=0.15\textwidth]{sup_adm/001_diff.png} &
        \includegraphics[width=0.15\textwidth]{sup_adm/007_diff.png} &
        \includegraphics[width=0.15\textwidth]{sup_adm/009_diff.png} &
        \includegraphics[width=0.15\textwidth]{sup_adm/011_diff.png} &
        \includegraphics[width=0.15\textwidth]{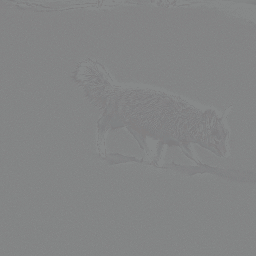} &
        \includegraphics[width=0.15\textwidth]{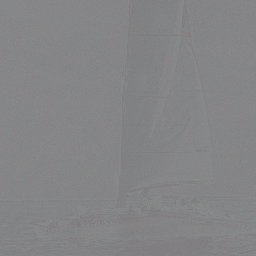} \\
    \end{tabular}
    \caption{
    \textbf{Visualization of ADM-generated fake images from the ImageNet subset constructed in the DIRE dataset.}
    For each ImageNet class (“A photo of class”, class from ImageNet~\cite{deng2009imagenet}), we show the ADM-generated image $x$, its first reconstruction $x'$, the second reconstruction $x''$, and their corresponding residual maps $\Delta x$, $\Delta x'$, and $\Delta^2 x$. 
    }
    \label{fig:sup_adm_matrix}
\end{figure*}

\begin{figure*}[t]
    \centering
    \setlength{\tabcolsep}{1pt}
    \renewcommand{\arraystretch}{0.1}
    \begin{tabular}{c@{\hspace{2pt}}cccccc}
        % x (real)
        \raisebox{0.07\textwidth}{\rotatebox[origin=c]{0}{$x$}} &
        \includegraphics[width=0.15\textwidth]{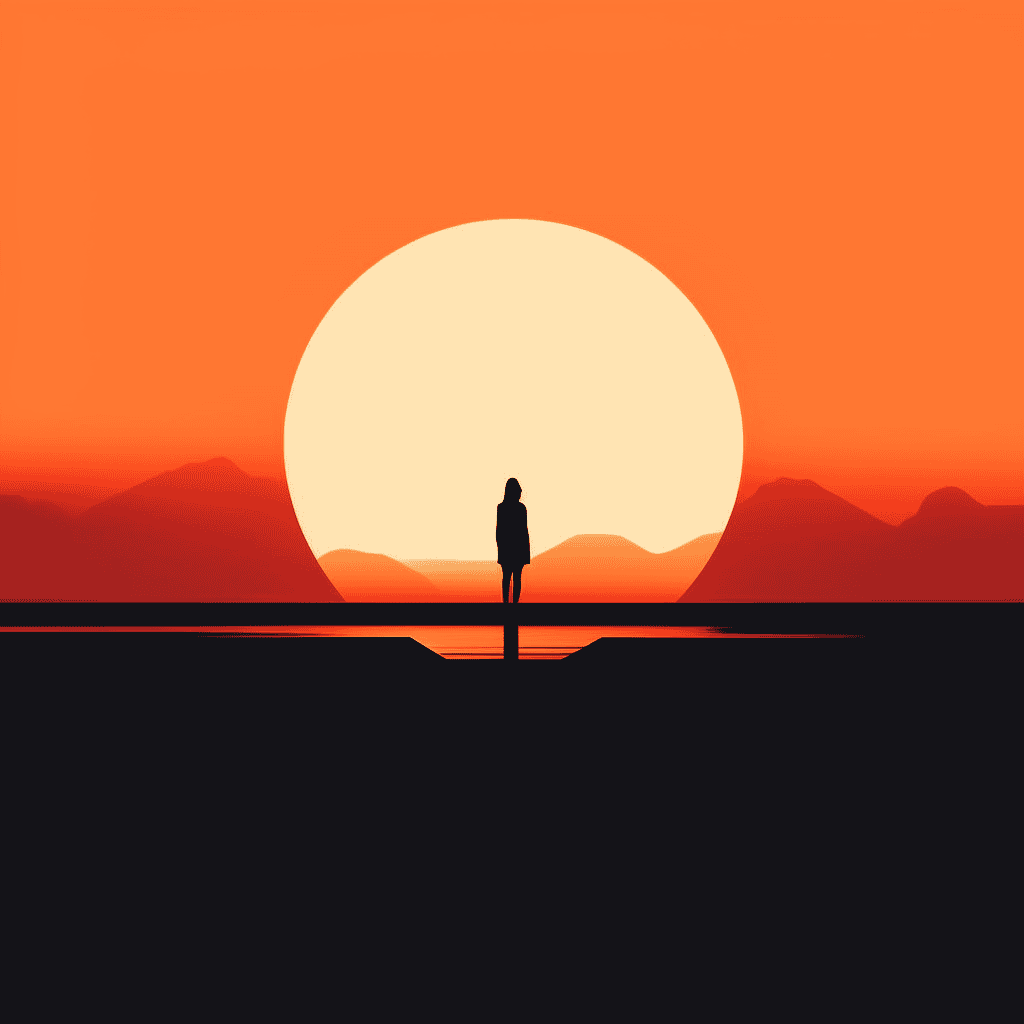} &
        \includegraphics[width=0.15\textwidth]{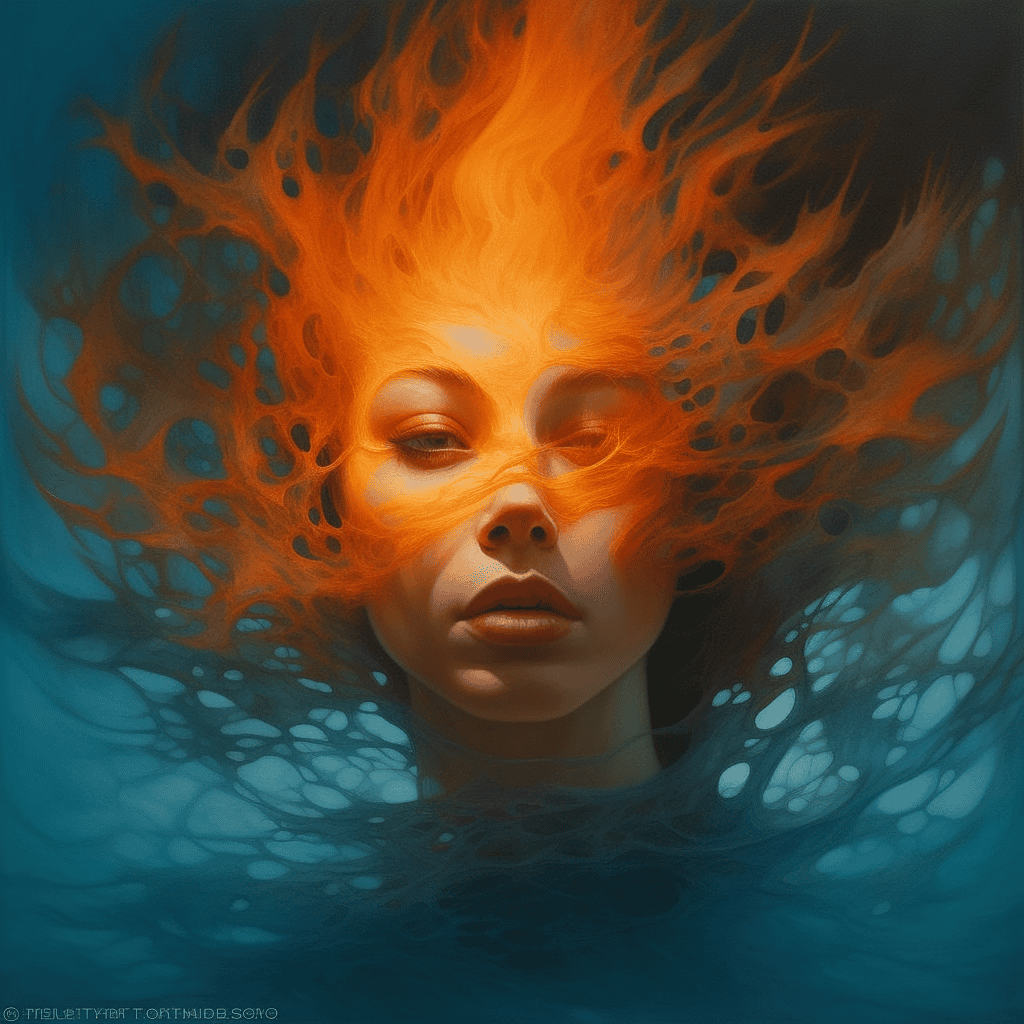} &
        \includegraphics[width=0.15\textwidth]{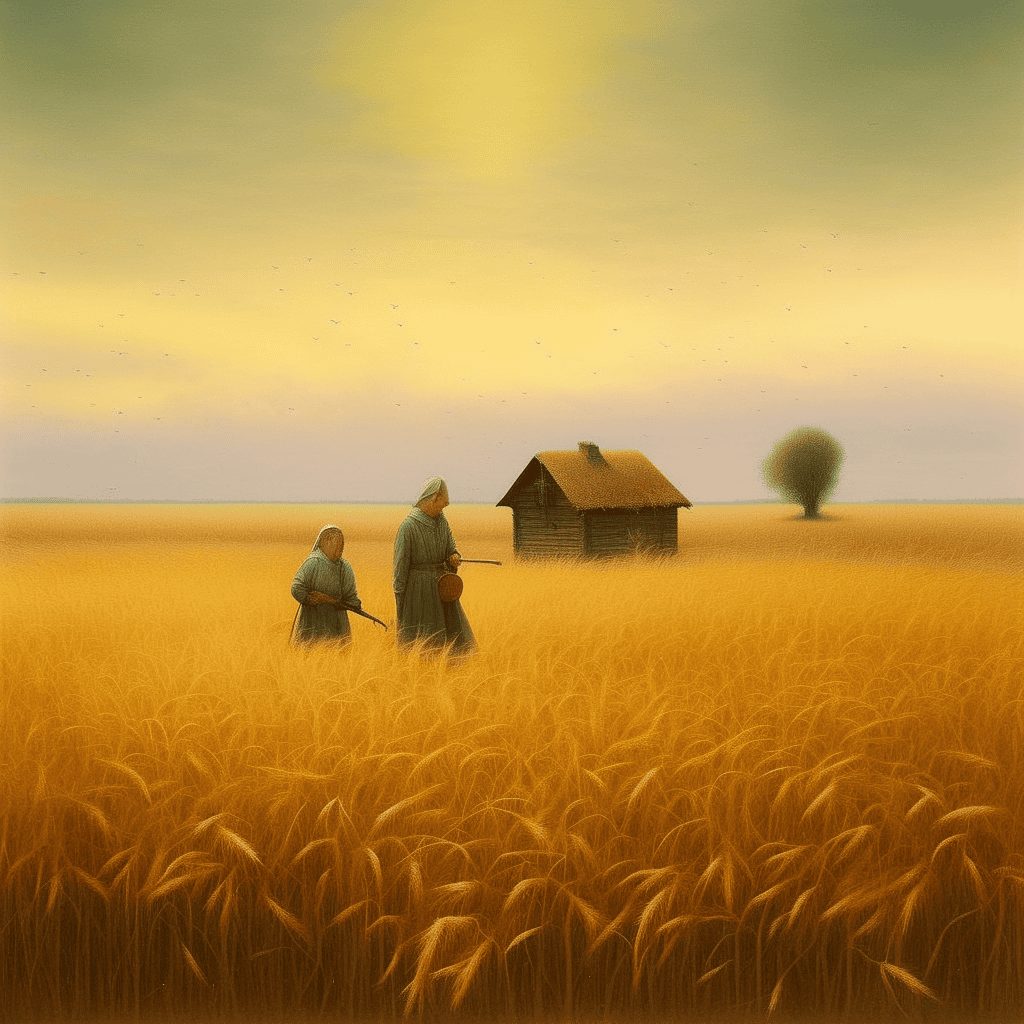} &
        \includegraphics[width=0.15\textwidth]{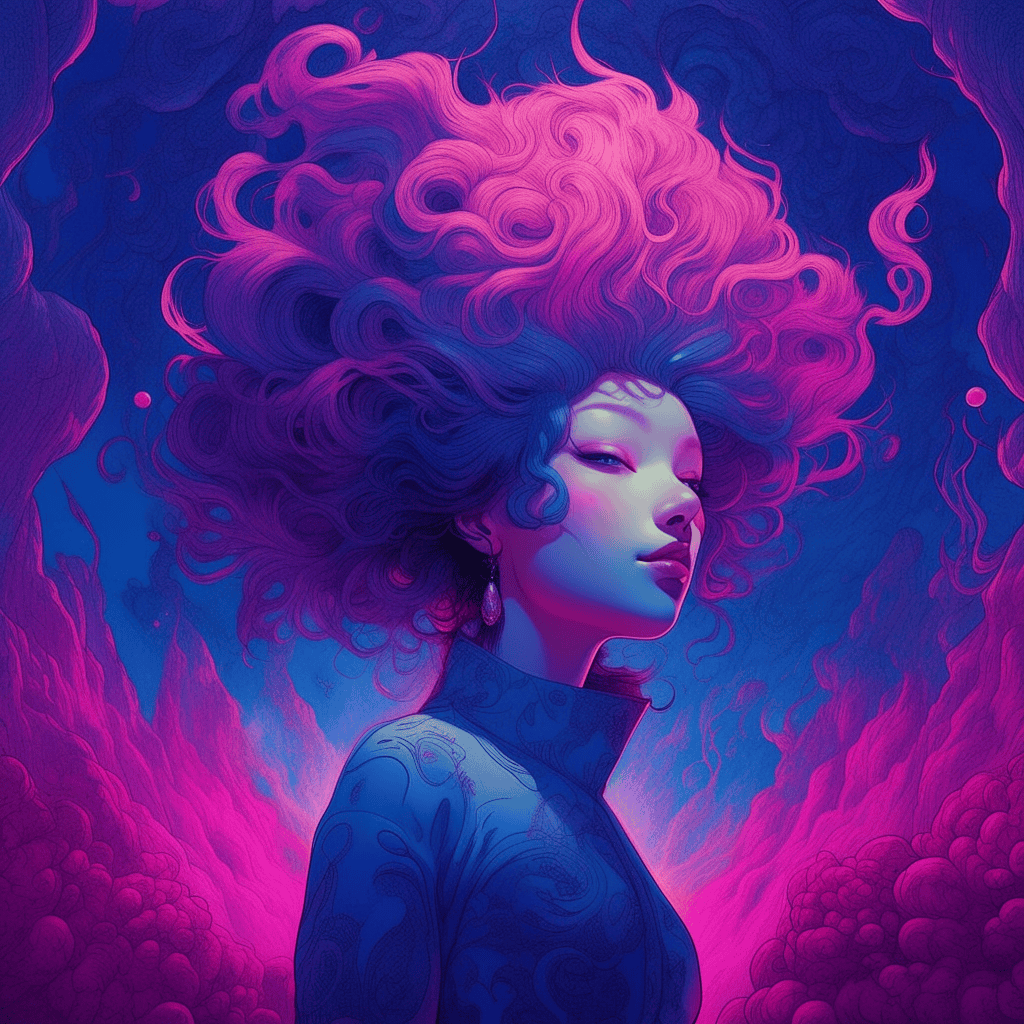} &
        \includegraphics[width=0.15\textwidth]{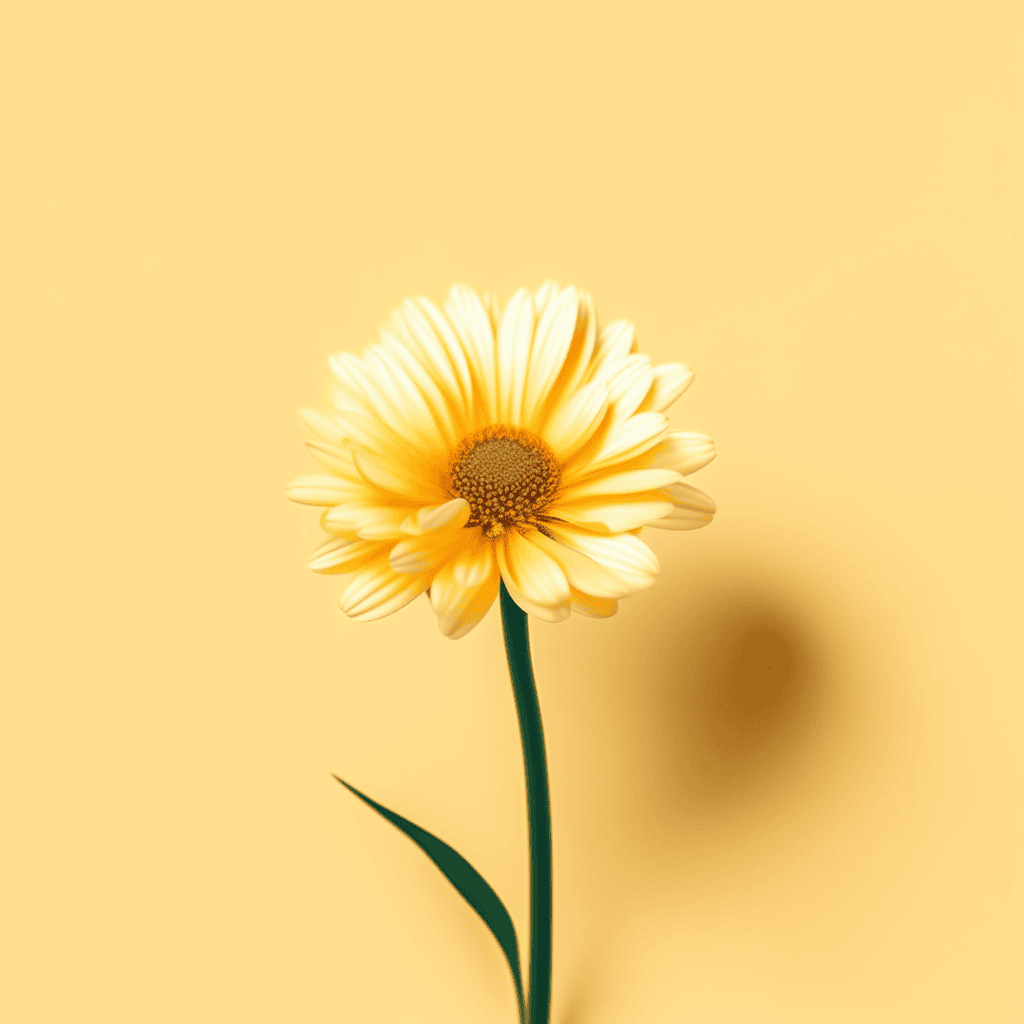} &
        \includegraphics[width=0.15\textwidth]{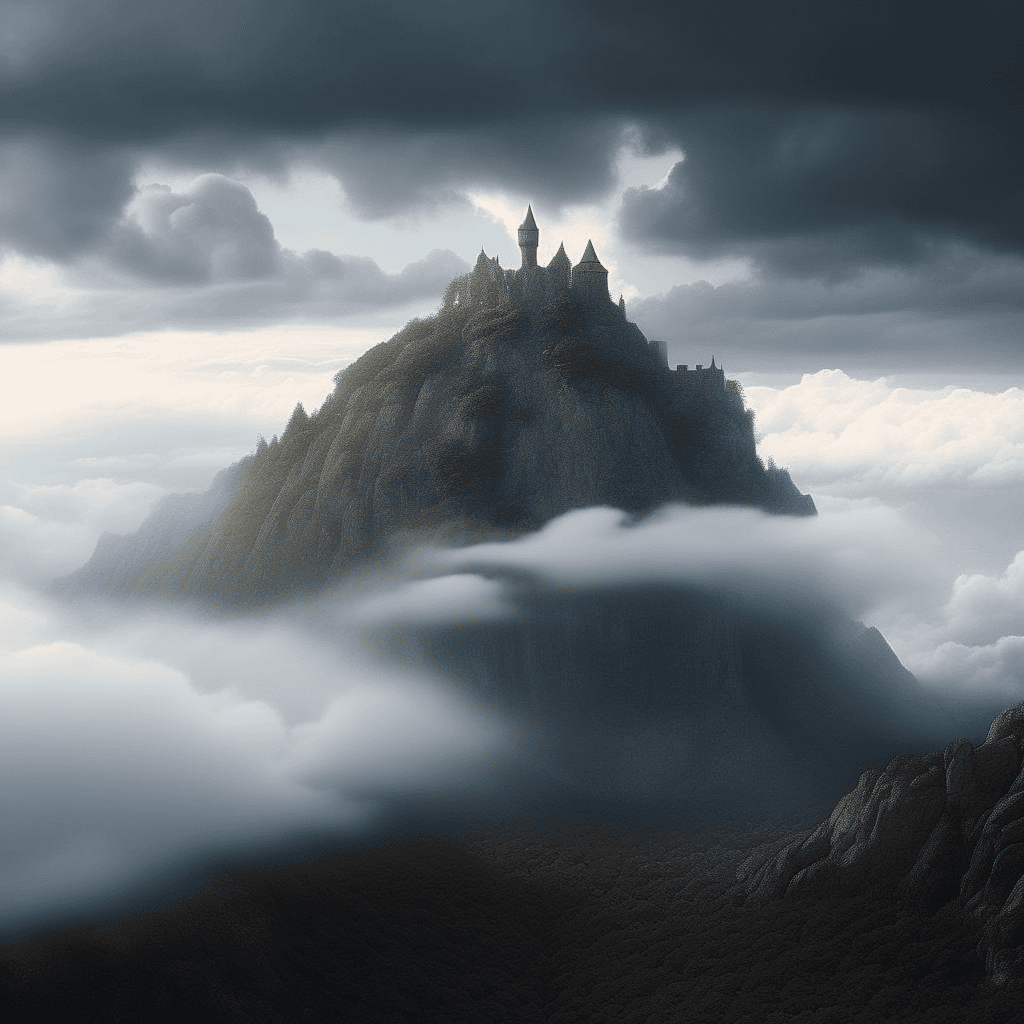} \\
        
        % x' (rescons)
        \raisebox{0.07\textwidth}{\rotatebox[origin=c]{0}{$x'$}} &
        \includegraphics[width=0.15\textwidth]{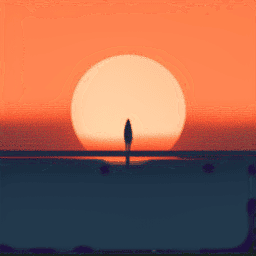} &
        \includegraphics[width=0.15\textwidth]{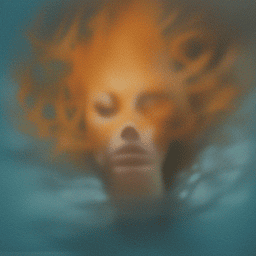} &
        \includegraphics[width=0.15\textwidth]{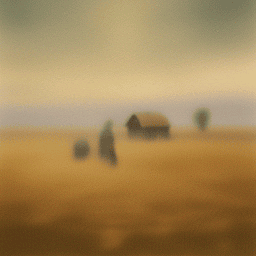} &
        \includegraphics[width=0.15\textwidth]{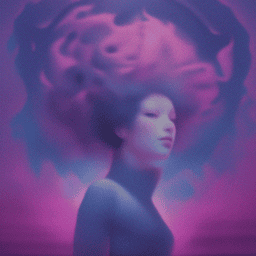} &
        \includegraphics[width=0.15\textwidth]{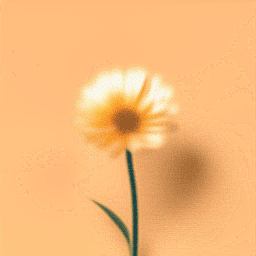} &
        \includegraphics[width=0.15\textwidth]{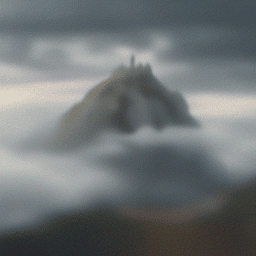} \\

        % Δx (dire)
        \raisebox{0.07\textwidth}{\rotatebox[origin=c]{0}{$\Delta x$}} &
        \includegraphics[width=0.15\textwidth]{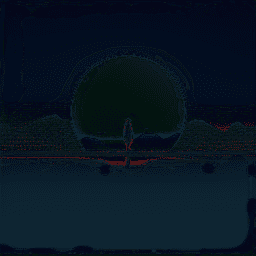} &
        \includegraphics[width=0.15\textwidth]{sup_kan3/019_dire.png} &
        \includegraphics[width=0.15\textwidth]{sup_kan3/020_dire.png} &
        \includegraphics[width=0.15\textwidth]{sup_kan3/028_dire.png} &
        \includegraphics[width=0.15\textwidth]{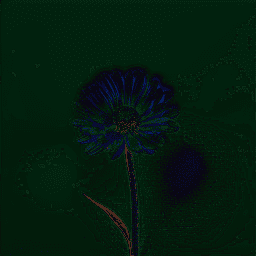} &
        \includegraphics[width=0.15\textwidth]{sup_kan3/071_dire.png} \\
        
        % x'' (rescons2)
        \raisebox{0.07\textwidth}{\rotatebox[origin=c]{0}{$x''$}} &
        \includegraphics[width=0.15\textwidth]{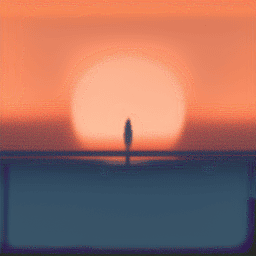} &
        \includegraphics[width=0.15\textwidth]{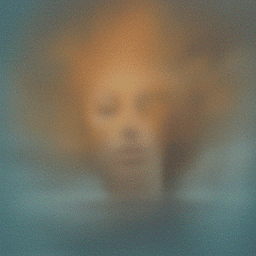} &
        \includegraphics[width=0.15\textwidth]{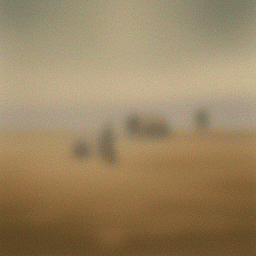} &
        \includegraphics[width=0.15\textwidth]{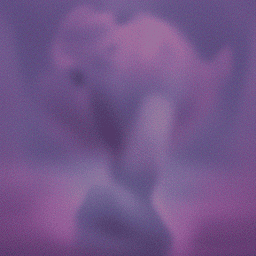} &
        \includegraphics[width=0.15\textwidth]{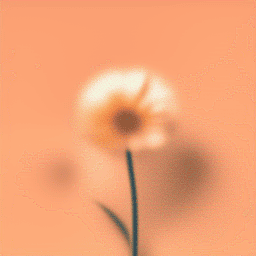} &
        \includegraphics[width=0.15\textwidth]{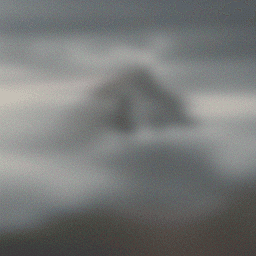} \\
        
        % Δx' (dire2)
        \raisebox{0.07\textwidth}{\rotatebox[origin=c]{0}{$\Delta x'$}} &
        \includegraphics[width=0.15\textwidth]{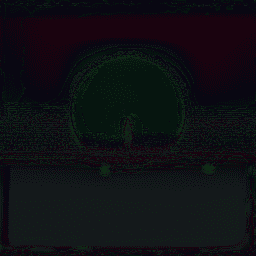} &
        \includegraphics[width=0.15\textwidth]{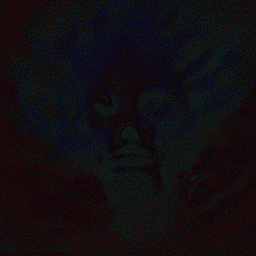} &
        \includegraphics[width=0.15\textwidth]{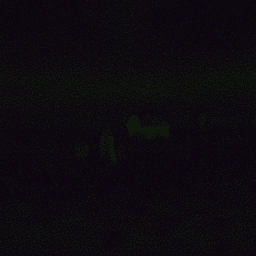} &
        \includegraphics[width=0.15\textwidth]{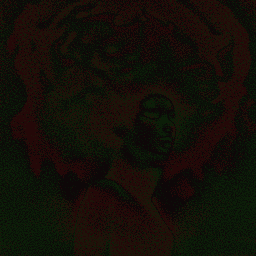} &
        \includegraphics[width=0.15\textwidth]{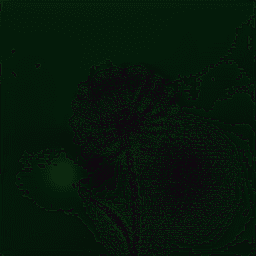} &
        \includegraphics[width=0.15\textwidth]{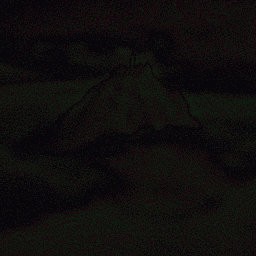} \\
        
        % Δ^2 x (diff)
        \raisebox{0.07\textwidth}{\rotatebox[origin=c]{0}{$\Delta^2 x$}} &
        \includegraphics[width=0.15\textwidth]{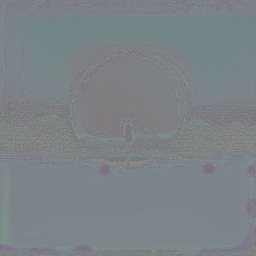} &
        \includegraphics[width=0.15\textwidth]{sup_kan3/019_diff.png} &
        \includegraphics[width=0.15\textwidth]{sup_kan3/020_diff.png} &
        \includegraphics[width=0.15\textwidth]{sup_kan3/028_diff.png} &
        \includegraphics[width=0.15\textwidth]{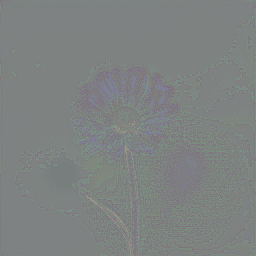} &
        \includegraphics[width=0.15\textwidth]{sup_kan3/071_diff.png} \\
    \end{tabular}
    \caption{
    \textbf{Visualization of Kan3-generated images constructed for our dataset.}
    For each sample, we show the original synthesized image $x$, the first reconstruction $x'$, the second reconstruction $x''$, and their corresponding residual maps $\Delta x$, $\Delta x'$, and $\Delta^2 x$. 
    All Kan3 images are generated using prompts collected from Midjourney. 
    }
    \label{fig:sup_kan3_matrix}
\end{figure*}

\begin{figure*}[t]
    \centering
    \setlength{\tabcolsep}{1pt}
    \renewcommand{\arraystretch}{0.1}
    \begin{tabular}{c@{\hspace{2pt}}cccccc}
        % x (real)
        \raisebox{0.07\textwidth}{\rotatebox[origin=c]{0}{$x$}} &
        \includegraphics[width=0.15\textwidth]{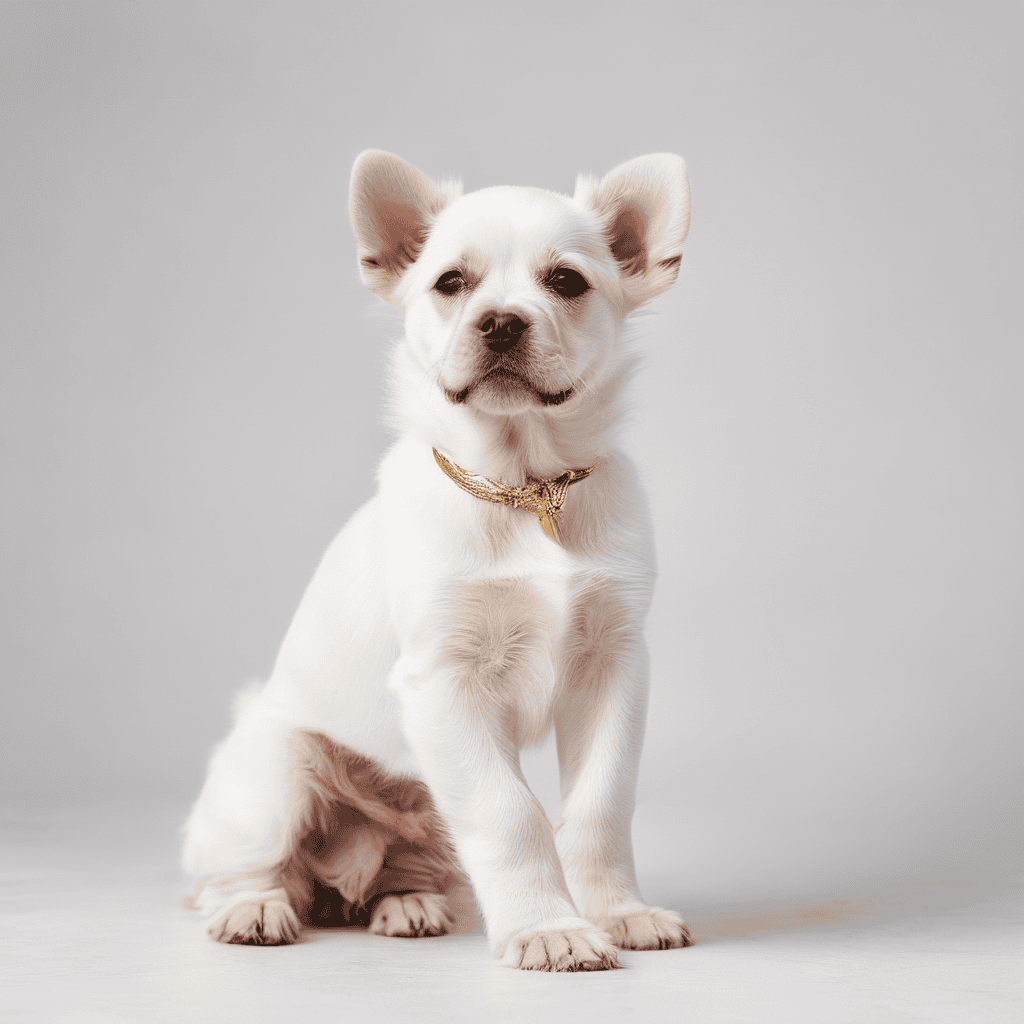} &
        \includegraphics[width=0.15\textwidth]{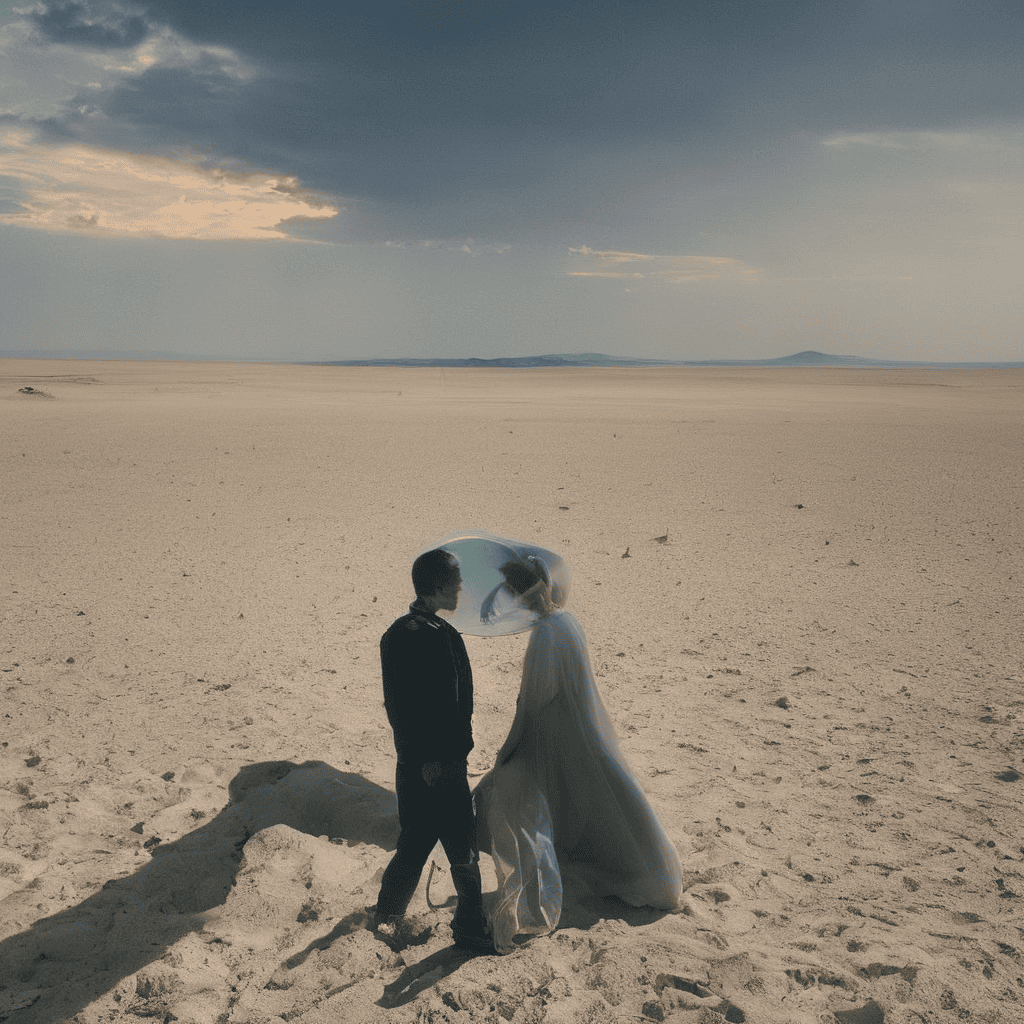} &
        \includegraphics[width=0.15\textwidth]{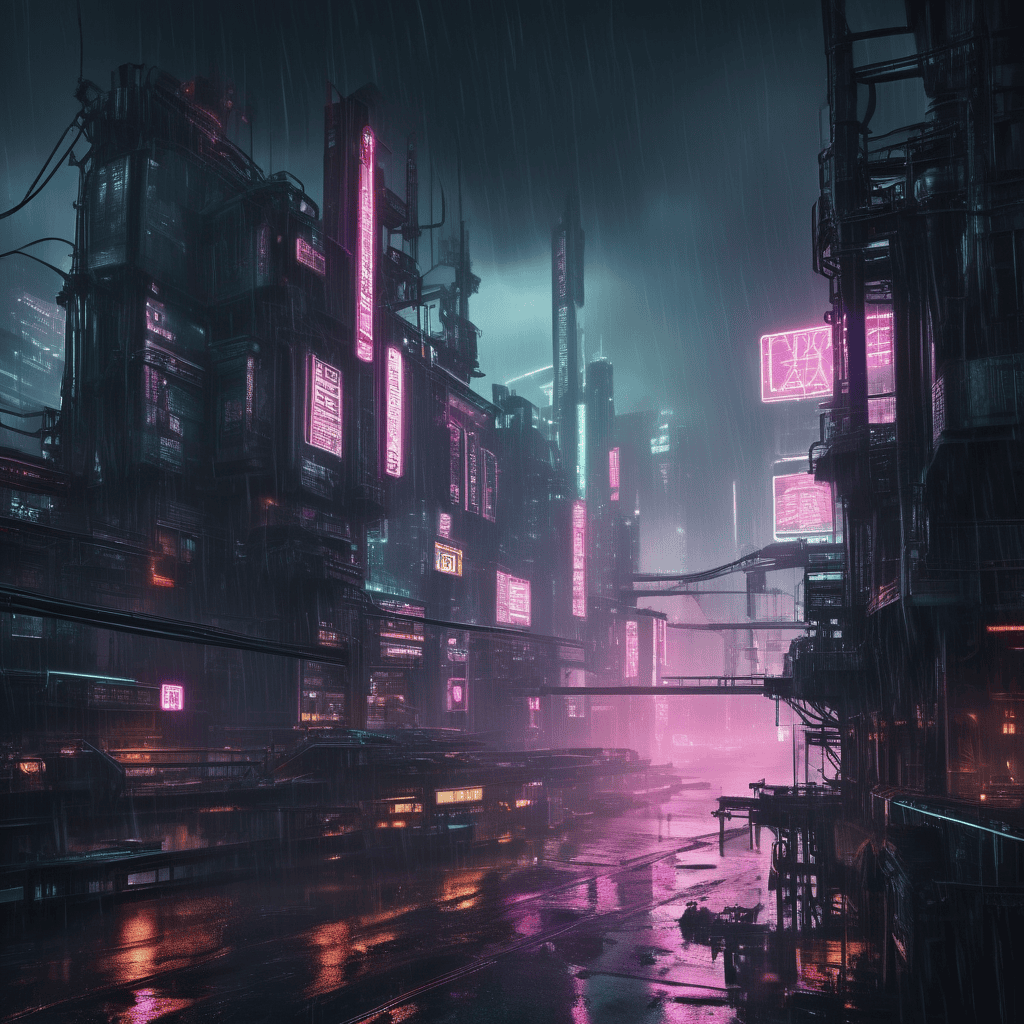} &
        \includegraphics[width=0.15\textwidth]{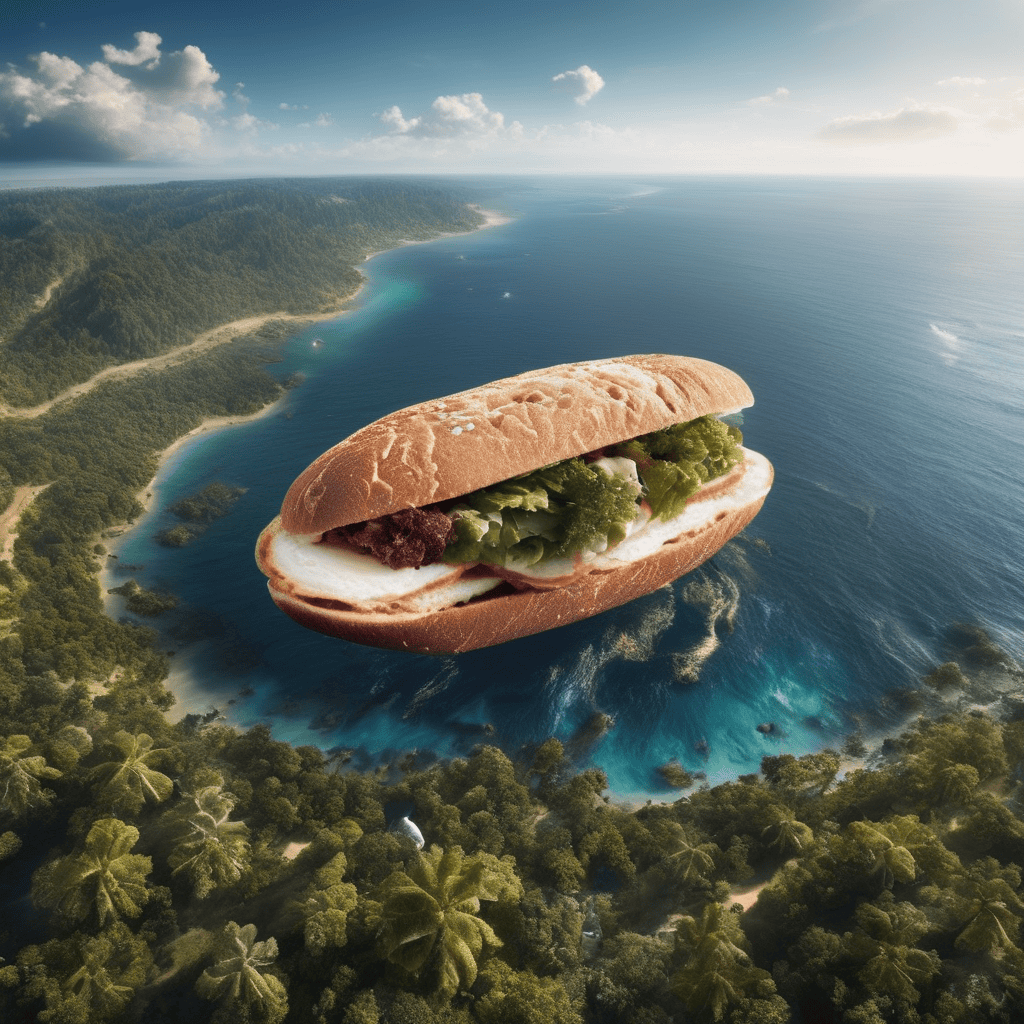} &
        \includegraphics[width=0.15\textwidth]{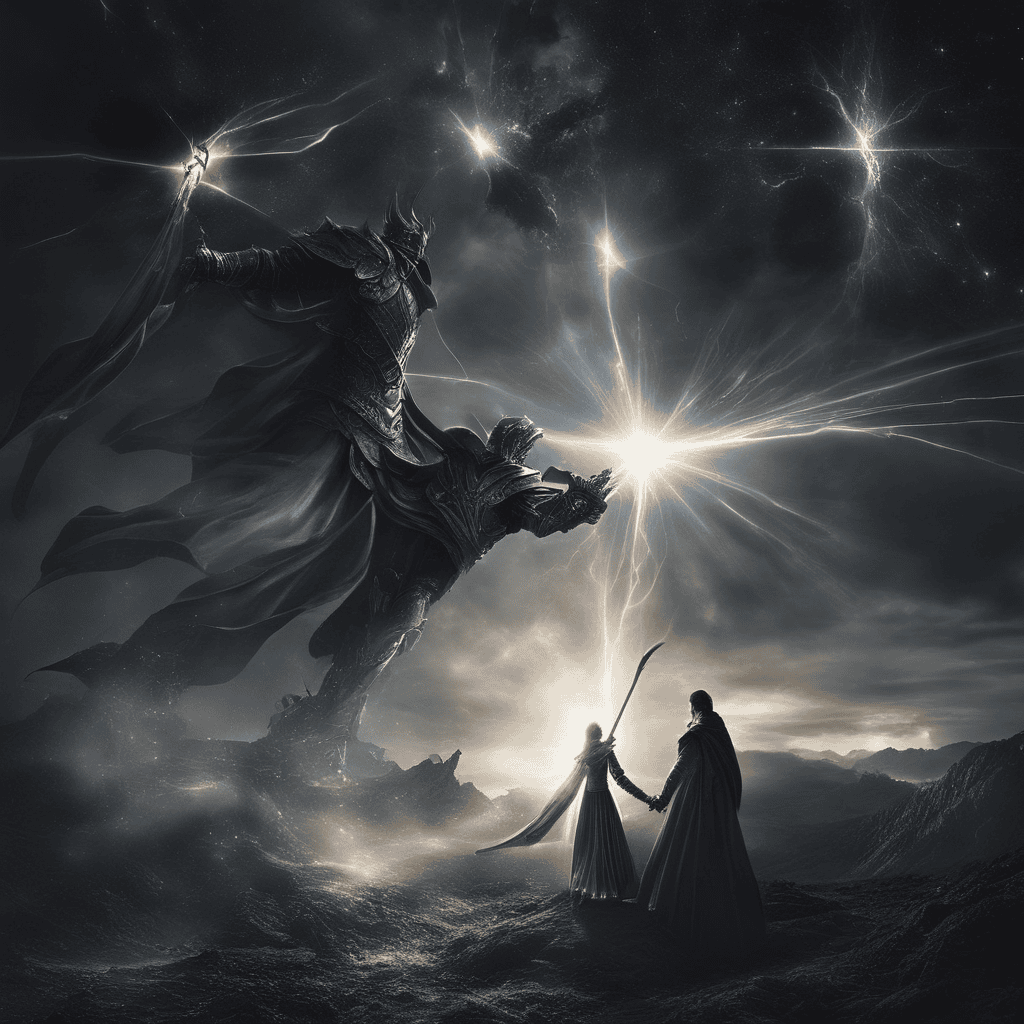} &
        \includegraphics[width=0.15\textwidth]{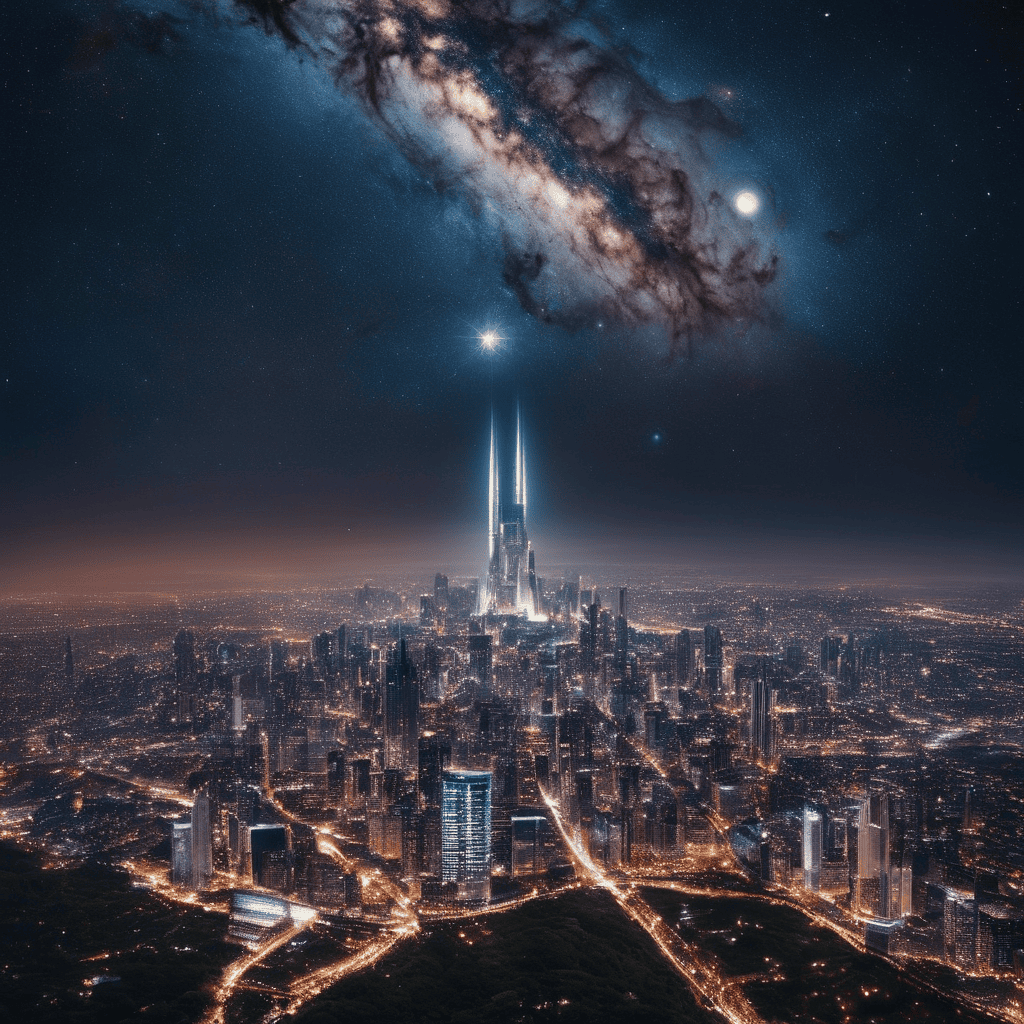} \\

        % x' (rescons)
        \raisebox{0.07\textwidth}{\rotatebox[origin=c]{0}{$x'$}} &
        \includegraphics[width=0.15\textwidth]{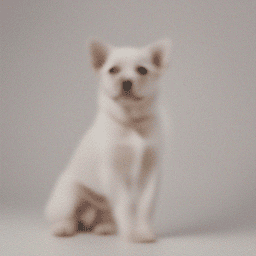} &
        \includegraphics[width=0.15\textwidth]{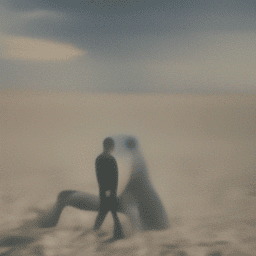} &
        \includegraphics[width=0.15\textwidth]{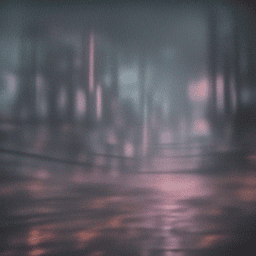} &
        \includegraphics[width=0.15\textwidth]{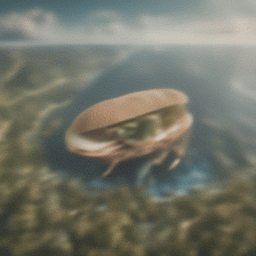} &
        \includegraphics[width=0.15\textwidth]{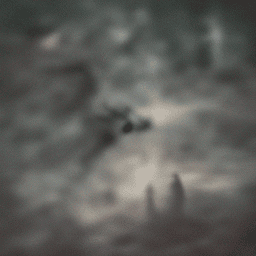} &
        \includegraphics[width=0.15\textwidth]{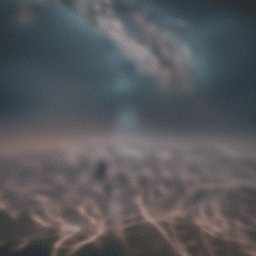} \\

        % Δx (dire)
        \raisebox{0.07\textwidth}{\rotatebox[origin=c]{0}{$\Delta x$}} &
        \includegraphics[width=0.15\textwidth]{sup_vega/002_dire.png} &
        \includegraphics[width=0.15\textwidth]{sup_vega/009_dire.png} &
        \includegraphics[width=0.15\textwidth]{sup_vega/059_dire.png} &
        \includegraphics[width=0.15\textwidth]{sup_vega/062_dire.png} &
        \includegraphics[width=0.15\textwidth]{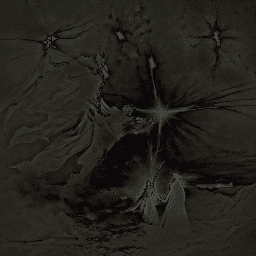} &
        \includegraphics[width=0.15\textwidth]{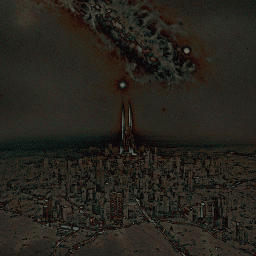} \\
        
        % x'' (rescons2)
        \raisebox{0.07\textwidth}{\rotatebox[origin=c]{0}{$x''$}} &
        \includegraphics[width=0.15\textwidth]{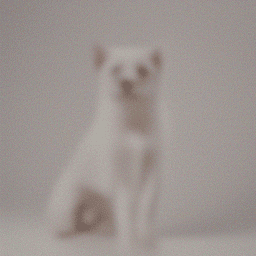} &
        \includegraphics[width=0.15\textwidth]{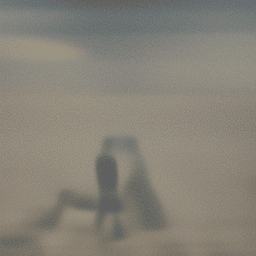} &
        \includegraphics[width=0.15\textwidth]{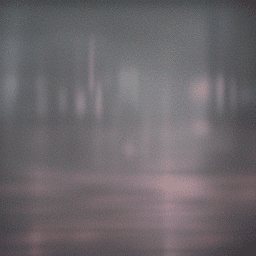} &
        \includegraphics[width=0.15\textwidth]{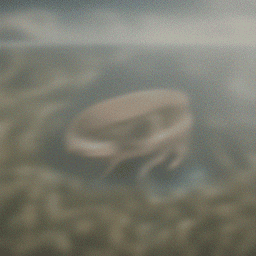} &
        \includegraphics[width=0.15\textwidth]{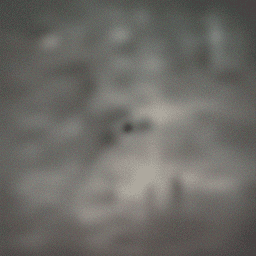} &
        \includegraphics[width=0.15\textwidth]{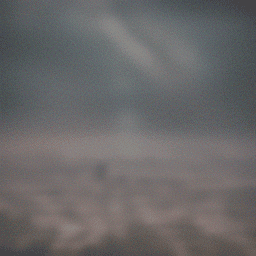} \\
        
        % Δx' (dire2)
        \raisebox{0.07\textwidth}{\rotatebox[origin=c]{0}{$\Delta x'$}} &
        \includegraphics[width=0.15\textwidth]{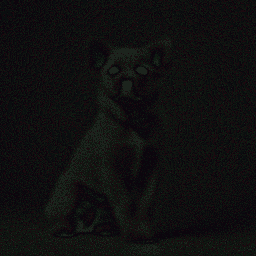} &
        \includegraphics[width=0.15\textwidth]{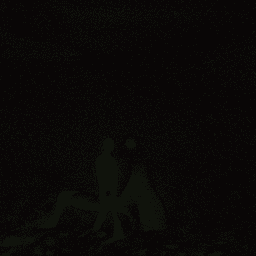} &
        \includegraphics[width=0.15\textwidth]{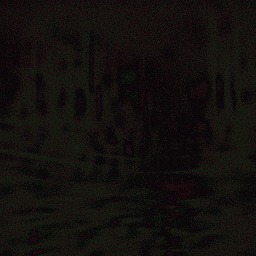} &
        \includegraphics[width=0.15\textwidth]{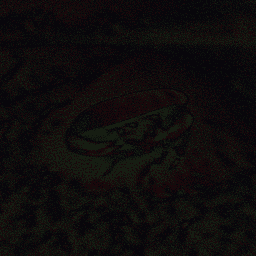} &
        \includegraphics[width=0.15\textwidth]{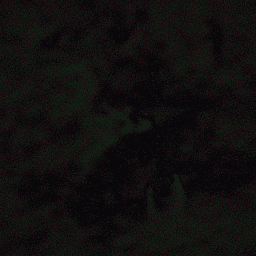} &
        \includegraphics[width=0.15\textwidth]{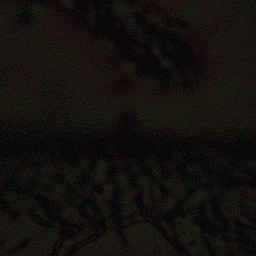} \\
        
        % Δ^2 x (diff)
        \raisebox{0.07\textwidth}{\rotatebox[origin=c]{0}{$\Delta^2 x$}} &
        \includegraphics[width=0.15\textwidth]{sup_vega/002_diff.png} &
        \includegraphics[width=0.15\textwidth]{sup_vega/009_diff.png} &
        \includegraphics[width=0.15\textwidth]{sup_vega/059_diff.png} &
        \includegraphics[width=0.15\textwidth]{sup_vega/062_diff.png} &
        \includegraphics[width=0.15\textwidth]{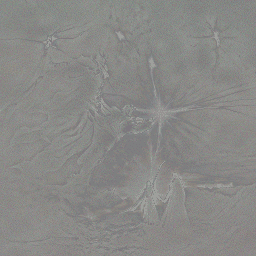} &
        \includegraphics[width=0.15\textwidth]{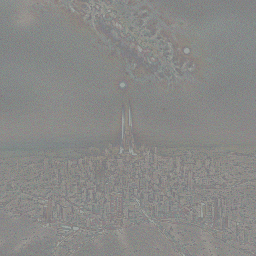} \\
    \end{tabular}
    \caption{
    \textbf{Visualization of Vega-generated images constructed for our dataset.}
    For each sample, we show the original synthesized image $x$, the first reconstruction $x'$, the second reconstruction $x''$, and their corresponding residual maps $\Delta x$, $\Delta x'$, and $\Delta^2 x$. 
    All Vega images are generated using prompts collected from Midjourney. 
    }
    \label{fig:sup_vega_matrix}
\end{figure*}

\begin{figure*}[t]
    \centering
    \setlength{\tabcolsep}{1pt}
    \renewcommand{\arraystretch}{0.1}
    \begin{tabular}{c@{\hspace{2pt}}cccccc}
        % x (real)
        \raisebox{0.07\textwidth}{\rotatebox[origin=c]{0}{$x$}} &
        \includegraphics[width=0.15\textwidth]{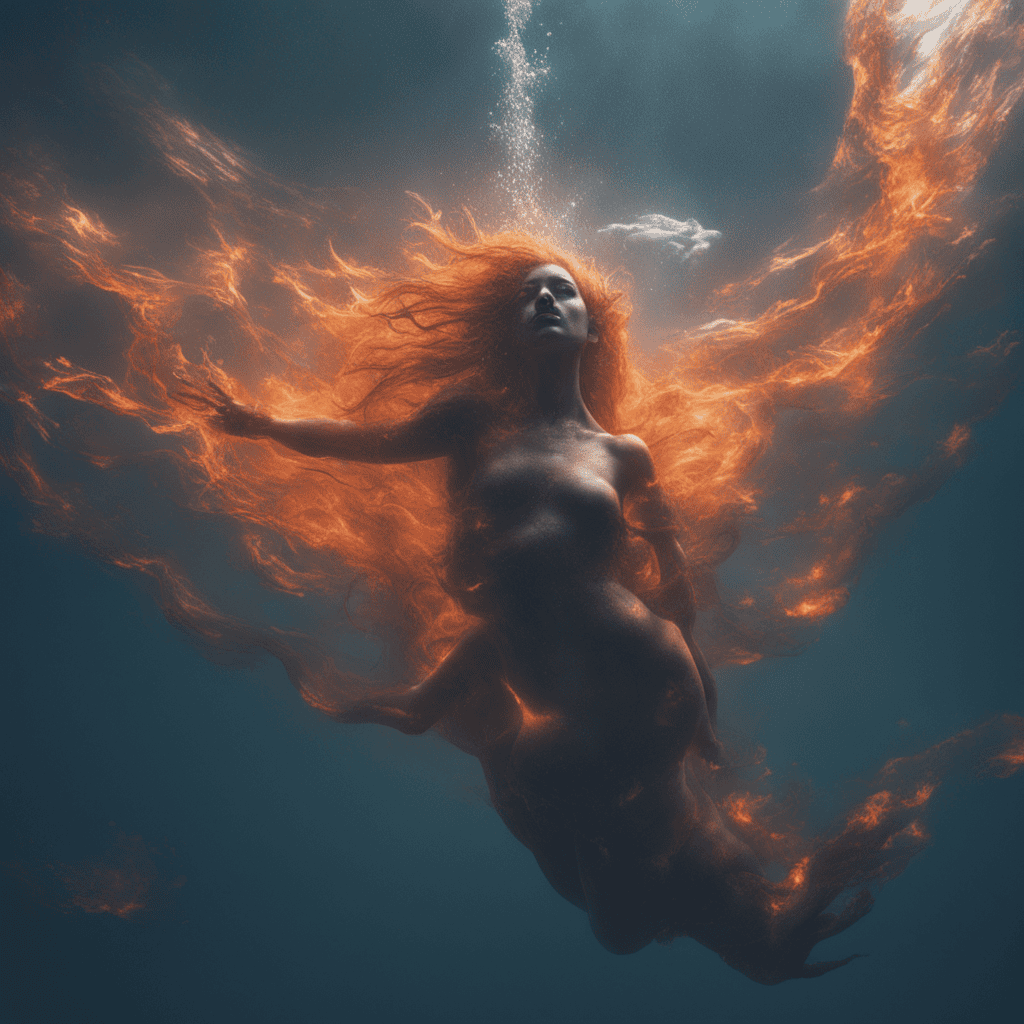} &
        \includegraphics[width=0.15\textwidth]{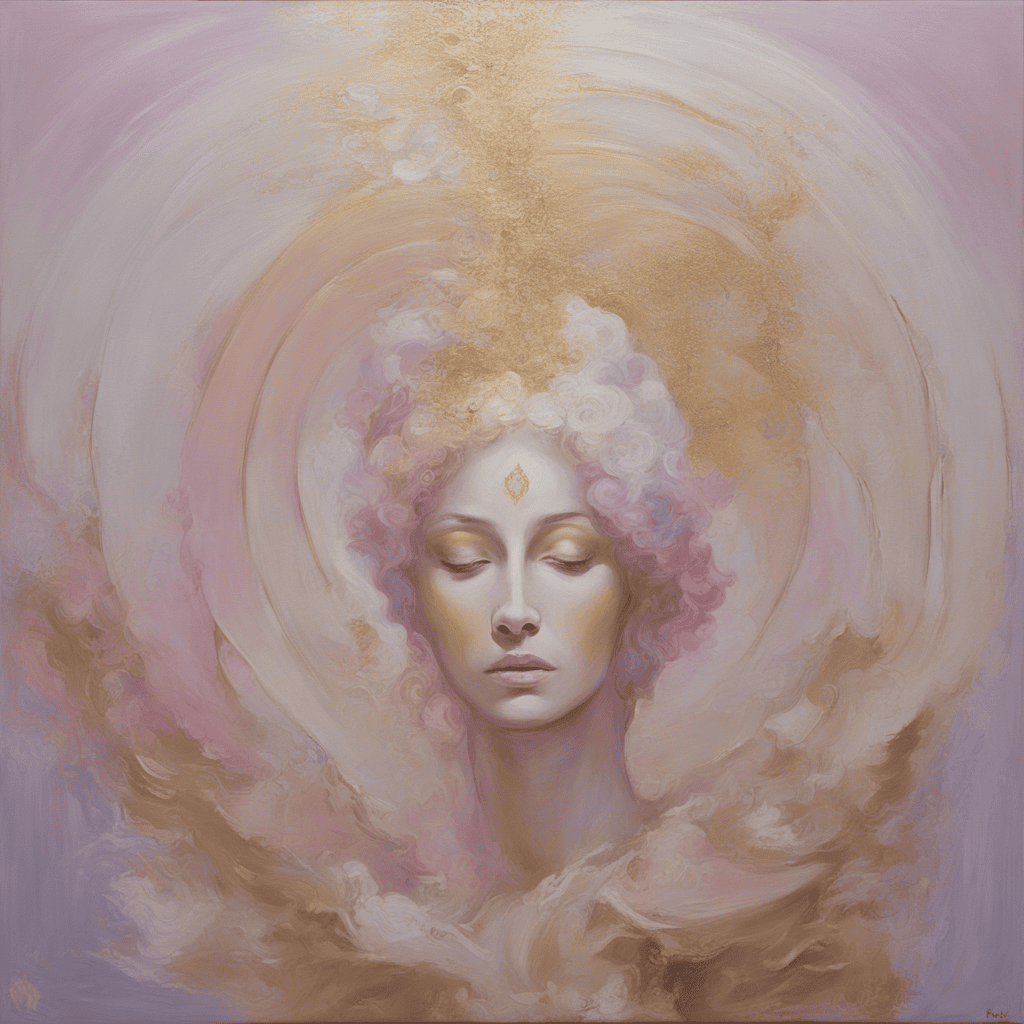} &
        \includegraphics[width=0.15\textwidth]{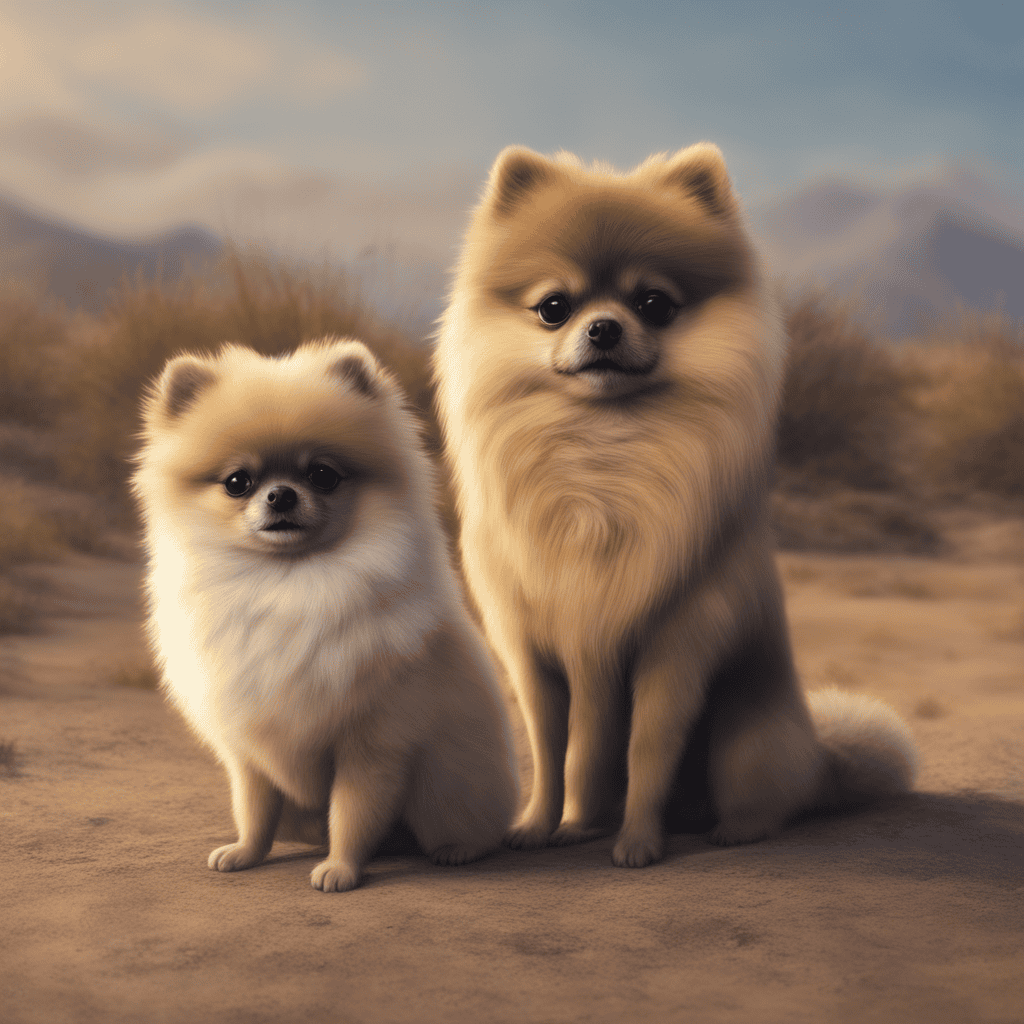} &
        \includegraphics[width=0.15\textwidth]{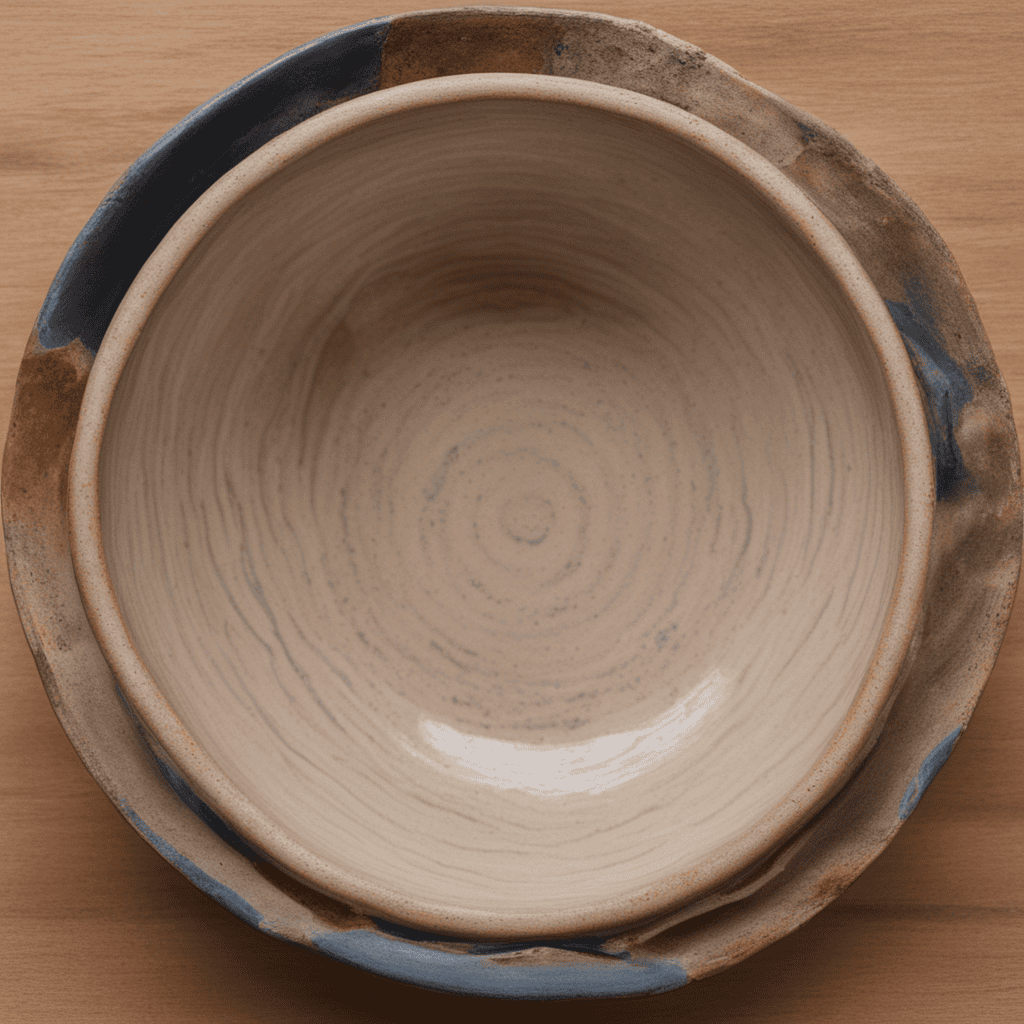} &
        \includegraphics[width=0.15\textwidth]{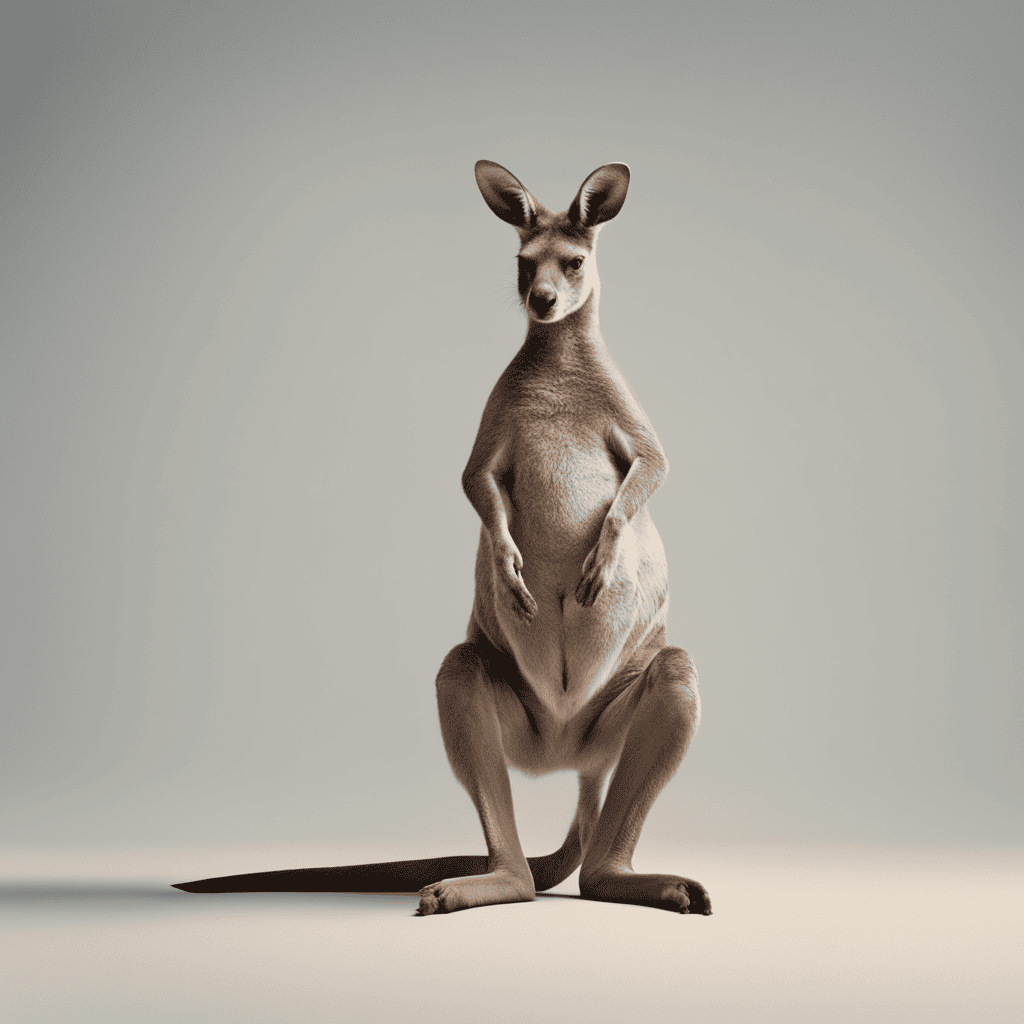} &
        \includegraphics[width=0.15\textwidth]{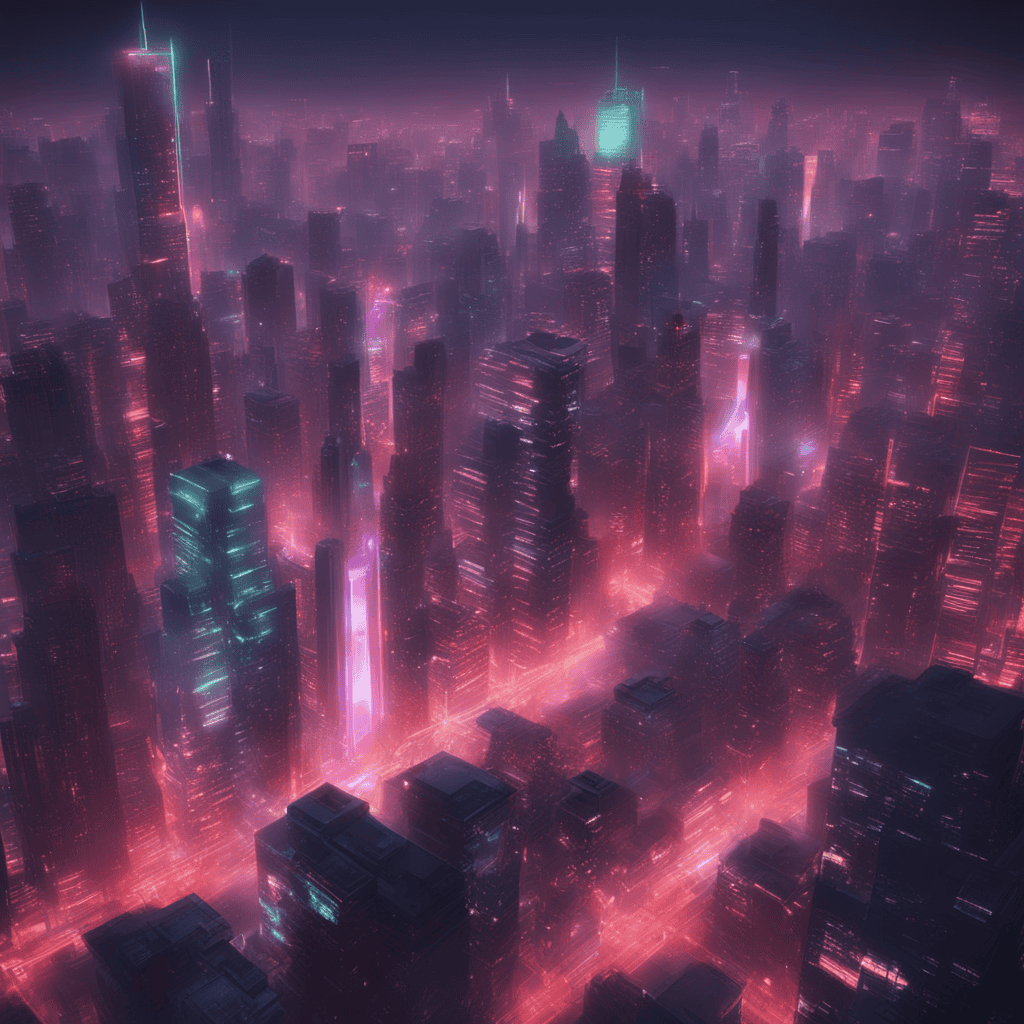} \\

        % x' (rescons)
        \raisebox{0.07\textwidth}{\rotatebox[origin=c]{0}{$x'$}} &
        \includegraphics[width=0.15\textwidth]{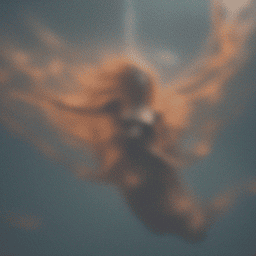} &
        \includegraphics[width=0.15\textwidth]{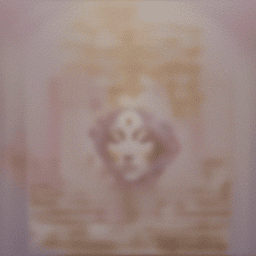} &
        \includegraphics[width=0.15\textwidth]{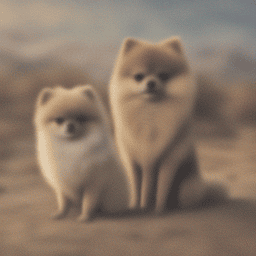} &
        \includegraphics[width=0.15\textwidth]{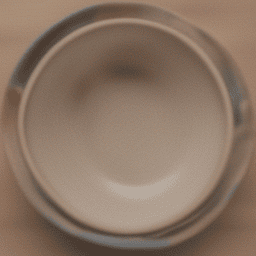} &
        \includegraphics[width=0.15\textwidth]{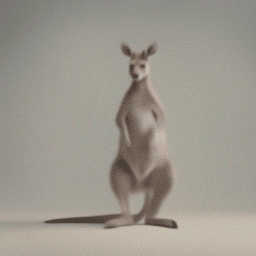} &
        \includegraphics[width=0.15\textwidth]{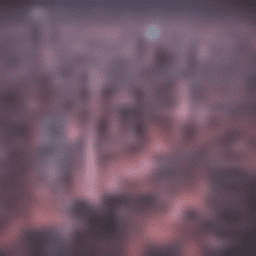} \\

                % Δx (dire)
        \raisebox{0.07\textwidth}{\rotatebox[origin=c]{0}{$\Delta x$}} &
        \includegraphics[width=0.15\textwidth]{sup_sdxl/030_dire.png} &
        \includegraphics[width=0.15\textwidth]{sup_sdxl/032_dire.png} &
        \includegraphics[width=0.15\textwidth]{sup_sdxl/046_dire.png} &
        \includegraphics[width=0.15\textwidth]{sup_sdxl/053_dire.png} &
        \includegraphics[width=0.15\textwidth]{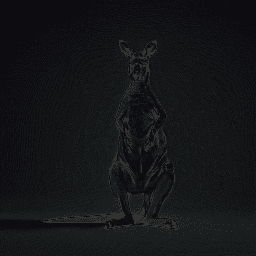} &
        \includegraphics[width=0.15\textwidth]{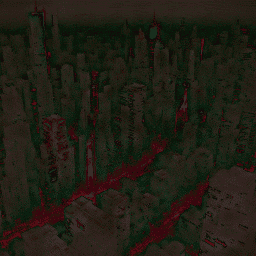} \\
        
        % x'' (rescons2)
        \raisebox{0.07\textwidth}{\rotatebox[origin=c]{0}{$x''$}} &
        \includegraphics[width=0.15\textwidth]{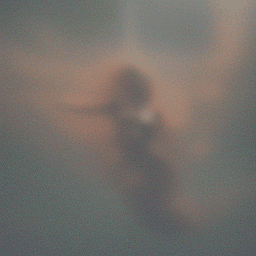} &
        \includegraphics[width=0.15\textwidth]{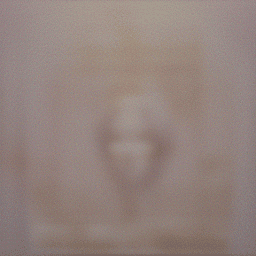} &
        \includegraphics[width=0.15\textwidth]{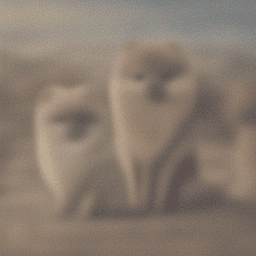} &
        \includegraphics[width=0.15\textwidth]{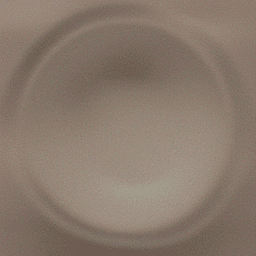} &
        \includegraphics[width=0.15\textwidth]{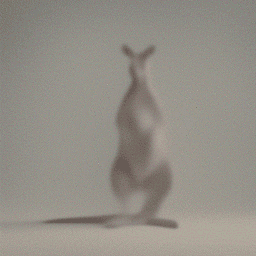} &
        \includegraphics[width=0.15\textwidth]{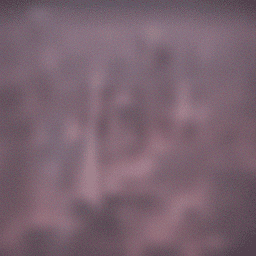} \\
        
        % Δx' (dire2)
        \raisebox{0.07\textwidth}{\rotatebox[origin=c]{0}{$\Delta x'$}} &
        \includegraphics[width=0.15\textwidth]{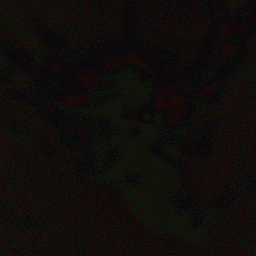} &
        \includegraphics[width=0.15\textwidth]{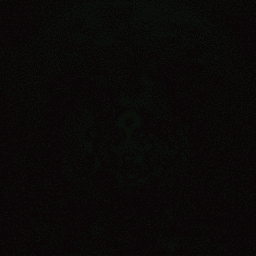} &
        \includegraphics[width=0.15\textwidth]{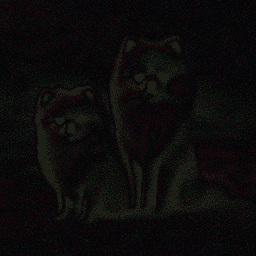} &
        \includegraphics[width=0.15\textwidth]{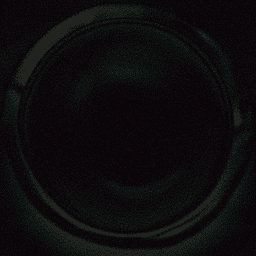} &
        \includegraphics[width=0.15\textwidth]{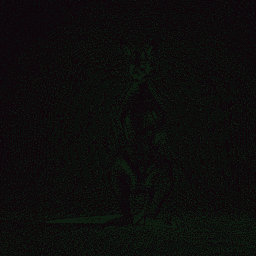} &
        \includegraphics[width=0.15\textwidth]{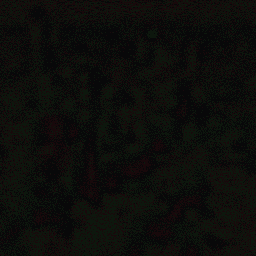} \\
        
        % Δ^2 x (diff)
        \raisebox{0.07\textwidth}{\rotatebox[origin=c]{0}{$\Delta^2 x$}} &
        \includegraphics[width=0.15\textwidth]{sup_sdxl/030_diff.png} &
        \includegraphics[width=0.15\textwidth]{sup_sdxl/032_diff.png} &
        \includegraphics[width=0.15\textwidth]{sup_sdxl/046_diff.png} &
        \includegraphics[width=0.15\textwidth]{sup_sdxl/053_diff.png} &
        \includegraphics[width=0.15\textwidth]{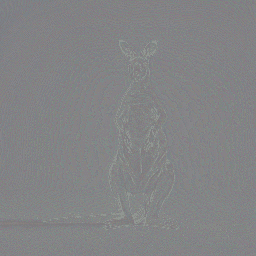} &
        \includegraphics[width=0.15\textwidth]{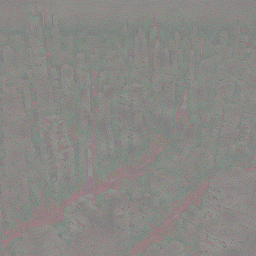} \\
    \end{tabular}
    \caption{    
    \textbf{Visualization of SDXL-generated images constructed for our dataset.}
    For each sample, we show the original synthesized image $x$, the first reconstruction $x'$, the second reconstruction $x''$, and their corresponding residual maps $\Delta x$, $\Delta x'$, and $\Delta^2 x$. 
    All SDXL images are generated using prompts collected from Midjourney. 
    }
    \label{fig:sup_sdxl_matrix}
\end{figure*}

\begin{figure*}[t]
    \centering
    \setlength{\tabcolsep}{1pt}
    \renewcommand{\arraystretch}{0.1}
    \begin{tabular}{c@{\hspace{2pt}}cccccc}
        % x (real)
        \raisebox{0.07\textwidth}{\rotatebox[origin=c]{0}{$x$}} &
        \includegraphics[width=0.15\textwidth]{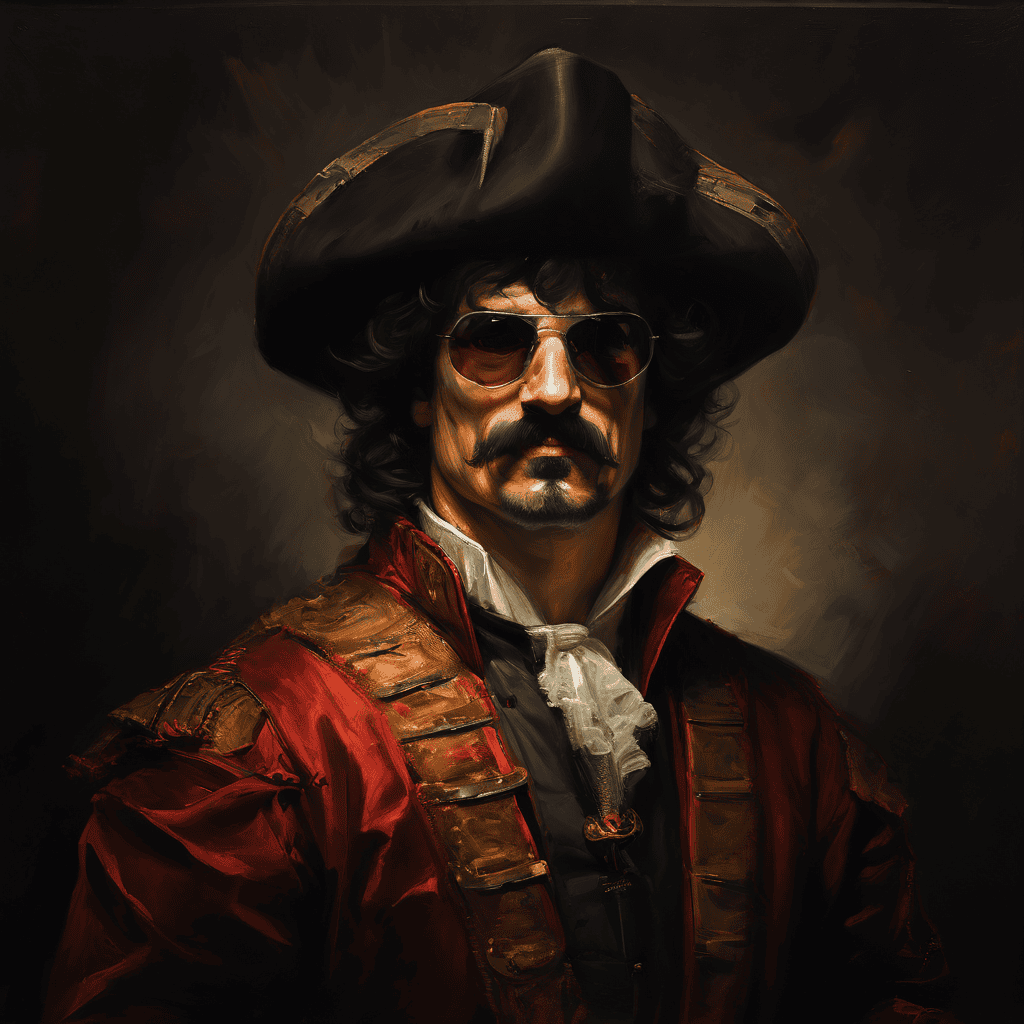} &
        \includegraphics[width=0.15\textwidth]{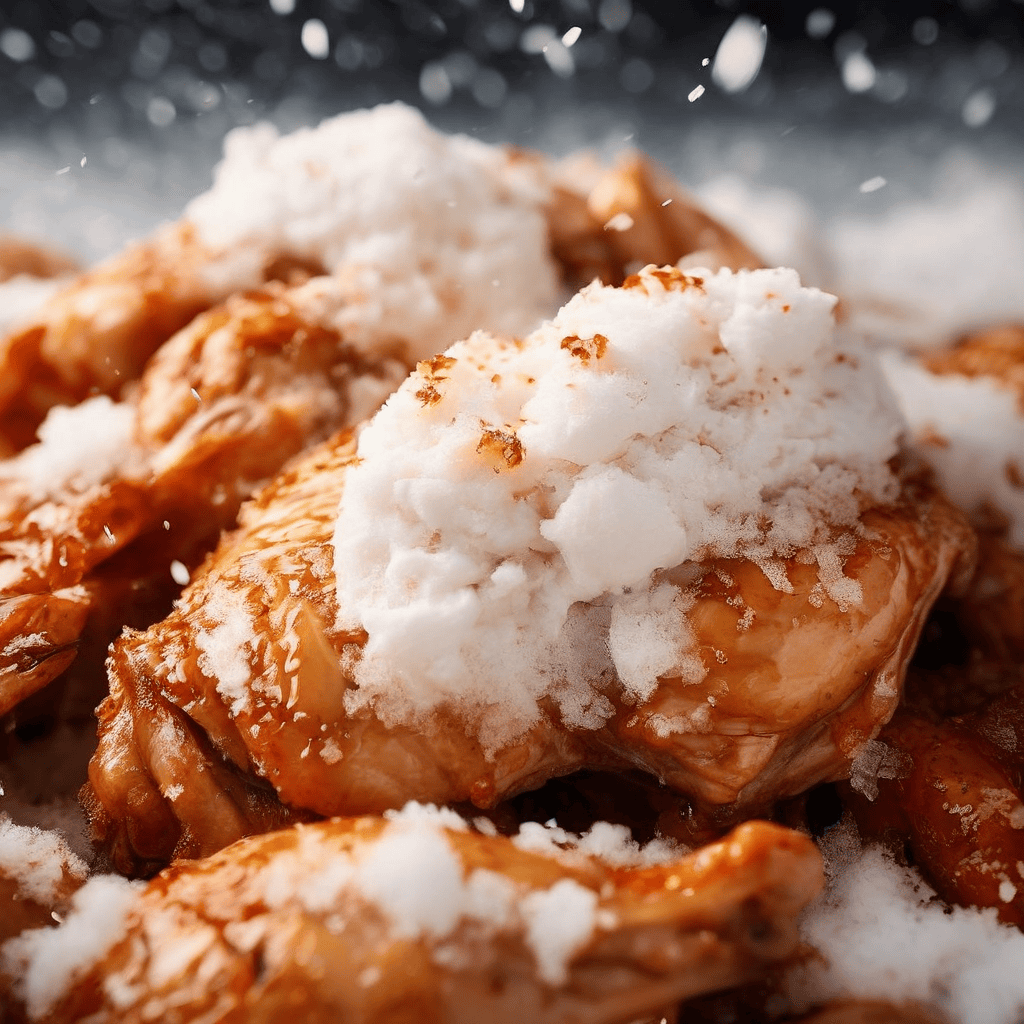} &
        \includegraphics[width=0.15\textwidth]{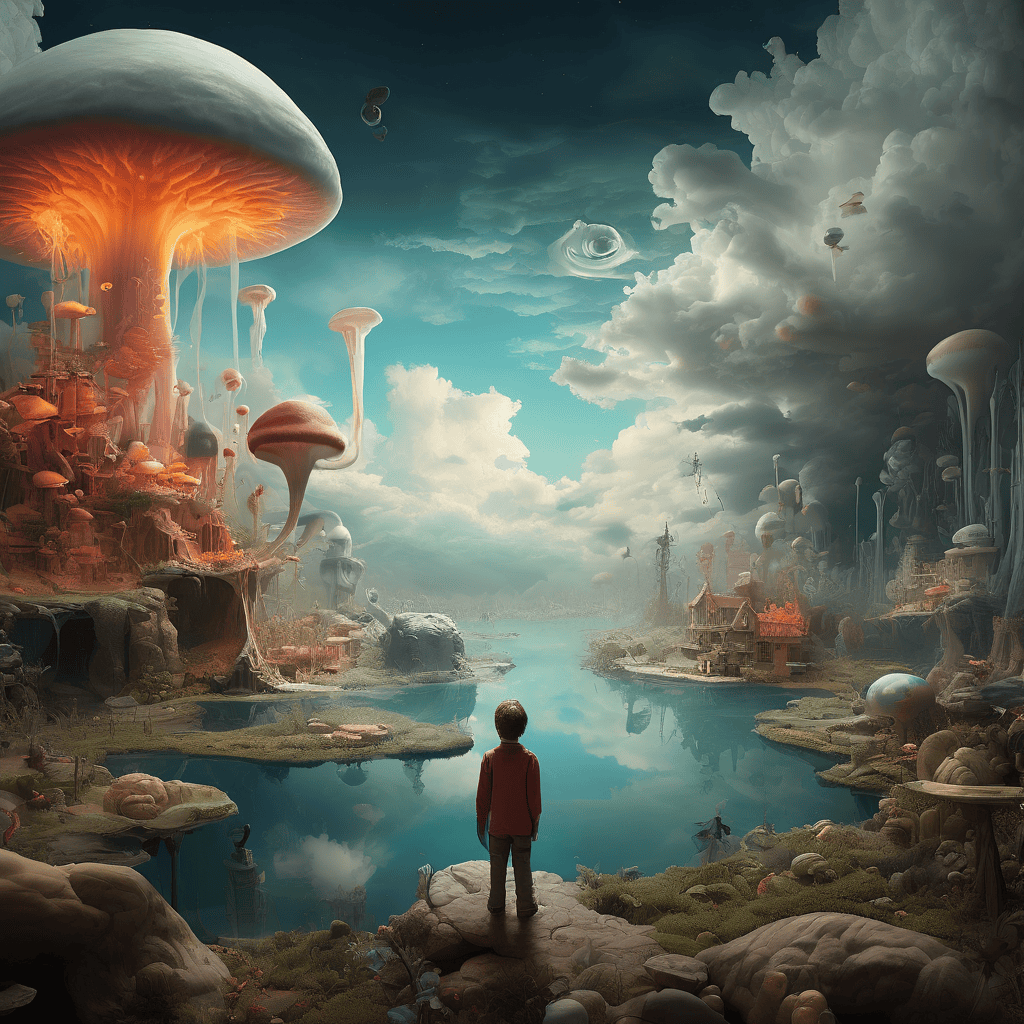} &
        \includegraphics[width=0.15\textwidth]{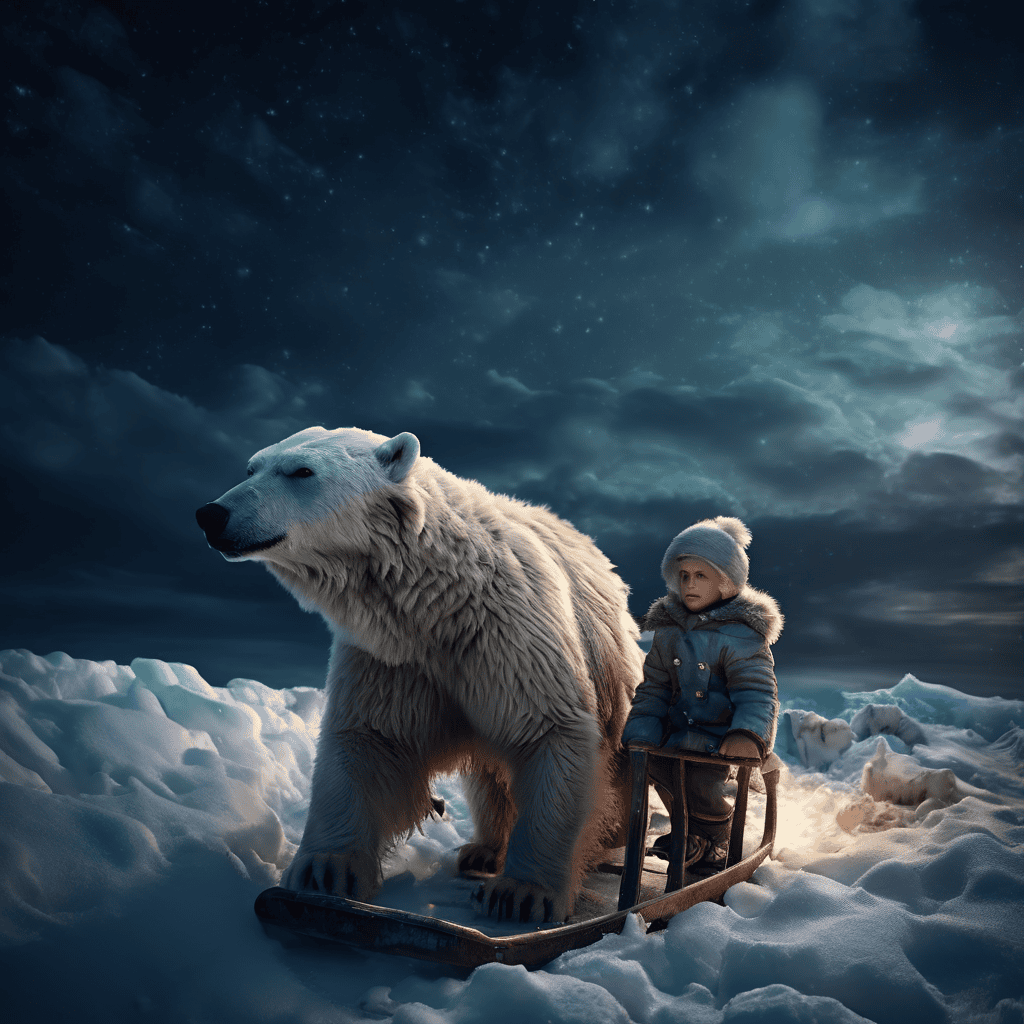} &
        \includegraphics[width=0.15\textwidth]{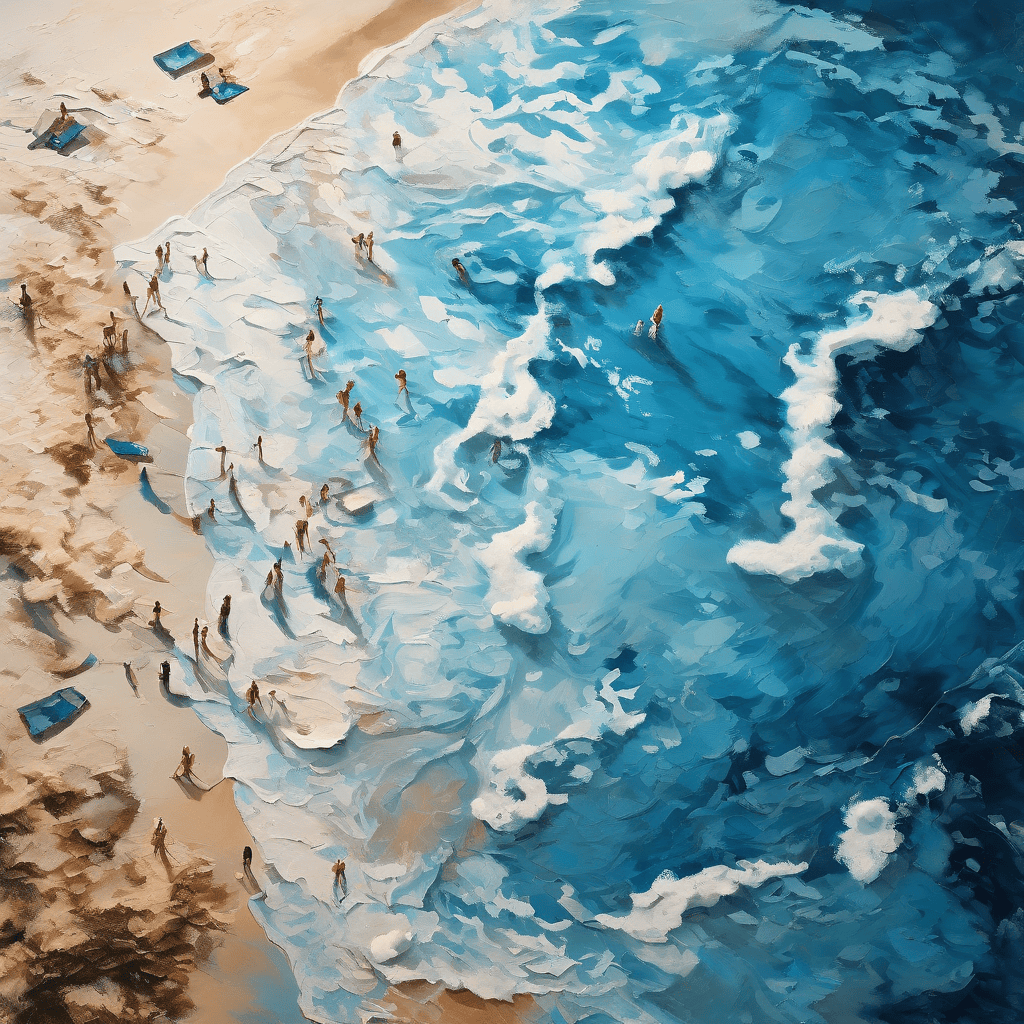} &
        \includegraphics[width=0.15\textwidth]{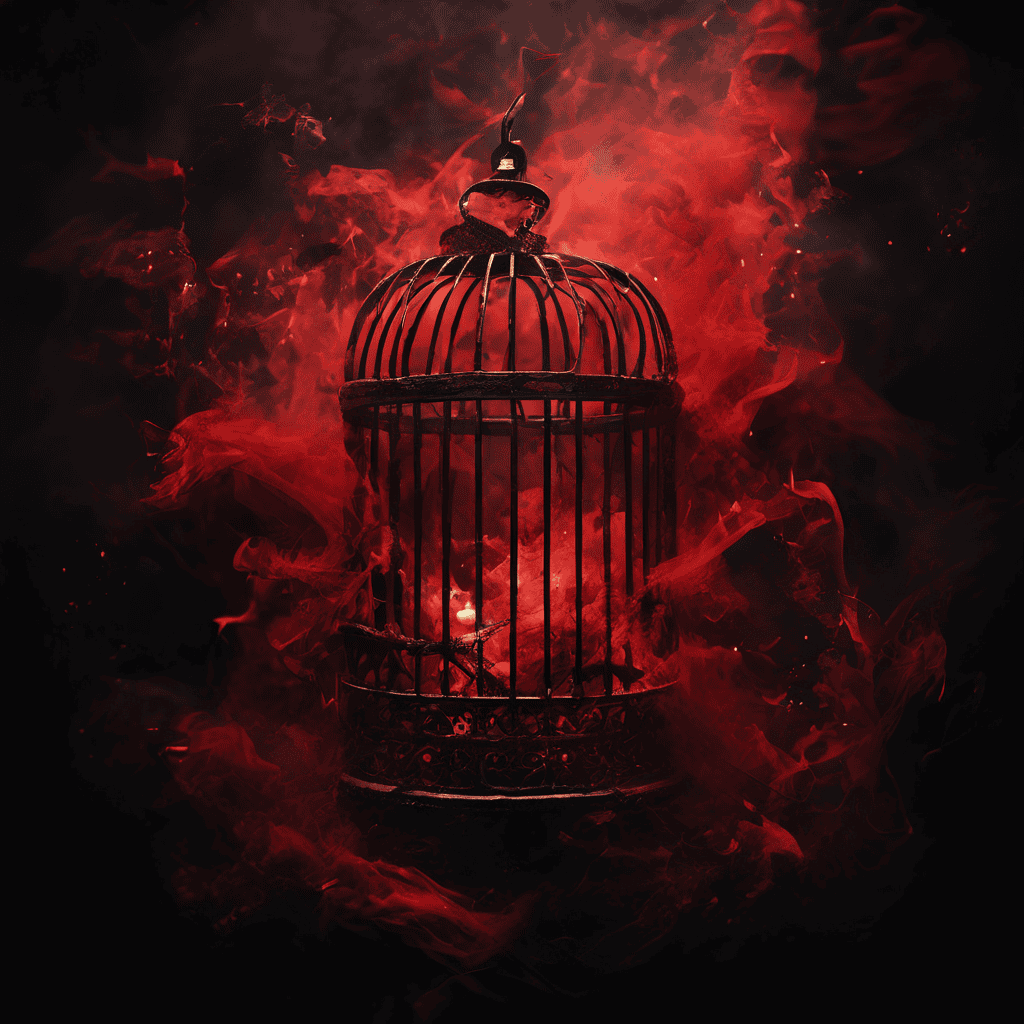} \\
        
        % x' (rescons)
        \raisebox{0.07\textwidth}{\rotatebox[origin=c]{0}{$x'$}} &
        \includegraphics[width=0.15\textwidth]{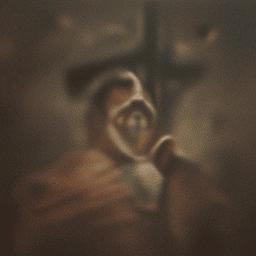} &
        \includegraphics[width=0.15\textwidth]{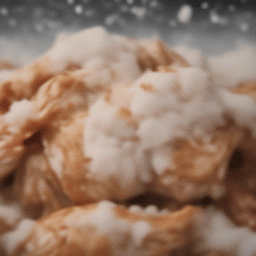} &
        \includegraphics[width=0.15\textwidth]{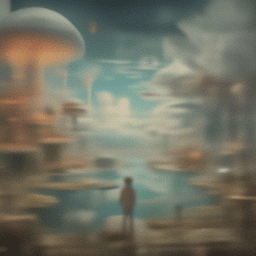} &
        \includegraphics[width=0.15\textwidth]{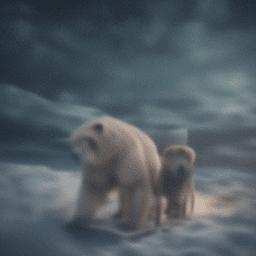} &
        \includegraphics[width=0.15\textwidth]{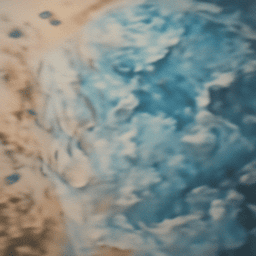} &
        \includegraphics[width=0.15\textwidth]{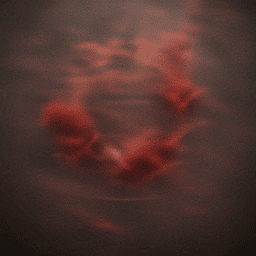} \\

        % Δx (dire)
        \raisebox{0.07\textwidth}{\rotatebox[origin=c]{0}{$\Delta x$}} &
        \includegraphics[width=0.15\textwidth]{sup_playground/017_dire.png} &
        \includegraphics[width=0.15\textwidth]{sup_playground/024_dire.png} &
        \includegraphics[width=0.15\textwidth]{sup_playground/031_dire.png} &
        \includegraphics[width=0.15\textwidth]{sup_playground/036_dire.png} &
        \includegraphics[width=0.15\textwidth]{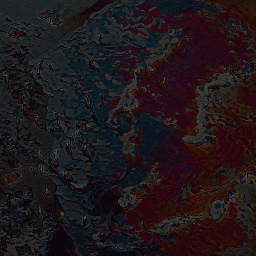} &
        \includegraphics[width=0.15\textwidth]{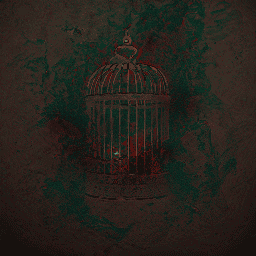} \\
        
        % x'' (rescons2)
        \raisebox{0.07\textwidth}{\rotatebox[origin=c]{0}{$x''$}} &
        \includegraphics[width=0.15\textwidth]{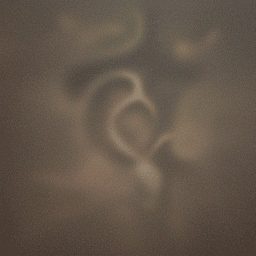} &
        \includegraphics[width=0.15\textwidth]{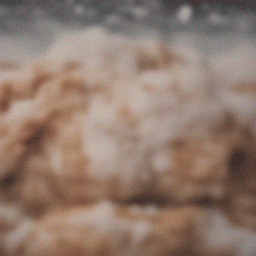} &
        \includegraphics[width=0.15\textwidth]{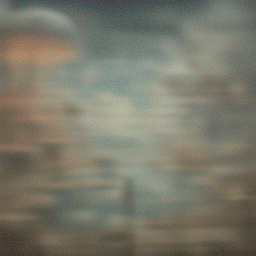} &
        \includegraphics[width=0.15\textwidth]{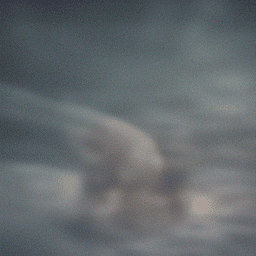} &
        \includegraphics[width=0.15\textwidth]{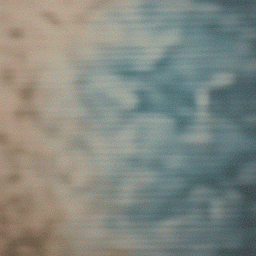} &
        \includegraphics[width=0.15\textwidth]{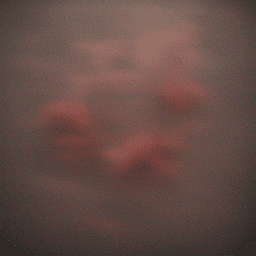} \\
        
        % Δx' (dire2)
        \raisebox{0.07\textwidth}{\rotatebox[origin=c]{0}{$\Delta x'$}} &
        \includegraphics[width=0.15\textwidth]{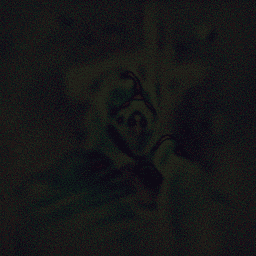} &
        \includegraphics[width=0.15\textwidth]{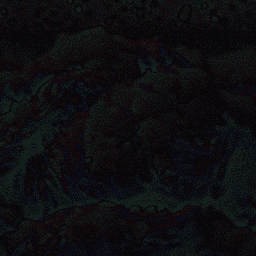} &
        \includegraphics[width=0.15\textwidth]{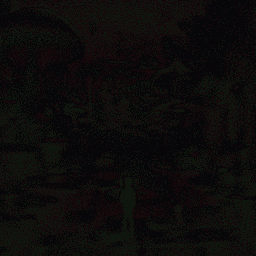} &
        \includegraphics[width=0.15\textwidth]{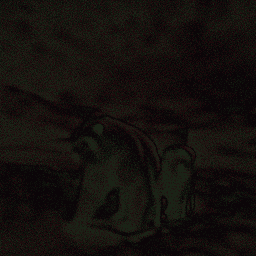} &
        \includegraphics[width=0.15\textwidth]{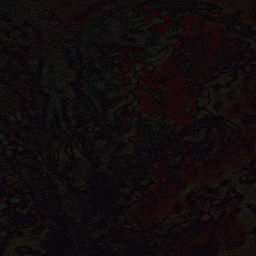} &
        \includegraphics[width=0.15\textwidth]{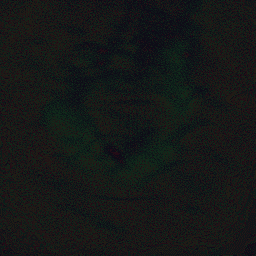} \\
        
        % Δ^2 x (diff)
        \raisebox{0.07\textwidth}{\rotatebox[origin=c]{0}{$\Delta^2 x$}} &
        \includegraphics[width=0.15\textwidth]{sup_playground/017_diff.png} &
        \includegraphics[width=0.15\textwidth]{sup_playground/024_diff.png} &
        \includegraphics[width=0.15\textwidth]{sup_playground/031_diff.png} &
        \includegraphics[width=0.15\textwidth]{sup_playground/036_diff.png} &
        \includegraphics[width=0.15\textwidth]{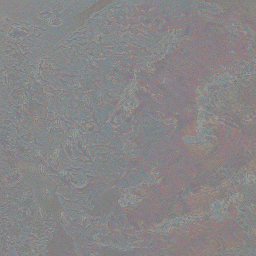} &
        \includegraphics[width=0.15\textwidth]{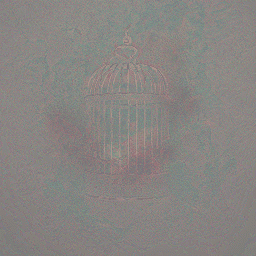} \\
    \end{tabular}
    \caption{    
    \textbf{Visualization of playground-2.5-generated images constructed for our dataset.}
    For each sample, we show the original synthesized image $x$, the first reconstruction $x'$, the second reconstruction $x''$, and their corresponding residual maps $\Delta x$, $\Delta x'$, and $\Delta^2 x$. 
    All playground-2.5 images are generated using prompts collected from Midjourney. 
    }
    \label{fig:sup_playground_matrix}
\end{figure*}

\begin{figure*}[t]
    \centering
    \setlength{\tabcolsep}{1pt}
    \renewcommand{\arraystretch}{0.1}
    \begin{tabular}{c@{\hspace{2pt}}cccccc}
        % x (real)
        \raisebox{0.07\textwidth}{\rotatebox[origin=c]{0}{$x$}} &
        \includegraphics[width=0.15\textwidth]{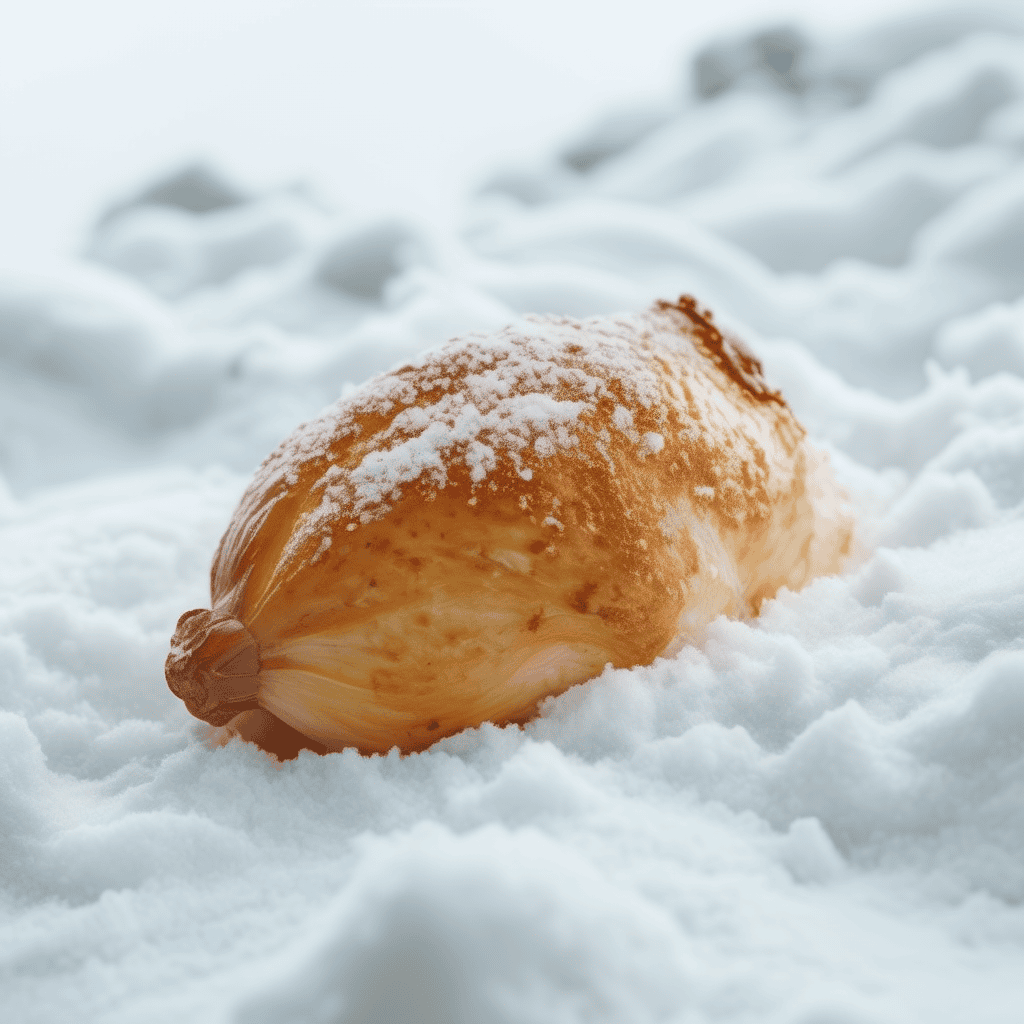} &
        \includegraphics[width=0.15\textwidth]{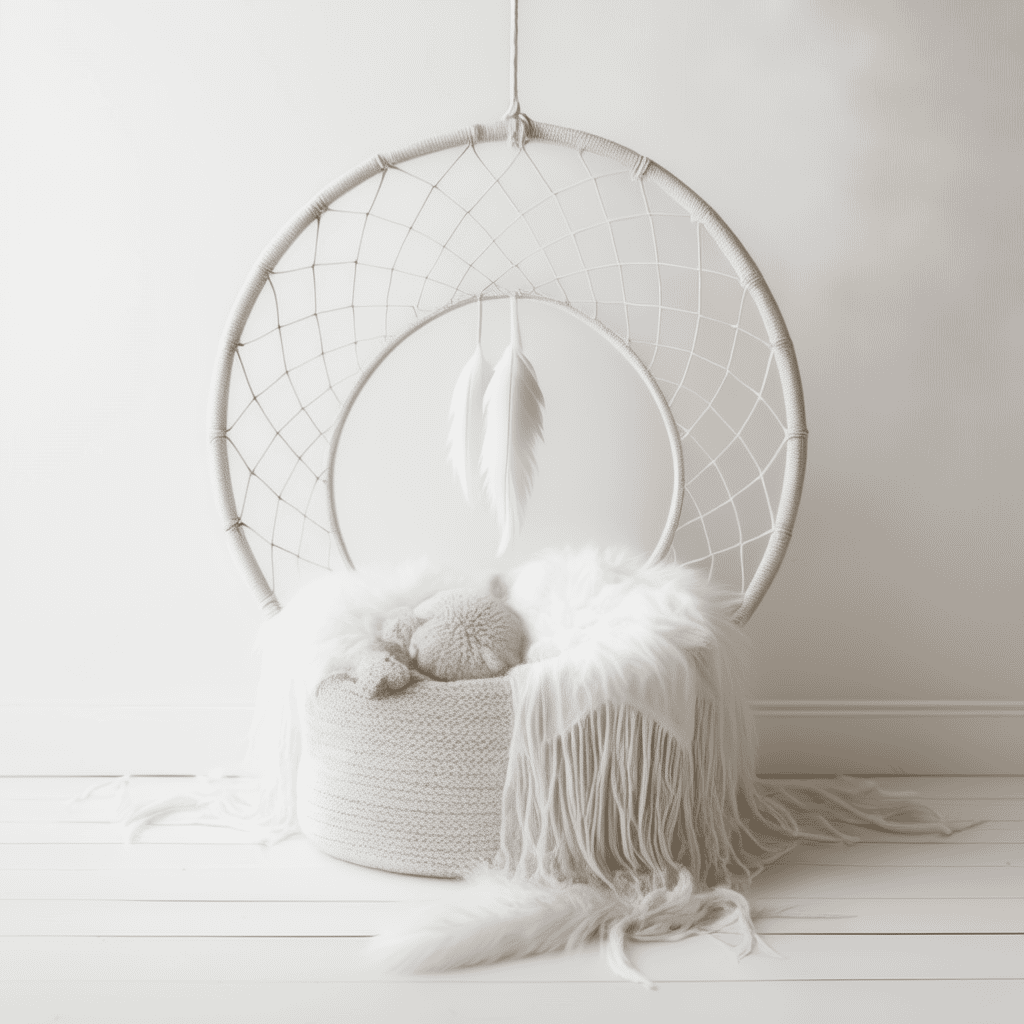} &
        \includegraphics[width=0.15\textwidth]{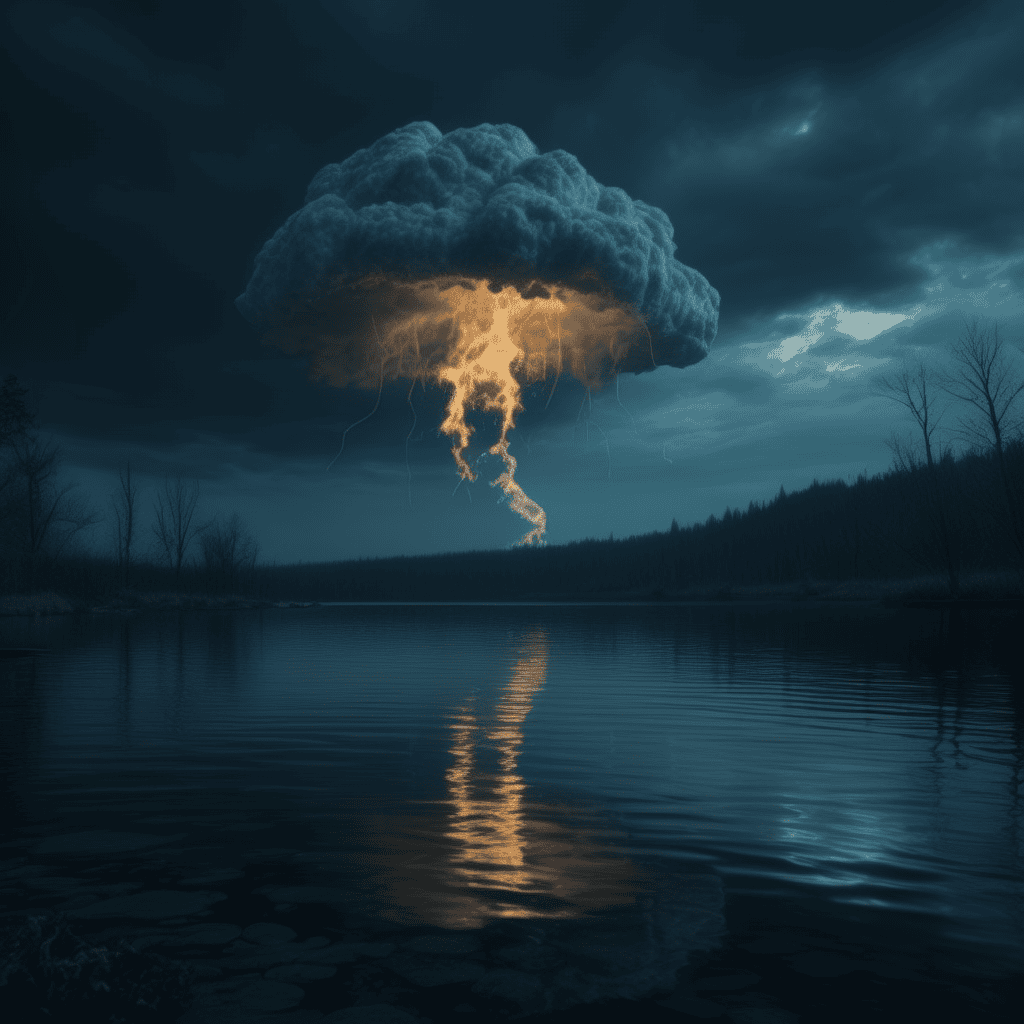} &
        \includegraphics[width=0.15\textwidth]{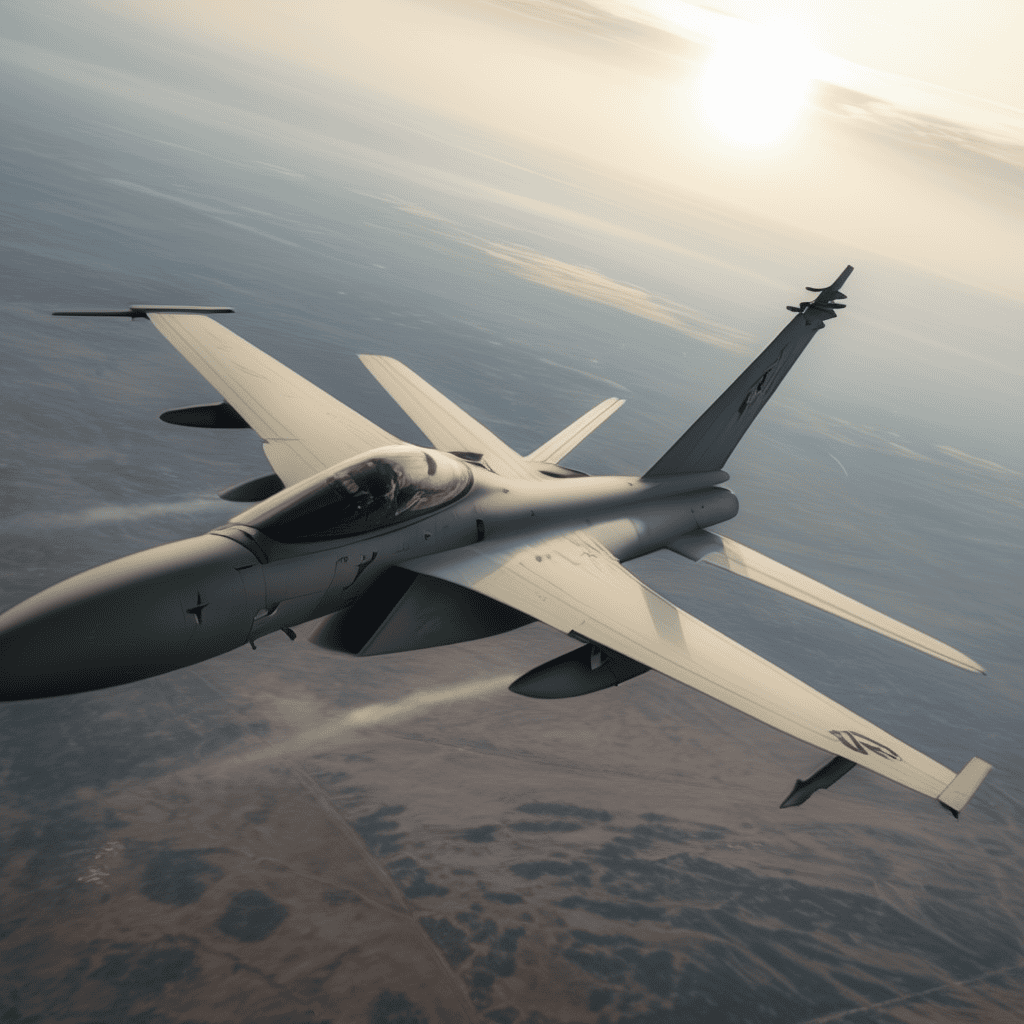} &
        \includegraphics[width=0.15\textwidth]{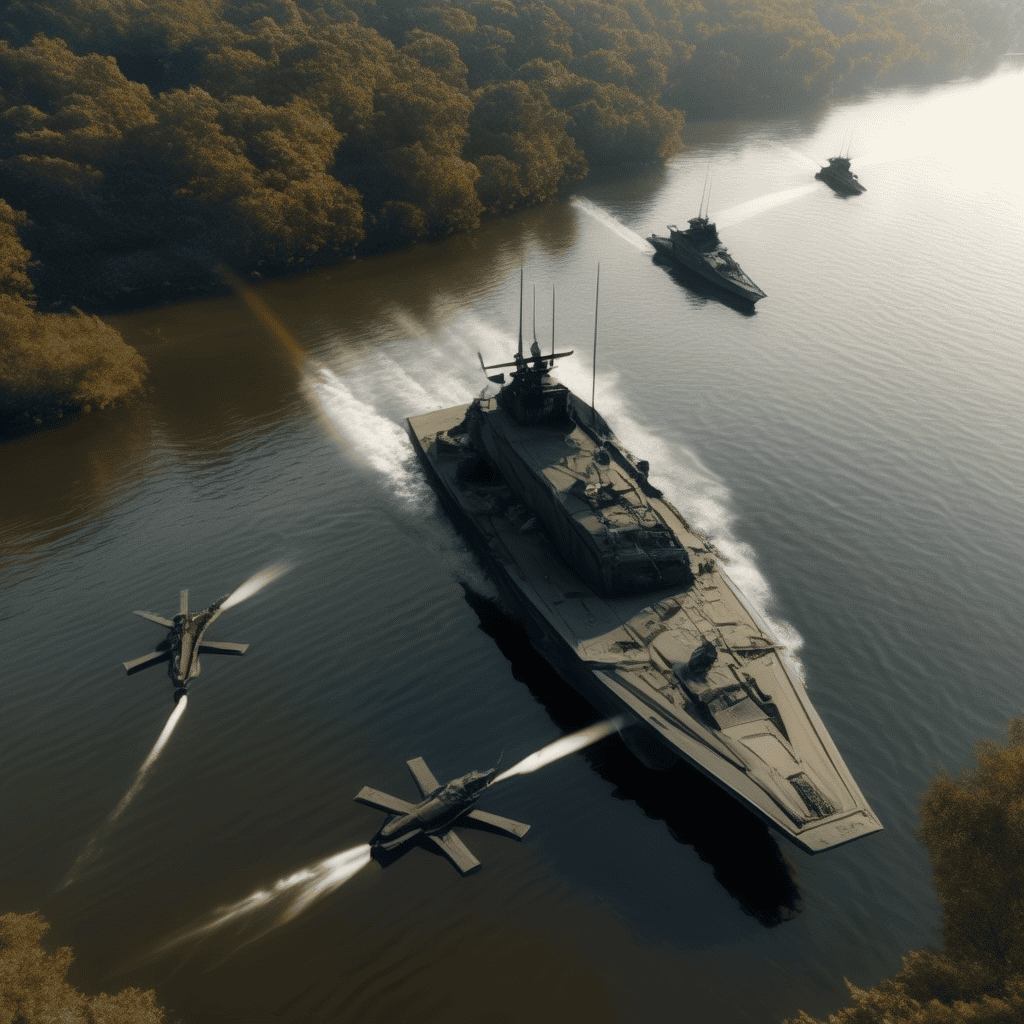} &
        \includegraphics[width=0.15\textwidth]{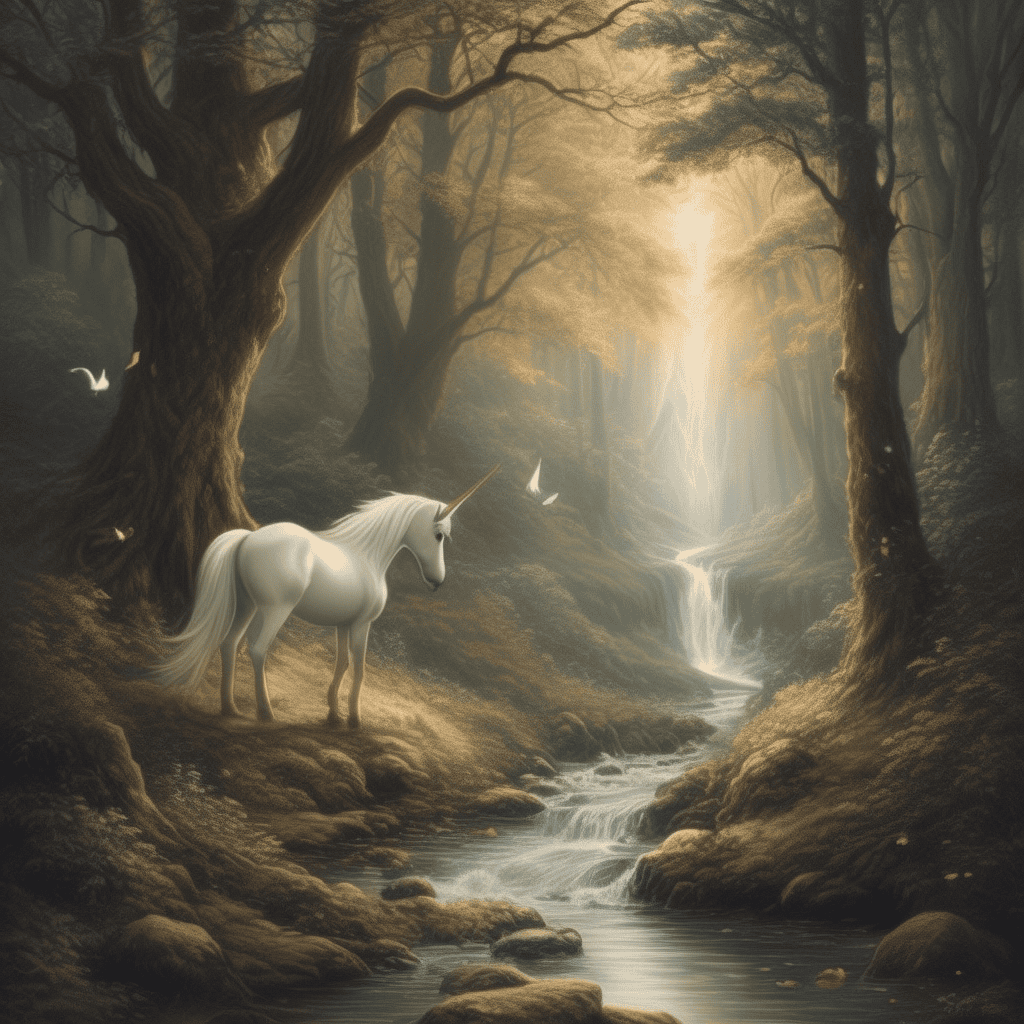} \\
        
        % x' (rescons)
        \raisebox{0.07\textwidth}{\rotatebox[origin=c]{0}{$x'$}} &
        \includegraphics[width=0.15\textwidth]{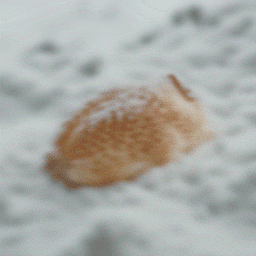} &
        \includegraphics[width=0.15\textwidth]{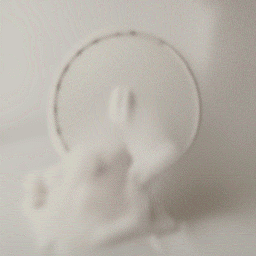} &
        \includegraphics[width=0.15\textwidth]{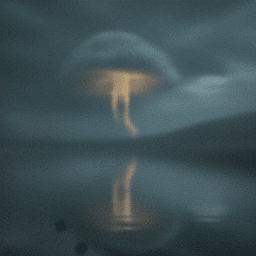} &
        \includegraphics[width=0.15\textwidth]{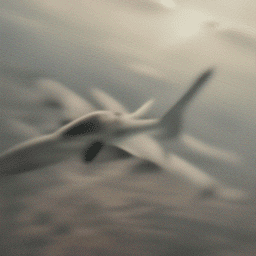} &
        \includegraphics[width=0.15\textwidth]{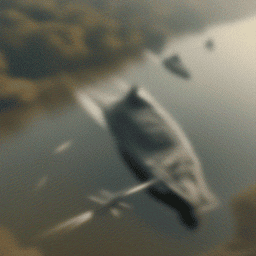} &
        \includegraphics[width=0.15\textwidth]{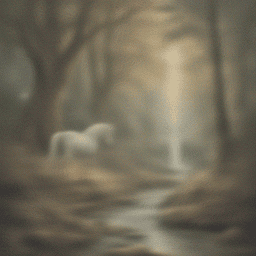} \\

        % Δx (dire)
        \raisebox{0.07\textwidth}{\rotatebox[origin=c]{0}{$\Delta x$}} &
        \includegraphics[width=0.15\textwidth]{sup_stable/003_dire.png} &
        \includegraphics[width=0.15\textwidth]{sup_stable/010_dire.png} &
        \includegraphics[width=0.15\textwidth]{sup_stable/014_dire.png} &
        \includegraphics[width=0.15\textwidth]{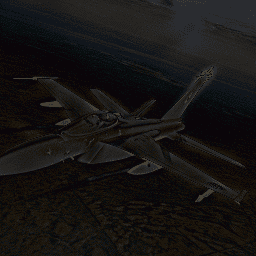} &
        \includegraphics[width=0.15\textwidth]{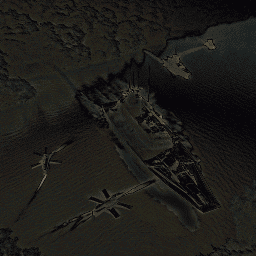} &
        \includegraphics[width=0.15\textwidth]{sup_stable/088_dire.png} \\
        
        % x'' (rescons2)
        \raisebox{0.07\textwidth}{\rotatebox[origin=c]{0}{$x''$}} &
        \includegraphics[width=0.15\textwidth]{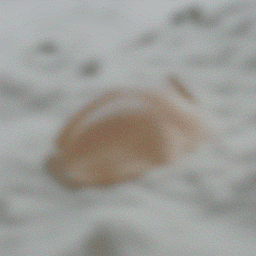} &
        \includegraphics[width=0.15\textwidth]{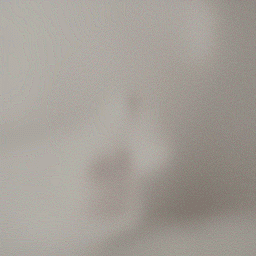} &
        \includegraphics[width=0.15\textwidth]{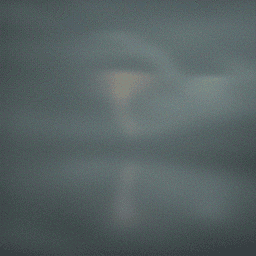} &
        \includegraphics[width=0.15\textwidth]{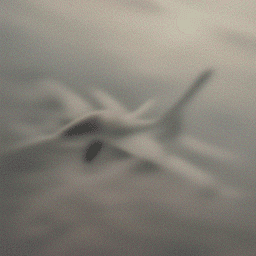} &
        \includegraphics[width=0.15\textwidth]{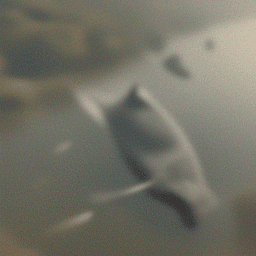} &
        \includegraphics[width=0.15\textwidth]{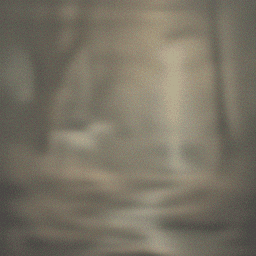} \\
        
        % Δx' (dire2)
        \raisebox{0.07\textwidth}{\rotatebox[origin=c]{0}{$\Delta x'$}} &
        \includegraphics[width=0.15\textwidth]{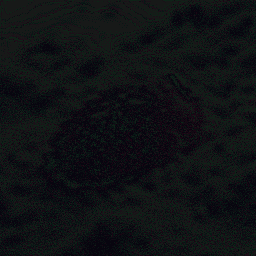} &
        \includegraphics[width=0.15\textwidth]{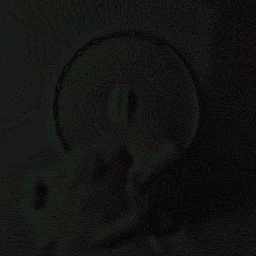} &
        \includegraphics[width=0.15\textwidth]{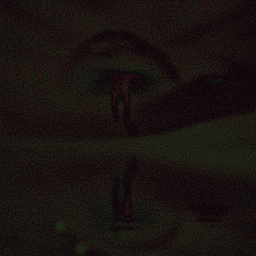} &
        \includegraphics[width=0.15\textwidth]{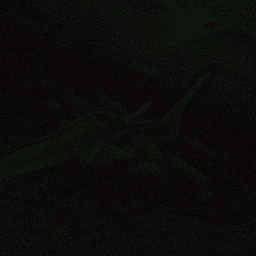} &
        \includegraphics[width=0.15\textwidth]{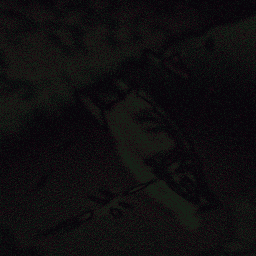} &
        \includegraphics[width=0.15\textwidth]{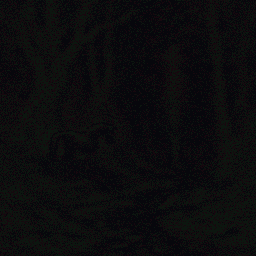} \\
        
        % Δ^2 x (diff)
        \raisebox{0.07\textwidth}{\rotatebox[origin=c]{0}{$\Delta^2 x$}} &
        \includegraphics[width=0.15\textwidth]{sup_stable/003_diff.png} &
        \includegraphics[width=0.15\textwidth]{sup_stable/010_diff.png} &
        \includegraphics[width=0.15\textwidth]{sup_stable/014_diff.png} &
        \includegraphics[width=0.15\textwidth]{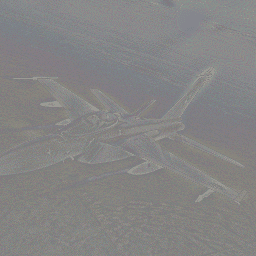} &
        \includegraphics[width=0.15\textwidth]{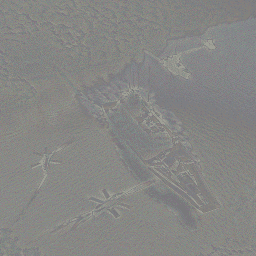} &
        \includegraphics[width=0.15\textwidth]{sup_stable/088_diff.png} \\
    \end{tabular}
    \caption{    
    \textbf{Visualization of Stable Cascade-generated images constructed for our dataset.}
    For each sample, we show the original synthesized image $x$, the first reconstruction $x'$, the second reconstruction $x''$, and their corresponding residual maps $\Delta x$, $\Delta x'$, and $\Delta^2 x$. 
    All Stable Cascad images are generated using prompts collected from Midjourney. 
    }
    \label{fig:sup_stable_matrix}
\end{figure*}

In this section, we provide additional visualizations to illustrate how DID behaves across real images and a wide range of generative models. Figure~\ref{fig:part1} summarizes the key differences between first--order and second--order residuals, while Figures~\ref{fig:sup_real_matrix}--\ref{fig:sup_stable_matrix} present more detailed residual matrices for each model. Together, these visualizations highlight the limitations of first--order residuals and the improved consistency achieved by our second--order module.

\end{document}